\documentclass[sigconf]{acmart}

\setcopyright{none}
\renewcommand\footnotetextcopyrightpermission[1]{}
\usepackage{amsmath,amsfonts,bm}

\def\eqref#1{eq.~(\ref{#1})}

\def\1{\bm{1}}

\DeclareMathAlphabet{\mathsfit}{\encodingdefault}{\sfdefault}{m}{sl}
\SetMathAlphabet{\mathsfit}{bold}{\encodingdefault}{\sfdefault}{bx}{n}

\usepackage{colortbl}       
\usepackage{amsfonts}
\usepackage{nicefrac}
\usepackage{comment}
\usepackage{threeparttable}
\usepackage{enumitem}
\usepackage{adjustbox}
\usepackage{tabularray}
\usepackage{nicematrix}
\usepackage{pifont}

\usepackage{multirow}
\usepackage{wrapfig}
\usepackage{setspace}
\usepackage{algorithm}
\usepackage{algorithmic}
\usepackage{subcaption}
\usepackage{float}
\usepackage{bbm}
\usepackage{titletoc}
\usepackage{placeins}    

\setlist[itemize]{leftmargin=0.3cm, itemsep=0.2pt, parsep=0pt, topsep=1pt}
\setlist[enumerate]{leftmargin=0.3cm, itemsep=0.2pt, parsep=0pt, topsep=1pt}
\extrafloats{200}

\newcommand{\red}[1]{\textcolor{red}{#1}}
\newcommand{\blue}[1]{\textcolor{blue}{#1}}
\newcommand{\mr}[2]{\multirow{#1}{*}{#2}}
\newcommand{\hz}[1]{\textcolor{gray}{\scriptsize$\cdot$}#1}

\definecolor{Gray}{gray}{.9}
\definecolor{LightYellow}{RGB}{255, 255, 204}
\definecolor{LightGreen}{RGB}{220, 255, 220}
\definecolor{LightGreen1}{RGB}{230, 255, 230}
\definecolor{LightGreen2}{RGB}{240, 255, 240}
\definecolor{LightGreen3}{RGB}{248, 255, 248}
\definecolor{LightGray1}{RGB}{240, 240, 240}
\definecolor{LightGray2}{RGB}{230, 230, 230}
\definecolor{LightGray3}{RGB}{220, 220, 220}
\definecolor{LightGray4}{RGB}{210, 210, 210}
\definecolor{darkgreen1}{RGB}{0,100,0}
\definecolor{darkgreen2}{RGB}{0,80,0}
\definecolor{LightTextGray}{RGB}{100,100,100}

\title[Rethinking Multimodal Fusion for Time Series Forecasting: Text Modalities Need Constrained Fusion]{Rethinking Multimodal Fusion for Time Series Forecasting: \\Text Modalities Need Constrained Fusion}

\author{Seunghan Lee}
\affiliation{%
  \institution{LG AI Research}
  \city{Seoul}
  \country{South Korea}
}
\author{Jun Seo}
\affiliation{%
  \institution{LG AI Research}
  \city{Seoul}
  \country{South Korea}
}
\author{Jaehoon Lee}
\affiliation{%
  \institution{LG AI Research}
  \city{Seoul}
  \country{South Korea}
}
\author{Sungdong Yoo}
\affiliation{%
  \institution{LG AI Research}
  \city{Seoul}
  \country{South Korea}
}
\author{Minjae Kim}
\affiliation{%
  \institution{LG AI Research}
  \city{Seoul}
  \country{South Korea}
}
\author{Tae Yoon Lim}
\affiliation{%
  \institution{LG AI Research}
  \city{Seoul}
  \country{South Korea}
}
\author{Dongwan Kang}
\affiliation{%
  \institution{LG AI Research}
  \city{Seoul}
  \country{South Korea}
}
\author{Hwanil Choi}
\affiliation{%
  \institution{LG AI Research}
  \city{Seoul}
  \country{South Korea}
}
\author{SoonYoung Lee}
\affiliation{%
  \institution{LG AI Research}
  \city{Seoul}
  \country{South Korea}
}
\author{Wonbin Ahn}
\affiliation{%
  \institution{LG AI Research}
  \city{Seoul}
  \country{South Korea}
}

\renewcommand{\shortauthors}{Lee et al.}

\acmConference[KDD '26 MILETS Workshop]%
  {12th International Workshop on Mining and Learning from Time Series}%
  {August 9--13, 2026}{Jeju, Korea}

\begin{document}

\begin{abstract}
Recent advances in multimodal learning have motivated the integration of auxiliary modalities such as text or vision into time series (TS) forecasting. However, most existing methods provide limited gains, often improving performance only in specific datasets or relying on architecture-specific designs that limit generalization.
In this paper, we show that multimodal models with naive fusion strategies (e.g., simple addition or concatenation) often underperform unimodal TS models,
which we attribute to
the \textit{uncontrolled integration of auxiliary modalities} which may introduce irrelevant information.
Motivated by this observation, we explore various \textbf{constrained fusion} methods designed to control such integration and find that they consistently outperform naive fusion methods.
Furthermore, we propose \textbf{Controlled Fusion Adapter} (\textbf{CFA}), a simple plug-in method that
enables controlled cross-modal interactions without modifying the TS backbone, integrating \textit{only relevant textual information aligned with TS dynamics}.
CFA employs
low-rank
adapters to filter irrelevant textual information before fusing it into temporal representations.
We conduct over 20K experiments across various datasets and
TS/text
models, demonstrating
the effectiveness of the constrained fusion methods.
Code is available at: \url{https://github.com/seunghan96/cfa}.
\end{abstract}

\maketitle

\section{Introduction}
Time series (TS) forecasting is widely used across domains such as finance~\cite{lee2026finstar}, traffic~\cite{cirstea2022towards}, and climate~\cite{angryk2020multivariate}.
With the emergence of large language models (LLMs), multimodal TS forecasting has gained 
attention~\cite{jiang2025multi}. 
Recent studies attempt to enhance TS forecasting by incorporating auxiliary modalities, including text~\cite{UniCast2508,CAPTime2505,BALMTSF2509,SpecTF2602}, vision~\cite{TimeVLM2502,ekambaram2020attention}, and tabular~\cite{chattopadhyay2024context}, under the assumption that 
contextual information 
enhances 
the modeling of temporal dependencies.

Existing
methods often adopt \textit{naive fusion} strategies (e.g., simple addition or concatenation)~\cite{li2025language,MultiModalForecaster2411,UniCast2508} without 
considering
how modalities should interact,
or rely on model-specific architectures\cite{liu2024timecma,CAPTime2505,SpecTF2602}
that are not easily integrated into unimodal TS models.
Additionally, prior work shows that fusion does not consistently improve performance~\cite{MultiModalForecaster2411}, 
attributing this to the difficulty of 
alignment among modalities.
Moreover, even when improvements are reported, 
we find that they 
often depend on the dataset or the TS model, indicating limitations of naive fusion strategies.

\setlength{\columnsep}{1.0pt}  
\begin{figure}[t]
\vspace{3pt}
  \centering
  \begin{adjustbox}{max width=\linewidth}
\includegraphics[width=0.385\textwidth]{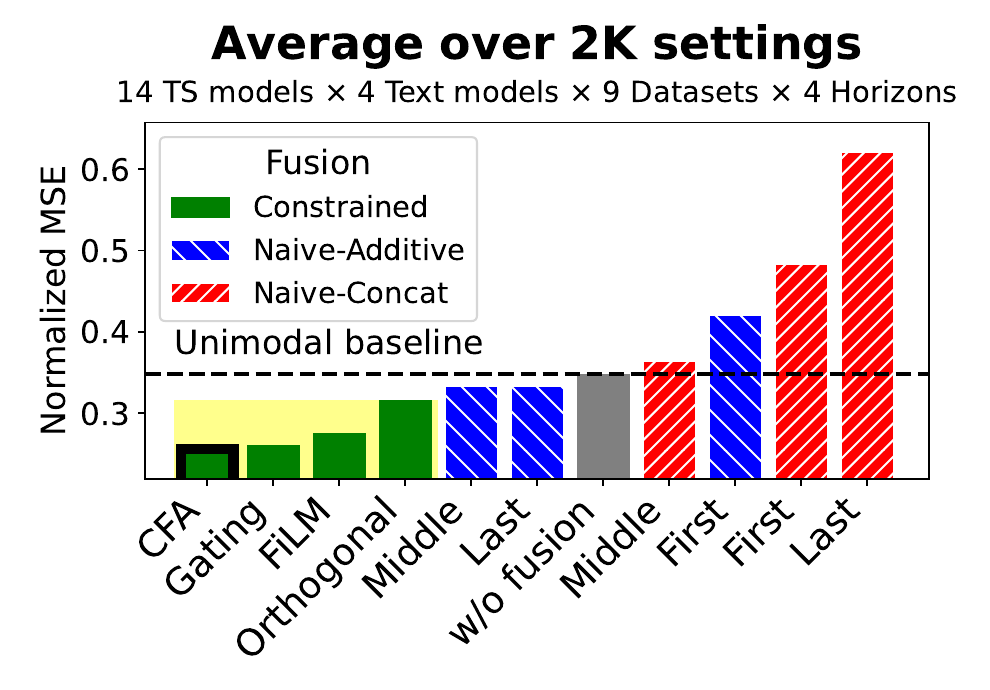}
\end{adjustbox}
\vspace{-8pt}
  \caption{Constrained vs. naive fusion. Across 2K settings, naive fusion often underperforms the unimodal baseline, while constrained fusion consistently improves over it.
  }
  \label{fig:intro1}
\vspace{-18pt}
\end{figure}
In this paper, we conduct extensive 
experiments
across diverse datasets and models (Table~\ref{sec:exp_setting}), 
evaluating fusion at the first, middle, and final layers with 
additive and concatenation operators. 
As shown in Figure~\ref{fig:intro1}, we find that 
\textit{multimodal models with naive fusion often underperform unimodal models},
as both concatenation-based (\textcolor{red}{red}) and additive (\textcolor{blue}{blue}) fusion
frequently fall below the unimodal (\textcolor{gray}{gray}) baseline.
Note that each setting is tuned with 10 learning rates, selecting the best-performing configuration.
This observation motivates the need for improved fusion mechanisms, suggesting that 
\textit{controlled fusion between TS and text is required}.

We attribute this phenomenon to the intrinsic nature of multimodal TS forecasting, where TS serve as the \textit{primary} modality, while other modalities serve as \textit{auxiliary} modalities that provide contextual guidance~\cite{su2025multimodal,TimeVLM2502, lin2026timi}, \textit{which may introduce irrelevant or conflicting information}~\cite{liu2025pa} that is misaligned with TS.
Therefore, indiscriminate fusion can degrade forecasting performance, highlighting the necessity of \textbf{constrained fusion},
which we refer to as a fusion strategy that \textit{incorporates auxiliary signals in a controlled manner} while preserving core temporal (TS) representations.

To this end, we explore various constrained fusion methods (Section~\ref{sec:constrained_fusion}), 
which consistently outperform naive methods, as shown in Figure~\ref{fig:intro1} (\textcolor{darkgreen1}{green}). 
Furthermore, we propose \textbf{C}ontrolled \textbf{F}usion \textbf{A}dapter (\textbf{CFA}), a 
plug-in
method that 
injects textual information via a residual connection constrained to a low-rank subspace,
filtering irrelevant textual information and enabling incorporation of auxiliary signals while preserving temporal dynamics.
CFA
outperforms other constrained fusion methods and 
is
effective even when naive fusion degrades performance.
The main contributions are:
\setlist[itemize]{leftmargin=0.3cm,itemsep=-1.5pt,topsep=-1.5pt, partopsep=0pt}
\begin{itemize}
\item We show that \textit{naive} multimodal fusion often underperforms 
unimodal TS forecasting and demonstrate that \textit{constrained} fusion consistently improves over naive strategies across diverse datasets and models,
where we evaluate four constrained fusion methods, including our proposed method.
\item 
We propose Controlled Fusion Adapter (CFA), a simple yet effective fusion method for multimodal TS forecasting,
which injects auxiliary textual information into TS representations via a residual connection constrained to a low-rank subspace, enabling controlled integration without modifying the backbone.
In contrast to prior multimodal TS forecasting works---which are either \textit{architecture-specific} and thus tied to a particular backbone~\cite{chowdhury2025t3time, liu2024timecma, SpecTF2602, CAPTime2505, TimeVLM2502, BALMTSF2509}, or \textit{plug-in but restricted to naive fusion}~\cite{jia2024gpt4mts, jin2023time, MultiModalForecaster2411, li2025language, liu2024time, UniCast2508}---CFA is the only method that is \textit{both} a plug-in module applicable to \textit{any unimodal TS model} and a \textit{constrained} fusion approach that filters irrelevant auxiliary information (Table~\ref{tab:fusion_plugin_summary}).
\item We conduct over 20K experiments across various settings 
(9 
multimodal datasets, 4 forecasting horizons, 14 TS models, and 4 text models) with 10 fusion strategies, demonstrating the effectiveness of constrained fusion including CFA. Additionally, we provide detailed analyses explaining \textit{why constrained fusion strategies yield superior performance} compared to naive fusion strategies.
\end{itemize}

\begin{table}[t]
\caption{Multimodal forecasting models. 
Existing multimodal forecasting models are either 1) tied to a specific backbone or 2) rely on naive fusion, while CFA combines plug-in applicability with constrained fusion. A detailed discussion of related work is provided in Appendix~\ref{sec:related_works}.}
\vspace{-3pt}
\centering
\begin{adjustbox}{max width=0.5\textwidth}
\begin{tabular}{cc|c}
\toprule
\multicolumn{2}{c|}{Type} & Methods \\
\midrule
\multicolumn{2}{c|}{\cellcolor{LightGray1}Architecture-specific} & \cellcolor{LightGray1}\cite{chowdhury2025t3time, liu2024timecma, SpecTF2602,CAPTime2505, TimeVLM2502, BALMTSF2509} \\
\midrule
\multirow{2.5}{*}{\textbf{Plug-in}} & 
Naive & \cite{jia2024gpt4mts, jin2023time, MultiModalForecaster2411, li2025language, liu2024time, UniCast2508} \\
\cmidrule{2-3}
& \cellcolor{LightYellow} \textbf{Constrained} & \cellcolor{LightYellow} \textbf{CFA} (Ours) \\
\bottomrule
\end{tabular}
\end{adjustbox}
\label{tab:fusion_plugin_summary}
\vspace{-7pt}
\end{table}

\section{Necessity of Constrained Fusion for Multimodal TS Forecasting}
\label{sec:problem_formulation}
\definecolor{fusionblue}{RGB}{0, 51, 153}  
In multimodal forecasting,
a model predicts future values
$\mathbf{y} = (\mathbf{x}_{L+1}, \ldots, \mathbf{x}_{L+H})$
given a lookback window $\mathbf{x} = (\mathbf{x}_1, \ldots, \mathbf{x}_L)$ and a paired text sequence $\mathbf{t} = (\mathbf{t}_1, \ldots, \mathbf{t}_L)$.
Each $\mathbf{x}_i \in \mathbb{R}^C$ denotes 
observations at time step $i$, where $L$, $H$, and $C$ represent the lookback length, forecast horizon, and number of channels, respectively.
Each text $\mathbf{t}_i$ is encoded by a 
language model as $\mathbf{z}_{\text{Text},i} = g_{\text{Text}}(\mathbf{t}_i)$, forming a text embedding
 $\mathbf{Z}_{\text{Text}} = 
(\mathbf{z}_{\text{Text},1}, \ldots, \mathbf{z}_{\text{Text},L})$.

\subsection{Naive Fusion}
\setlength{\columnsep}{8pt}  
\begin{figure}[t]
    \begin{minipage}{\linewidth}
      \begin{algorithm}[H]
        \caption{\textbf{Multimodal TS forecasting.} First / Middle / Last fusions are mutually exclusive.}
        \small
        \label{alg:MTSF}
        \renewcommand{\algorithmicrequire}{\textbf{Input:}}
        \renewcommand{\algorithmicensure}{\textbf{Output:}}
        \begin{algorithmic}
        \REQUIRE $\mathbf{X} = [\mathbf{X}_1, \ldots ,\mathbf{X}_L]$, $\mathbf{T} = [\mathbf{T}_1, \ldots ,\mathbf{T}_L]$
        \ENSURE $\hat{\mathbf{Y}} = [\hat{\mathbf{X}}_{L+1}, \ldots, \hat{\mathbf{X}}_{L+H}]$
        \STATE $\mathbf{Z}_{\text{TS}} \leftarrow g_{\text{TS}}(\mathbf{X})$ \hfill {// Input projection (TS)}
        \STATE $\mathbf{Z}_{\text{Text}} \leftarrow g_{\text{Text}}(\mathbf{T})$ \hfill {// Input projection (Text)}

        \STATE {\color{fusionblue}
        $\mathbf{Z}_{\text{TS}} \leftarrow 
        \mathcal{F}(\mathbf{Z}_{\text{TS}}, \mathbf{Z}_{\text{Text}})$}
        \hfill {\color{fusionblue} // First fusion}

        \FOR{$m$ in \text{encoder layers}}
            \STATE $\mathbf{Z}_{\text{TS}} \leftarrow f(\mathbf{Z}_{\text{TS}})$
            \STATE {\color{fusionblue}
            $\mathbf{Z}_{\text{TS}} \leftarrow 
            \mathcal{F}(\mathbf{Z}_{\text{TS}}, \mathbf{Z}_{\text{Text}})$}
            \hfill {\color{fusionblue} // Middle fusion}
        \ENDFOR

        \STATE {\color{fusionblue}
        $\mathbf{Z}_{\text{TS}} \leftarrow 
        \mathcal{F}(\mathbf{Z}_{\text{TS}}, \mathbf{Z}_{\text{Text}})$}
        \hfill {\color{fusionblue} // Last fusion}

        \STATE $\hat{\mathbf{Y}} \leftarrow h(\mathbf{Z}_{\text{TS}})$ \hfill {// Output projection (TS)}
        \end{algorithmic}
      \end{algorithm}
    \end{minipage}
\vspace{-10pt}
\end{figure}

\begin{figure*}[t]
\centering
\begin{adjustbox}{max width=\linewidth}
\includegraphics[width=.999\textwidth]{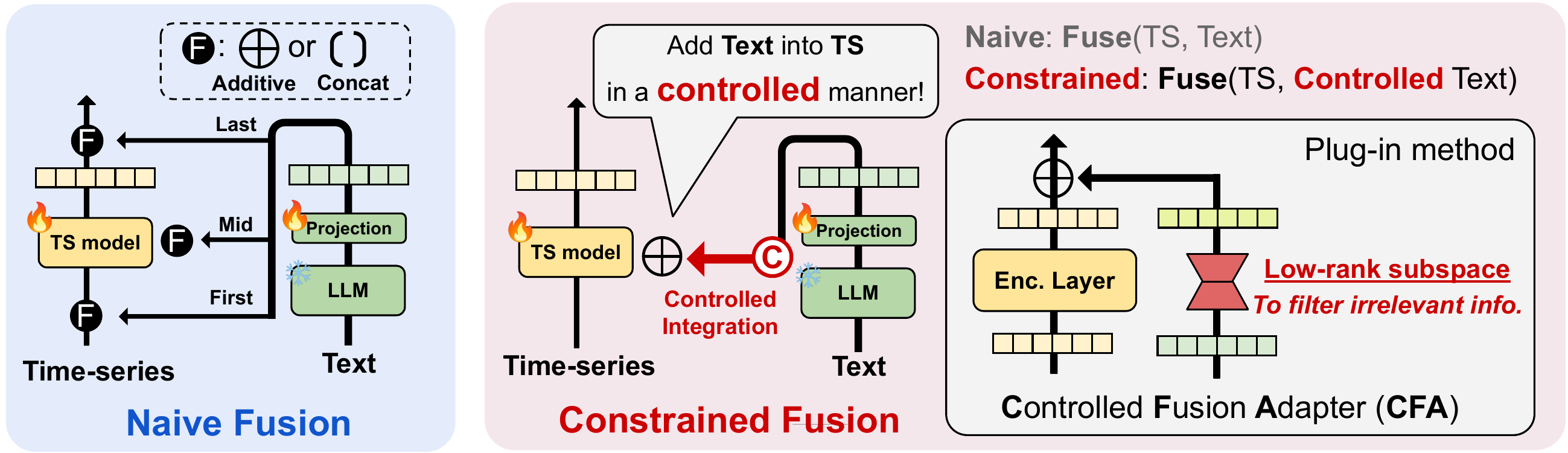}
\end{adjustbox}
\caption{
\textbf{Comparison of multimodal fusion strategies for TS.}
(\textbf{Left}) \textit{Naive fusion} applies simple additive or concatenation operators at first, middle, or last stages without considering modality relevance.
(\textbf{Right}) 
\textit{Constrained fusion} incorporates textual information in a controlled manner by considering its relevance to TS.
\textit{CFA} injects textual signals via a residual connection constrained to a low-rank subspace to filter irrelevant information while preserving TS representations.
}
\label{fig:main}
\vspace{-10pt}
\end{figure*}

Naive fusion integrates textual information using a fusion operator (e.g., addition or concatenation), where the text embeddings produced by the text encoder are \textit{directly} fused with TS embeddings \textit{without any constraint}.
Fusion can be applied before the encoder (first fusion), within intermediate encoder layers (middle fusion), or after temporal encoding 
(last fusion), as shown in Algorithm~\ref{alg:MTSF}.
Many multimodal TS forecasting methods follow this approach~\cite{jia2024gpt4mts, jin2023time, MultiModalForecaster2411, li2025language, liu2024time, UniCast2508},
whereas methods that do not follow this paradigm typically adopt architecture-specific designs to fuse different modalities~\cite{chowdhury2025t3time, SpecTF2602, CAPTime2505, TimeVLM2502, BALMTSF2509}, which are not generally applicable to existing TS models.


\subsection{Constrained Fusion}
\label{sec:constrained_fusion}
In this paper, we argue that TS forecasting primarily relies on learning temporal representations, while auxiliary modalities provide contextual guidance \textit{that may contain information irrelevant to TS}. However, naive fusion methods 
do not account for this misalignment, which may disrupt temporal representations and underperform a unimodal (TS-only) model.
 To investigate this effect, we explore various constrained fusion strategies\footnote{Details of the constrained fusion strategies are discussed in Appendix~\ref{app:constrained}.}, which incorporate auxiliary information from other modalities in a constrained manner while preserving temporal representations as follows:

\setlist[itemize]{leftmargin=0.3cm,itemsep=-1.5pt,topsep=-1.5pt, partopsep=0pt}
\begin{itemize}
\item \textbf{[1] Gating mechanism} determines the relevance of text information at each time step using a learned gate, 
allowing the model to utilize only the necessary textual information. 
\label{sec:related_works_v2}

\item \textbf{[2] FiLM (Feature-wise Linear Modulation)}\footnote{Note that
FiLM~\cite{perez2018film}
is different from
(TS model) FiLM~\cite{Zhou2022FiLM}, 
discussed in Section~\ref{sec:related_works}.}~\cite{perez2018film} modulates the scale and bias of TS embeddings based on text embeddings, 
preserving temporal structure while adjusting 
representations.
\item \textbf{[3] Orthogonal fusion} projects 
text embeddings
into 
TS embedding space and injects only the orthogonal component, explicitly preserving temporal information without overwriting it.
\end{itemize}

\textbf{Naive fusion vs. Constrained fusion.}
As shown in Figure~\ref{fig:intro1}, 
these \textit{constrained fusion strategies consistently outperform naive fusion strategies} across various settings.
Nonetheless, they can \textit{still underperform unimodal baselines under certain settings}, as shown in Table~\ref{tab:horizon_avg}. 
This observation motivates the need to more effectively filter irrelevant textual signals while preserving informative ones.
A comparison of naive and constrained fusion is provided in Figure~\ref{fig:main}.

\subsection{Controlled Fusion Adapter (CFA)}

\label{sec:CFA}
Building on these insights, we propose Controlled Fusion Adapter (\textbf{CFA}), a model-agnostic fusion method where text embeddings are projected through a low-dimensional bottleneck and added to the TS embedding as a small residual.
It is important to note that our goal is \textit{not to develop a novel adapter} but to design a \textit{controlled fusion formulation} that constrains the textual signal effectively.
To implement this, we adopt a LoRA-style parameterization~\cite{hu2022lora}, enabling lightweight integration into diverse TS backbones.
The low-rank bottleneck limits textual capacity and encourages retention of information useful for TS forecasting while filtering irrelevant components.
This design preserves the TS backbone and enables text-guided integration into any TS model as:
\begin{subequations}
\begin{align}
    \mathbf{z}_{\text{Adapter},t}   &= \mathbf{W}_\text{down}\, \mathbf{z}_{\text{Text},t} \;\in \mathbb{R}^{D/r} \\
    \mathbf{z}_{\text{Adapter},t}'  &= \text{ReLU}(\text{LayerNorm}(\mathbf{z}_{\text{Adapter},t})) \\
    \mathbf{z}_{\text{Adapter},t}'' &= \mathbf{W}_\text{up}\, \mathbf{z}_{\text{Adapter},t}' \;\in \mathbb{R}^{D} \\
    \tilde{\mathbf{z}}_{\text{TS},t} &= \mathbf{z}_{\text{TS},t} + \textcolor{red!80!black}{\mathbf{z}_{\text{Adapter},t}''}.
\end{align}
\end{subequations}

\begin{table}[t]
\centering
\caption{Comparison of \textcolor{red!80!black}{constrained fusions.}}
\label{tbl:compare_select}
\begin{adjustbox}{max width=\linewidth}
\begin{tabular}{c|l|c}
\toprule
\multirow{5.5}{*}{\rotatebox{90}{Methods}} &  Gating & $\mathbf{z}_{\text{TS},t} + \textcolor{red!80!black}{\mathbf{g}_t \odot \mathbf{z}_{\text{Text},t}}$ \\
\cmidrule{2-3}
& FiLM & $\textcolor{red!80!black}{\boldsymbol{\gamma}_t} \odot \mathbf{z}_{\text{TS},t} + \textcolor{red!80!black}{\boldsymbol{\beta}_t}$ \\
\cmidrule{2-3}
& Orthogonal & $\mathbf{z}_{\text{TS},t} + \textcolor{red!80!black}{\mathbf{z}_{\text{Text},t}^{\perp}}$ \\
\cmidrule{2-3}
 & \cellcolor{LightYellow} \textbf{CFA} & \cellcolor{LightYellow} $\mathbf{z}_{\text{TS},t} + \textcolor{red!80!black}{\mathbf{W}_{\text{up}}\,\phi(\mathbf{W}_{\text{down}}\mathbf{z}_{\text{Text},t})}$. \\
\bottomrule
\end{tabular}
\end{adjustbox}
\end{table}

Here, $\tilde{\mathbf{z}}_{\text{TS},t}$ denotes the fused embedding at time step $t$, 
and the residual addition is applied at each encoder layer of any TS model.
The bottleneck dimension $D/r$ controls parameter efficiency, 
where we set 
$r=8$.
Note that initializing $\mathbf{W}_\text{up}$ near zero ensures that textual influence is minimal at the early stages of training.
A comparison of how the four constrained fusion strategies
construct the fused embedding is shown in Table~\ref{tbl:compare_select},
and robustness to the choice of $r$ is discussed in Appendix~\ref{app:sensitivity_analysis}.



\begin{table}[t]
\caption{Experimental settings across 20K configurations.
}
\centering
\begin{adjustbox}{width=\linewidth}
\begin{NiceTabular}{c|c|c|l}
\toprule
\multicolumn{4}{l}{\cellcolor{LightGray1}\textit{\textbf{Experimental settings}} ([1] $\times$ [2] $\times$ [3] $>2K$)} \\ 
\midrule
\multicolumn{3}{c}{\multirow{2}{*}{\textbf{[1] Datasets} (Time-MMD~\cite{liu2024time})}} & \multirow{2}{*}{\shortstack[l]{Agriculture, Climate, Economy, Energy, Environment \\ Public Health, Security, Social Good, Traffic}}
 \\
 \multicolumn{3}{c}{} & \\
\midrule
\multirow{8}{*}{\textbf{[2] Models}} & \multirow{6}{*}{TS} & \multirow{3}{*}{Transformer} & \multirow{3}{*}{\shortstack[l]{Nonstationary Transformer~\cite{Liu2022NonstationaryTransformer}, PatchTST~\cite{Nie2023PatchTST}, iTransformer~\cite{Liu2023iTransformer} \\ Crossformer~\cite{zhang2023crossformer}, FEDformer~\cite{zhou2022fedformer}, Autoformer~\cite{wu2021autoformer} \\ Reformer~\cite{kitaev2020reformer}, Informer~\cite{zhou2021informer}, Transformer~\cite{vaswani2017attention}}} \\
& & & \\
& & & \\
\cmidrule{3-4}
 &  & Linear/MLP & DLinear~\cite{Zeng2022DLinear}, TiDE~\cite{Das2023TiDE}, TSMixer~\cite{Chen2023TSMixer} \\
 \cmidrule{3-4}
 &  & Others & Koopa~\cite{Koopa2023}, FiLM~\cite{Zhou2022FiLM} \\
 \cmidrule{2-4}
 & \multicolumn{2}{c}{Text} & BERT~\cite{devlin2018bert}, GPT2~\cite{radford2019gpt2}, Llama3~\cite{dubey2024llama3}, Doc2Vec~\cite{le2014doc2vec} \\

\midrule
\multicolumn{3}{c}{\textbf{[3] Forecasting horizons} ($H$)} &
Daily: \{48, 96, 192, 336\}.
Weekly: \{12, 24, 36, 48\}.
Monthly: \{6, 8, 10, 12\}.
\\
\midrule
\multicolumn{4}{l}{\cellcolor{LightGray1}\textit{\textbf{Fusion methods}} (10)} \\
\midrule
\multicolumn{2}{c}{\multirow{2.5}{*}{\textbf{Naive}}} & Additive & First~\cite{li2025language, jia2024gpt4mts}, Middle, Last~\cite{liu2024time} \\
\cmidrule{3-4}
 \multicolumn{2}{c}{ } & Concat & First~\cite{UniCast2508, MultiModalForecaster2411, jin2023time}, Middle, Last \\
 \midrule
\multicolumn{3}{c}{\textbf{Constrained}} & Orthogonal, FiLM~\cite{perez2018film}, Gating~\cite{TimeVLM2502,chowdhury2025t3time}, \textbf{CFA} (Ours) \\
\bottomrule
\end{NiceTabular}
\end{adjustbox}
\label{sec:exp_setting}
\end{table}

\begin{table*}[!tbp]
\caption{\textbf{Comparison of multimodal fusion strategies for TS forecasting.}
While naive fusion strategies
often underperform the unimodal baseline, \textit{constrained} fusion methods consistently perform better, with our \textit{CFA showing robust improvements across diverse
settings}.
\textbf{\red{Red}}/\blue{blue} denotes \textcolor{red}{improvements}/\textcolor{blue}{degradations} over the unimodal model, respectively, averaged across four $H$s. Divergence (\blue{\textit{Div.}}) indicates cases where the MSE exceeds that of the unimodal baseline by more than $10\times$.
}
\centering
\begin{adjustbox}{max width=0.9\linewidth, max totalheight=0.815\textheight}
\begin{tabular}{c|c|l|c|ccc|ccc|ccc>{\columncolor{LightYellow}}c}
\toprule
\multicolumn{1}{c|}{}
 & \multicolumn{1}{c|}{\multirow{5}{*}{\shortstack{TS \\ Model}}}
 & \multicolumn{1}{c|}{\multirow{5}{*}{Dataset}}
 & \multicolumn{1}{c|}{\multirow{5}{*}{\shortstack{\textbf{Unimodal}\\\textbf{(w/o text)}}}}
 & \multicolumn{6}{c|}{\textbf{Naive}}
 & \multicolumn{4}{c}{\multirow{2.5}{*}{\textbf{Constrained}}}
 \\
\cmidrule(lr){5-10}
 & & & & \multicolumn{3}{c|}{\textbf{Additive}} & \multicolumn{3}{c|}{\textbf{Concat}} & \multicolumn{4}{c}{} \\
\cmidrule(lr){5-7}\cmidrule(lr){8-10}\cmidrule(lr){11-14}
 & & & & \shortstack{First \\ \cite{li2025language, jia2024gpt4mts}} & \shortstack{Middle \\ (---)} & \shortstack{Last \\ \cite{liu2024time}} & \shortstack{First \\ \cite{UniCast2508, MultiModalForecaster2411, jin2023time}} & \shortstack{Middle \\ (---)} & \shortstack{Last \\ (---)} & \shortstack{Orthogonal \\ (---)} & \shortstack{FiLM \\ \cite{perez2018film}} & \shortstack{Gating \\ \cite{TimeVLM2502,chowdhury2025t3time}} & \shortstack{\textbf{CFA} \\ (Ours)} \\
\midrule
\multirow{21}{*}{\rotatebox{90}{\textbf{Transformer}}}
 & \multirow{10.5}{*}{\rotatebox{90}{
    \textbf{\shortstack{Nonstationary\\Transformer~\cite{Liu2022NonstationaryTransformer}}}
    }} & Agriculture & \cellcolor{gray!20}    0.084 & \blue{0.089} & \textbf{\red{0.082}} & \textbf{\red{0.079}} & \blue{0.090} & \blue{0.084} & \blue{0.892} & \cellcolor{LightGreen1} \textbf{\red{0.084}} & \cellcolor{LightGreen1} \blue{0.085} & \cellcolor{LightGreen1} \blue{0.085} & \textbf{\red{0.082}} \\
 & & Climate & \cellcolor{gray!20}   1.248 & \blue{1.287} & \textbf{\red{1.229}} & \textbf{\red{1.210}} & \textbf{\red{1.215}} & \textbf{\red{1.230}} & \textbf{\red{1.153}} & \cellcolor{LightGreen1} \blue{1.264} & \cellcolor{LightGreen1} \textbf{\red{1.208}} & \cellcolor{LightGreen1} \textbf{\red{1.232}} & \textbf{\red{1.170}} \\
\bottomrule
\end{tabular}
\end{adjustbox}
\label{tab:horizon_avg}
\vspace{-10pt}
\end{table*}

\textbf{Role of low-rank bottleneck.}
To understand the role of the 
bottleneck, we conduct a toy experiment examining how 
it
affects 
fusion under 1) informative, 2) contradicting, and 3) irrelevant texts. The results show that \textit{the 
low-rank projection 
suppresses misleading textual signals while preserving useful guidance}. Detailed analyses are provided in Appendix~\ref{sec:appendix_toy}, with three complementary analyses:
\setlist[itemize]{leftmargin=0.25cm,itemsep=-1.5pt,topsep=-1.5pt, partopsep=0pt}
\begin{itemize}
    \item \textbf{a) Performance comparison (\ref{sec:appendix_toy_perf}).} We compare forecasting performance w/ and w/o the low-rank bottleneck across matching, contradicting, and irrelevant text. The results show that the low-rank bottleneck consistently improves performance and provides the \textit{largest gain when text is irrelevant}.
    \item \textbf{b) Representation analysis (\ref{sec:appendix_toy_repr}).} We measure the text-contri\-bution ratio at the adapter to quantify how much textual signal survives the bottleneck. Matching text exhibits stronger contribution than contradicting text, indicating that the bottleneck \textit{selectively preserves useful information}.
    \item \textbf{c) TS visualization (\ref{sec:appendix_toy_viz}).} Forecast trajectories show that the 
    bottleneck \textit{prevents misleading predictions under contradicting text} while \textit{preserving accurate forecasts when text is helpful}.
\end{itemize}
A theoretical perspective showing how the low-rank bottleneck constrains the textual signals to a low-dimensional subspace during fusion is provided in Appendix~\ref{app:theory}.

\section{Experiments}
\textbf{Experimental settings.}
To evaluate the generality of the proposed 
method,
we conduct experiments on 9 
multimodal datasets~\cite{liu2024time}, 14 TS backbones, and 4 language models across 4 $H$s, 
as shown in Table~\ref{sec:exp_setting}.
In all experiments, only the TS model is trained while the text model remains frozen. Each setting is evaluated over 10 learning rates, 
reporting the best result.
Following Time-MMD~\cite{liu2024time}, datasets are split into train, validation, and test sets with a ratio of 7:1:2.
Performance is evaluated using MSE and MAE. 

\begin{figure*}[t]
\centering
\begin{subfigure}[t]{0.46\textwidth}
\centering
\begin{adjustbox}{max width=\linewidth}
\includegraphics[width=\textwidth]{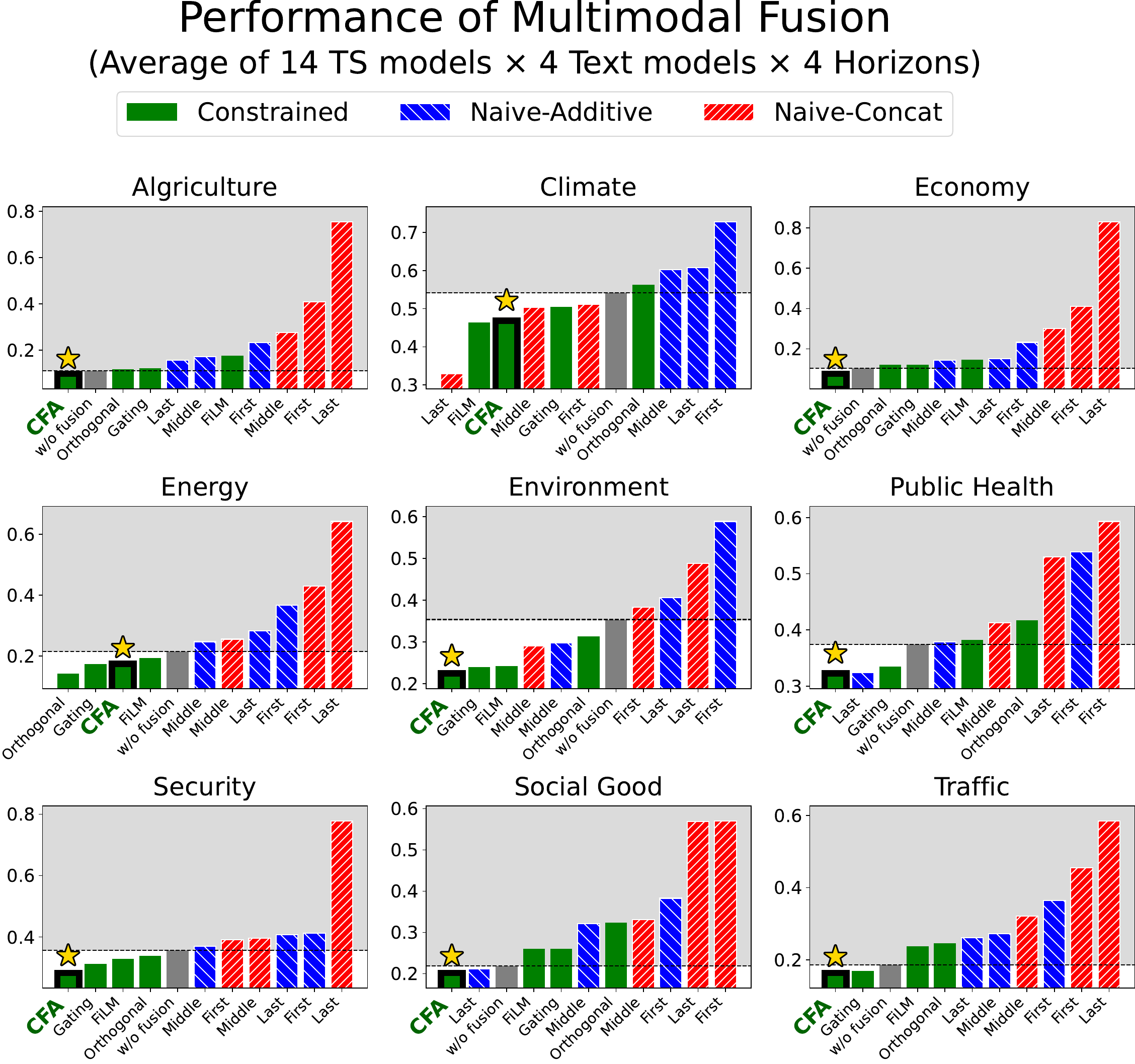}
\end{adjustbox}
\caption{Performance by \textbf{Dataset}.}
\label{fig:setting_dataset}
\end{subfigure}
\hfill
\begin{subfigure}[t]{0.46\textwidth}
\centering
\begin{adjustbox}{max width=\linewidth}
\includegraphics[width=\textwidth]{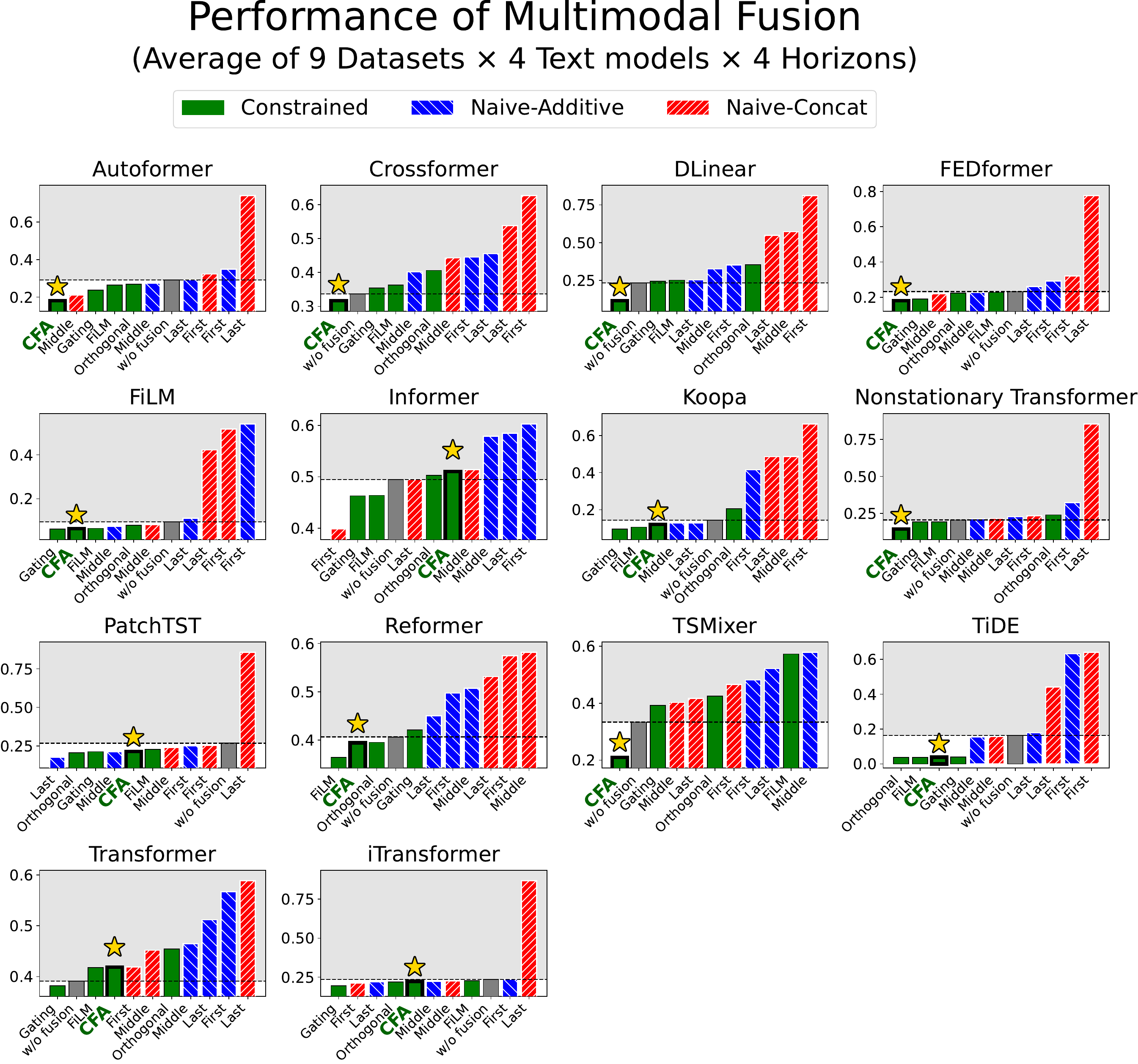}
\end{adjustbox}
\caption{Performance by \textbf{TS models}.}
\label{fig:setting_ts_model}
\end{subfigure}

\medskip
\begin{subfigure}[b]{0.56\textwidth}
\centering
\begin{adjustbox}{max width=\linewidth}
\includegraphics[width=\textwidth]{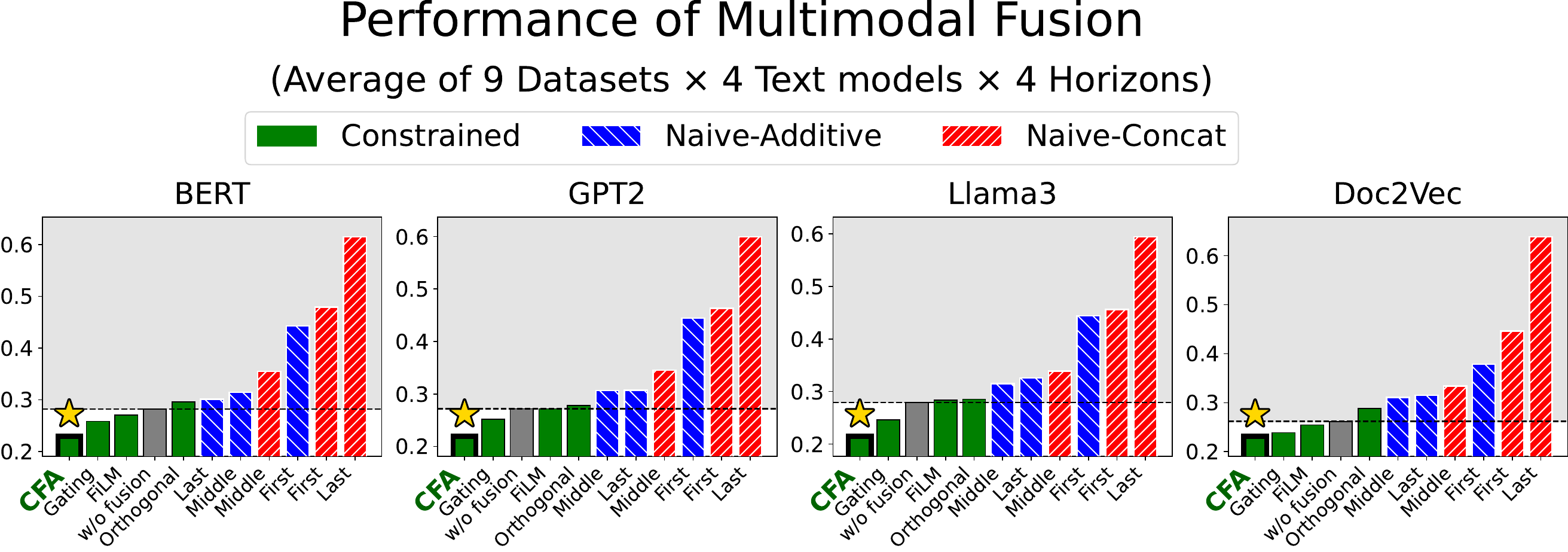}
\end{adjustbox}
\caption{Performance by \textbf{Text models}.}
\label{fig:setting_text_model}
\end{subfigure}
\hfill
\begin{subfigure}[b]{0.395\textwidth}
\centering
\label{fig:motivation_table}
\begin{adjustbox}{max width=\linewidth}
\begin{NiceTabular}{l|c}
\toprule
Method & Normalized MSE \\
\midrule
Concat - F/M/L
& \textcolor{blue}{0.484}/\textcolor{blue}{0.364}/\textcolor{blue}{0.621} \\
Additive - F/M/L
& \textcolor{blue}{0.421}/\textcolor{red}{0.333}/\textcolor{red}{0.334} \\
\cellcolor{LightGray1} w/o fusion & \cellcolor{LightGray1} 0.349 \\
\cellcolor{LightGreen1} Orth./FiLM/Gating
& \cellcolor{LightGreen1}\textcolor{red}{0.317}/\textcolor{red}{0.275}/\textcolor{red}{0.261} \\
\cellcolor{LightYellow} \textbf{CFA} (Ours) & \cellcolor{LightYellow} \textcolor{red}{\textbf{0.256}} \\
\bottomrule
\end{NiceTabular}
\end{adjustbox}
\caption{Average performance.}
\label{tbl:overall_summary}
\end{subfigure}
\caption{
\textbf{Performance across diverse settings.} 
(a), (b), and (c) show the performance (normalized MSE) by \textbf{Dataset}, \textbf{TS model}, and \textbf{Text model}, respectively. 
(d) shows the overall average performance across all settings. 
}
\label{fig:motivation_total}
\vspace{-12pt}
\end{figure*}

\subsection{Multimodal Time Series Forecasting}
Table~\ref{tab:horizon_avg} presents the 
forecasting performance (MSE) of 
multimodal fusion strategies across TS models and datasets
using BERT~\cite{devlin2018bert} as the language model. 
Due to space limitations, we report six TS models~\cite{Liu2022NonstationaryTransformer, Nie2023PatchTST, Zeng2022DLinear, Das2023TiDE, Koopa2023, Zhou2022FiLM}, selecting the two representative methods from each model category 
with the highest performance.
Overall, \textit{naive fusion} methods, whether additive or concatenation-based,
frequently fail to outperform the unimodal baseline and even lead to severe degradation or divergence.
 In contrast, \textit{constrained fusion} methods consistently yield more stable 
and
better performance
across models and domains. Notably, CFA achieves the 
best performance, indicating that 
controlled integration of textual information is crucial for robust multimodal TS forecasting. 
Comparison with other (architecture-specific) methods are shown in Appendix~\ref{app:compare_others}.

\begin{figure*}[t]
\centering
\begin{subfigure}[b]{0.55\textwidth}
\centering
\begin{adjustbox}{width=\linewidth}
\begin{NiceTabular}{l c c c}
\toprule
Text type & w/o bottleneck & w/ bottleneck & Improv. (\%) \\
\midrule
\textbf{Matching}      & 0.1683 & \textbf{0.1477} & $+12.19$ \\
\textbf{Contradicting} & 0.1635 & \textbf{0.1560} & $+4.59$  \\
\textbf{Irrelevant}    & 0.1851 & \textbf{0.1480} & $+20.04$ \\
\bottomrule
\end{NiceTabular}
\end{adjustbox}
\caption{Per-type MSE: w/ vs.\ w/o low-rank bottleneck.}
\label{tab:toy_perf_main}
\end{subfigure}
\hfill
\begin{subfigure}[b]{0.43\textwidth}
\centering
\includegraphics[width=\linewidth]{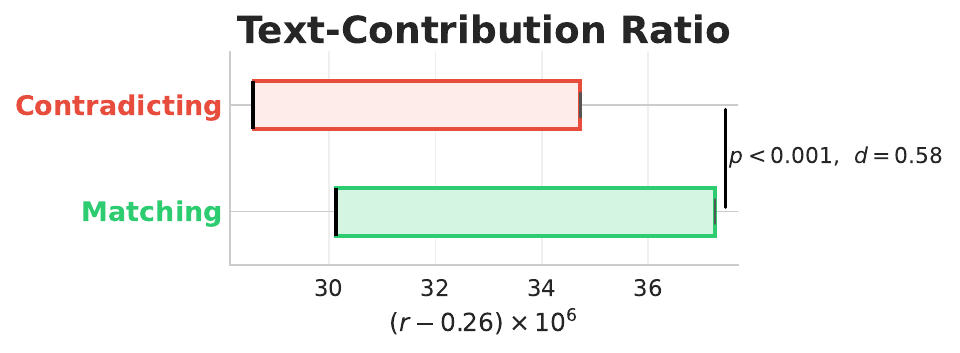}
\caption{Text-contribution ratio by text type.}
\label{fig:toy_repr_main}
\end{subfigure}
\vspace{-5pt}
\caption{
\textbf{Experiment on low-rank bottleneck.}
(a) CFA with a 
bottleneck consistently outperforms the version without a bottleneck across all text types, with the largest gain observed for irrelevant text.
(b) The text-contribution ratio
 shows that matching text is injected more strongly than contradicting text, indicating that it selectively suppresses conflicting signals.
}
\label{fig:toy_analysis}
\vspace{-7pt}
\end{figure*}

\subsection{Performance with Various Settings}
\label{sec:various_settings}
In this section,
we evaluate the proposed method across diverse 
\textbf{1) datasets}, 
\textbf{2) TS backbones}, and 
\textbf{3) text encoders} to verify its general effectiveness. 
To account for scale differences across datasets,
we report normalized MSE averaged over 10 fusion strategies.

\textbf{[1] Various datasets (Figure~\ref{fig:setting_dataset}).}
We conduct experiments on 9 real-world multimodal datasets~\cite{liu2024time} for TS forecasting. Across datasets, constrained fusion strategies (\textcolor{darkgreen1}{green}) generally outperform the additive (\textcolor{blue}{blue}) and concatenation-based (\textcolor{red}{red}) fusion. 
While many fusion methods exhibit dataset-dependent behavior,
CFA consistently outperforms the unimodal model on all datasets
and ranks first on 7 of the 9 datasets.

\textbf{[2] Various TS models (Figure~\ref{fig:setting_ts_model}).}
We 
evaluate the effectiveness of our method across 
diverse TS backbones (e.g., Transformer-based, linear/MLP-based). Similar to the dataset-level analysis, constrained fusion strategies (\textcolor{darkgreen1}{green}) generally outperform the additive (\textcolor{blue}{blue}) and concatenation-based (\textcolor{red}{red}) approaches. 
Specifically, CFA improves over the unimodal baseline on 13 out of 14 backbones, with the exception of Transformer~\cite{vaswani2017attention}.
We attribute this to the fact that the standard Transformer exhibits substantially lower performance than other TS backbones, indicating a limitation of the backbone itself rather than the fusion strategy. 
For Transformer variants specifically designed for 
TS forecasting~\cite{Liu2022NonstationaryTransformer, Liu2023iTransformer, zhang2023crossformer}, 
CFA consistently yields improvements over unimodal models.

\textbf{[3] Various text models (Figure~\ref{fig:setting_text_model}).}
To verify robustness to text encoders, we compare four text models, including three LLM-based encoders~\cite{devlin2018bert,radford2019gpt2,dubey2024llama3} and Doc2Vec~\cite{le2014doc2vec} to account for scenarios where LLM deployment is restricted. Consistent with previous observations, constrained fusion strategies (\textcolor{darkgreen1}{green}) outperform additive (\textcolor{blue}{blue}) and concatenation-based (\textcolor{red}{red}) methods.
Among them, CFA 
consistently 
achieves the 
best performance,
demonstrating robustness to the 
text model.

\textbf{[4] Overall comparison (Table~\ref{tbl:overall_summary}).}
Table~\ref{tbl:overall_summary}
reports the average normalized MSE across more than 2K settings (9 datasets × 14 TS models × 4 text models × 4 horizons)
with 10 fusion methods. Among simple fusion strategies, only additive fusion at the middle or last layer improves over the unimodal baseline, whereas other naive approaches fail to do so. This 
highlights the necessity of constrained fusion for consistent performance gains. Furthermore, CFA achieves the 
best performance,
confirming its superiority over 
various fusion strategies.
Note that we tune each setting over 10 learning rates and compare the best performance to ensure that  
gains or losses are not due to optimization effects.

\section{Analysis}
In this section, we analyze \textit{how and why the proposed method operates effectively},
using BERT~\cite{devlin2018bert} as the text model:
\setlist[itemize]{leftmargin=0.3cm,itemsep=-1pt,topsep=-1pt, partopsep=0pt}
\begin{itemize}
\item \textbf{1) Performance with irrelevant text.}
We inject \textit{mismatched textual inputs} from unrelated datasets and observe that CFA exhibits the smallest 
degradation relative to the unimodal baseline, 
demonstrating robustness to irrelevant information.
\item \textbf{2) 
Effect of low-rank bottleneck.}
Using a synthetic dataset with matched, contradicting, and irrelevant text, we show that the bottleneck reduces MSE across text types. The largest gain occurs for irrelevant text, and matching text is injected more strongly than contradicting text.

\item \textbf{3) Representation similarity.}
We compute \textit{cosine similarity between TS-only and TS+Text representations} and observe that the best performance does not correspond to the largest representation shift,
highlighting 
constrained modification.

\item \textbf{4) Visualization of TS forecasting.}
We visualize the predicted values in TS forecasting and observe that CFA captures late-stage trends missed by the unimodal model.

\item \textbf{5) Temporal attribution analysis.}
We analyze temporal attribution to quantify \textit{how each input 
step contributes to the prediction} and find that CFA yields 
a distinct temporal importance distribution, 
indicating that it selectively references 
time steps.

\item \textbf{6) Efficiency analysis.}
We compare the number of parameters and FLOPs across fusion methods and observe that CFA introduces 
marginal overhead relative to the unimodal baseline, 
while several strategies significantly increase computational overhead.

\item 
\textbf{7) Information analysis.}
We compute the rank correlation between MAE and effective rank across diverse settings and 
find that methods with higher effective rank tend to achieve lower MAE,
suggesting that models with more distributed representations tend to forecast more accurately.
\end{itemize}

\setlength{\columnsep}{6pt} 
\begin{figure}[t]
  \centering
  \begin{adjustbox}{max width=\linewidth}
\includegraphics[width=0.43\textwidth]{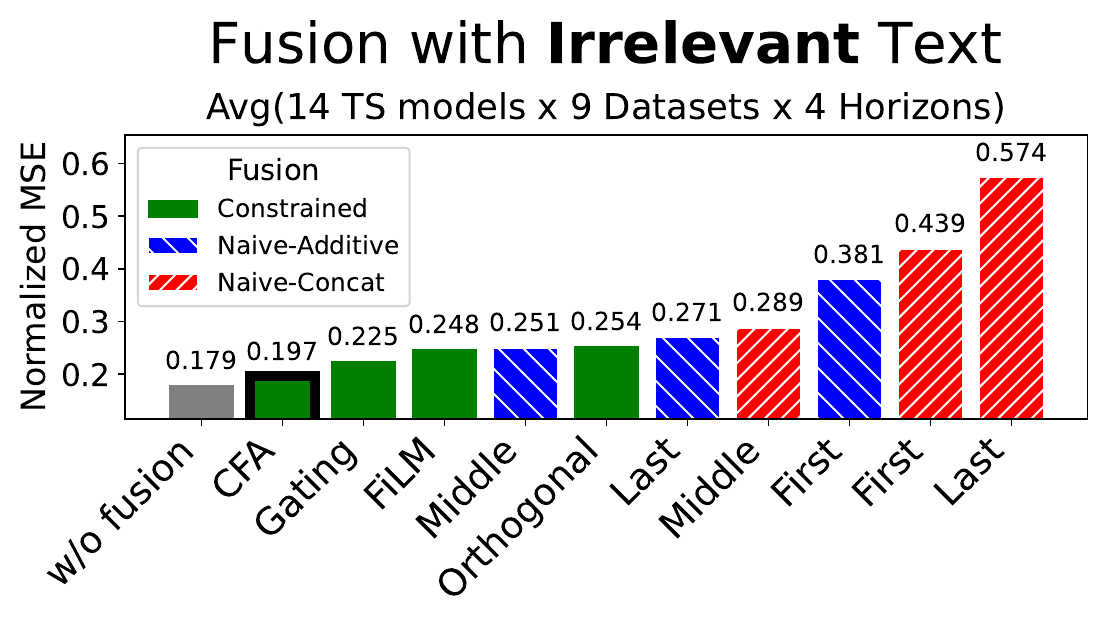}
\end{adjustbox}
\vspace{-10pt}
  \caption{Irrelevant text experiments.}
  \label{fig:irrelevant_text}
\vspace{-13pt}
\end{figure}

\textbf{1) Performance with irrelevant text (Figure~\ref{fig:irrelevant_text}).}
To assess the effect of multimodal fusion, we conduct an \textit{irrelevant text} experiment,
where we replace the text aligned with each TS by text sampled from entirely different datasets. 
For instance, when evaluating on the Agriculture dataset, we randomly sample text from the remaining eight datasets.
This design examines how each fusion strategy responds when irrelevant textual information is injected. 

A robust fusion strategy is expected to ignore unrelated information and rely primarily on TS representations, such that its performance does not substantially degrade compared to the unimodal setting.
As shown in Figure~\ref{fig:irrelevant_text}, 
CFA exhibits performance most similar to the unimodal model and shows the smallest degradation.
In addition, constrained fusion strategies except Orthogonal effectively filter out unnecessary information and primarily utilize TS representations.
These results confirm that CFA incorporates relevant textual signals while remaining robust to irrelevant inputs.

\begin{figure*}[t]
\centering
\begin{adjustbox}{max width=\linewidth}
\includegraphics[width=.85\textwidth]{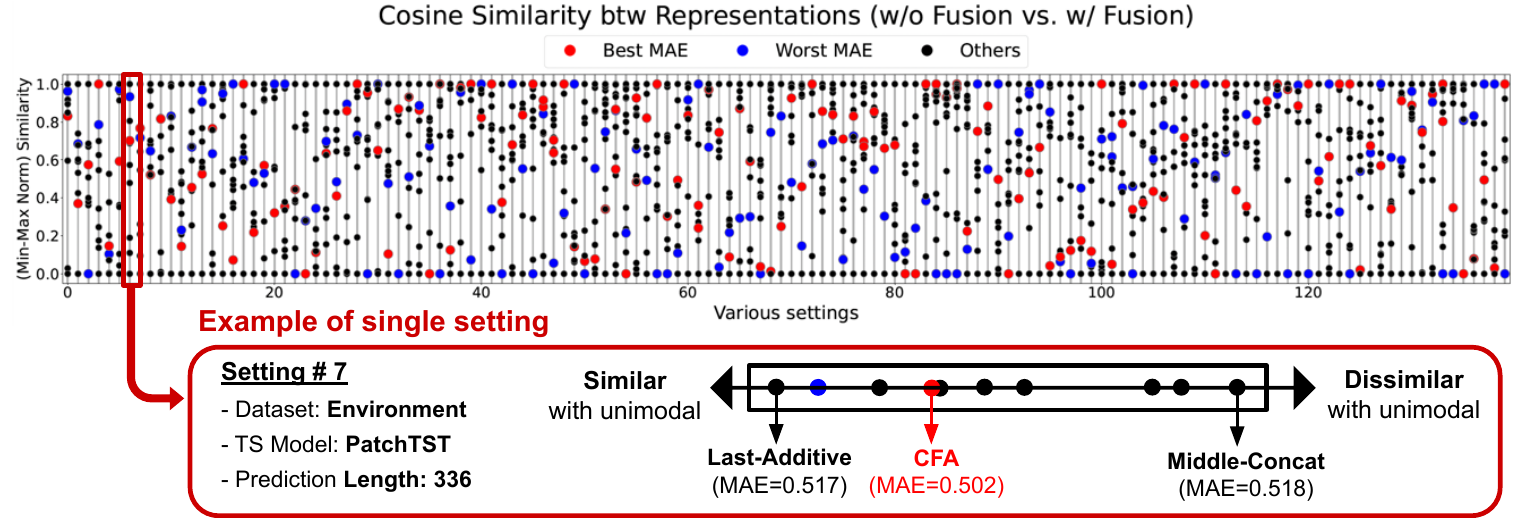}
\end{adjustbox} 
\vspace{-4pt}
\caption{
\textbf{Cosine similarity between representations of TS-only and TS+Text models.}
We apply multiple fusion strategies across diverse settings, where each dot represents a single fusion method.
The best-performing method (\textcolor{red}{red}) does \textit{not} consistently induce the largest representation shift from the unimodal baseline, 
highlighting the importance of controlled integration of textual information into temporal representations rather than simply increasing the magnitude of representation shift.
}
\label{fig:analysis2}
\end{figure*}

\begin{figure*}[t]
\centering
\begin{adjustbox}{max width=\linewidth}
\includegraphics[width=.85\textwidth]{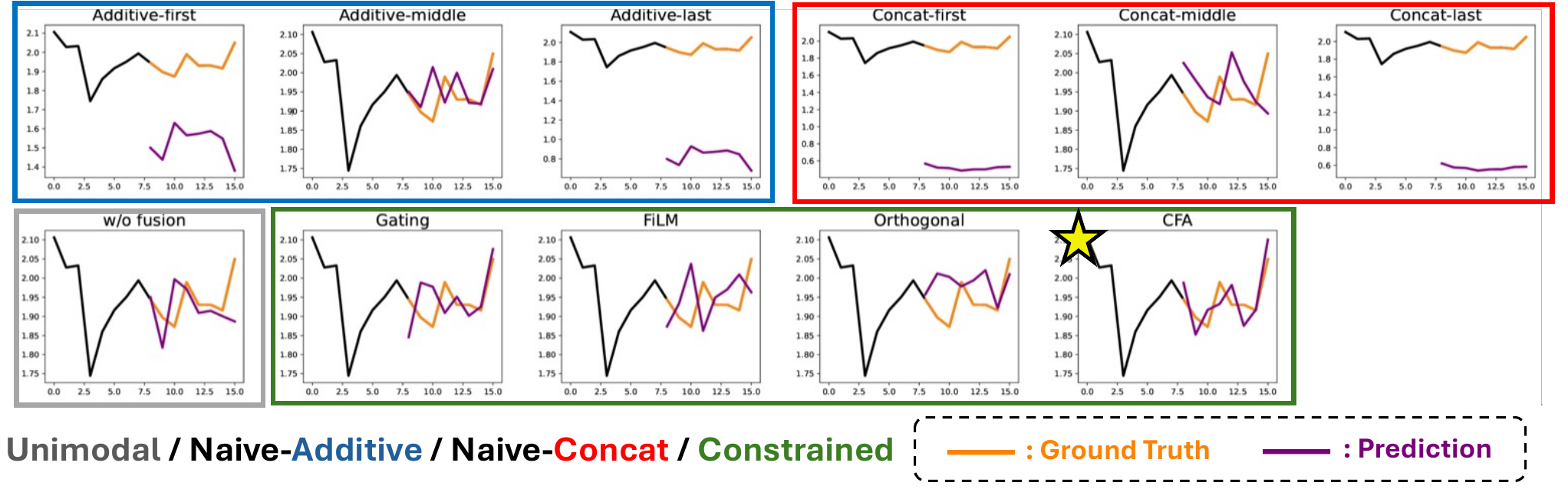}
\end{adjustbox}
\vspace{-8pt}
\caption{
\textbf{4) Visualization of multimodal TS forecasting.} 
Although the unimodal model captures the initial pattern, it fails to model the subsequent upward trend, whereas CFA accurately captures the rise in the later horizon.
In addition, some fusion strategies even 
fail to converge (e.g., concat-first),
indicating that naive fusion can hinder effective learning.
}
\label{fig:analysis3}
\vspace{-10pt}
\end{figure*}

\textbf{2) Effect of low-rank bottleneck (Figure~\ref{fig:toy_analysis}).}
To 
analyze the filtering behavior of CFA's low-rank adapter, we conduct a controlled toy experiment on a synthetic dataset containing three text types: \textbf{matching} (reflecting the TS trend), \textbf{contradicting} (opposing the TS trend), and \textbf{irrelevant} (topically unrelated).
We compare CFA w/ and w/o the bottleneck, 
with details in Appendix~\ref{sec:appendix_toy}.

Table~\ref{tab:toy_perf_main} shows that the bottleneck consistently reduces MSE across all text types.
The largest improvement occurs for irrelevant text, indicating that \textit{the bottleneck suppresses uninformative texts}.
Without the bottleneck, contradicting text yields \textit{lower} MSE than matching text, 
suggesting that all text is treated as undifferentiated noise.
With the bottleneck, the expected ordering appears, where matching text yields the lowest MSE, followed by irrelevant text and contradicting text.

Figure~\ref{fig:toy_repr_main} reports the text-contribution ratio (See Appendix~\ref{sec:appendix_toy_repr})
measured at the adapter output.
Matching text produces a higher mean ratio than contradicting text with strong statistical significance ($p{<}0.001$, Cohen’s $d{=}0.58$).
This indicates that CFA \textit{injects helpful text more strongly while suppressing conflicting signals}.
Visualization of forecast trajectories are illustrated in Appendix~\ref{sec:appendix_toy_viz}.

\textbf{3) Representation similarity (Figure~\ref{fig:analysis2}).}
To quantify how text fusion reshapes temporal representations, 
we measure the cosine similarity between the layer-wise representations 
$\mathbf{Z}^{\text{TS}}_{\ell}$ and 
$\mathbf{Z}^{\text{TS+Text}}_{\ell}$ at layer $\ell$, 
where $\ell \in \{1, \dots, L\}$ denotes the layer index.
We compute the layer-wise cosine similarity between 
$\mathbf{Z}^{\text{TS}}_{\ell}$ and 
$\mathbf{Z}^{\text{TS+Text}}_{\ell}$, 
and obtain the final similarity score by averaging across layers:

\begin{equation}
S = \frac{1}{L} \sum_{\ell=1}^{L}
\frac{
\left\langle \mathbf{Z}^{\text{TS}}_{\ell}, \mathbf{Z}^{\text{TS+Text}}_{\ell} \right\rangle
}{
\left\| \mathbf{Z}^{\text{TS}}_{\ell} \right\|_2
\left\| \mathbf{Z}^{\text{TS+Text}}_{\ell} \right\|_2
}.
\end{equation}
We repeat this analysis across different settings and fusion strategies.
As shown in Figure~\ref{fig:analysis2}, best-performing fusion method (\textcolor{red}{red}) does not consistently correspond to the largest deviation from the unimodal baseline.
This indicates that effective fusion 
requires \textit{controlled modification} of temporal representations with textual information rather than simply increasing representation shift.

\begin{figure*}[t]
\centering
\begin{subfigure}[b]{0.425\textwidth}
\centering
\includegraphics[width=\linewidth]{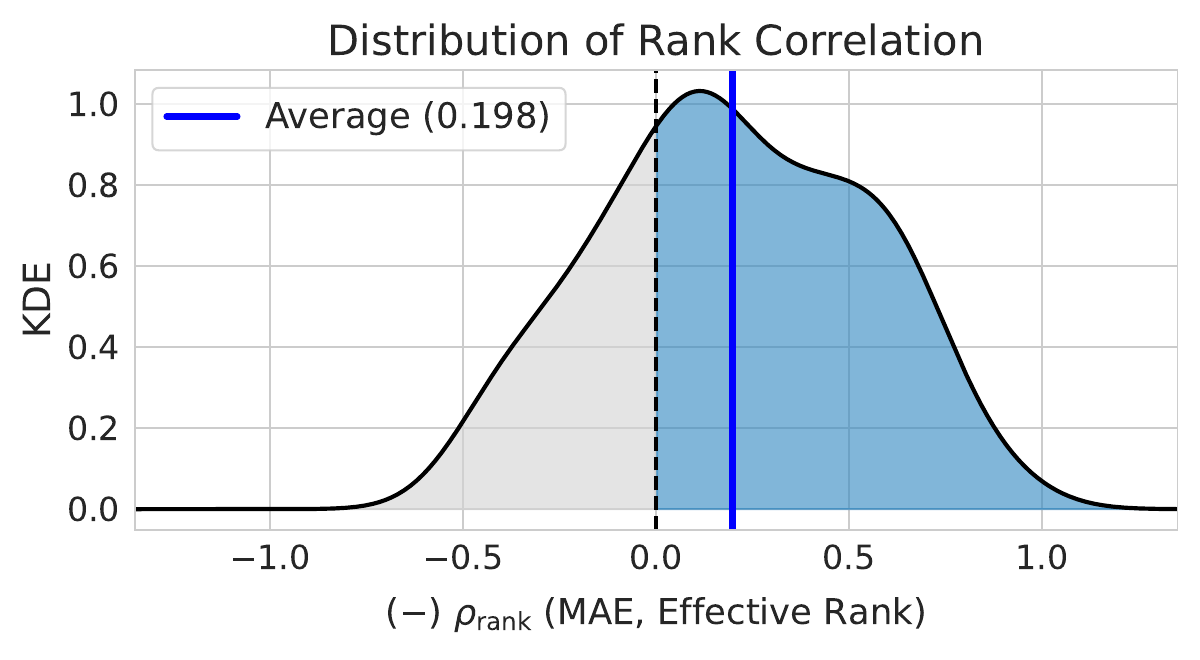}
\caption{Distribution of rank correlation.}
\label{fig:analysis1_1}
\end{subfigure}
\hspace{5pt}
\begin{subfigure}[b]{0.46\textwidth}
\centering
\begin{adjustbox}{width=\linewidth}
\begin{NiceTabular}{l|cc}
\toprule
Method & MAE $\downarrow$  & Effective Rank $\uparrow$ \\
\midrule
\cellcolor{LightYellow} \textbf{CFA} (Ours) & \cellcolor{LightYellow} 0.0929 (1/11) & \cellcolor{LightYellow} 24.05 (1/11) \\
\cellcolor{LightGreen1} Orthogonal & \cellcolor{LightGreen1} 0.0967 (4/11) & \cellcolor{LightGreen1} 19.37 (3/11) \\
\cellcolor{LightGray1} w/o fusion & \cellcolor{LightGray1} 0.0970 (5/11) & \cellcolor{LightGray1} 18.97 (4/11) \\
Additive (First) & 0.0983 (7/11) & 16.79 (9/11) \\ 
Concat (Middle) & 0.0996 (9/11) & 16.10 (10/11) \\ 
\midrule
Rank correlation ($\rho$) & \multicolumn{2}{c}{0.6727 (p-value: 0.023)} \\
\bottomrule
\end{NiceTabular}
\end{adjustbox}
\caption{Example of rank correlation.}
\label{fig:analysis1_2}
\end{subfigure}
\caption{
\textbf{Rank correlation(MAE, Effective rank).}
(a) Distn of rank correlations across various settings, showing that higher effective rank tends to associate with lower MAE.
(b) Methods with higher effective rank generally achieve lower MAE
(w/ positive rank correlation of $\rho = 0.6727$).
}
\label{fig:analysis1}
\end{figure*}

\textbf{4) Visualization of TS forecasting (Figure~\ref{fig:analysis3}).}
To assess how textual fusion influences temporal prediction, we visualize forecasting results on the Agriculture dataset~\cite{liu2024time} using TiDE~\cite{Das2023TiDE},
with both input and output horizons set to 8. 
While the unimodal model captures the initial pattern yet fails to model the subsequent upward trend, CFA accurately captures the rise in the later horizon. 
Moreover, most constrained fusion (\textcolor{darkgreen1}{green}) methods successfully capture the late-stage upward trend that the unimodal model fails to capture.
In contrast, several naive fusion strategies (\textcolor{red}{red}, \textcolor{blue}{blue}) even 
underperform the unimodal baseline, 
indicating that improper fusion can hinder effective learning. 

\setlength{\columnsep}{6pt} 
\begin{figure*}[t]
\centering
\begin{minipage}[b]{0.285\textwidth}
  \centering
  \begin{adjustbox}{max width=\linewidth}
    \includegraphics[width=1.0\textwidth]{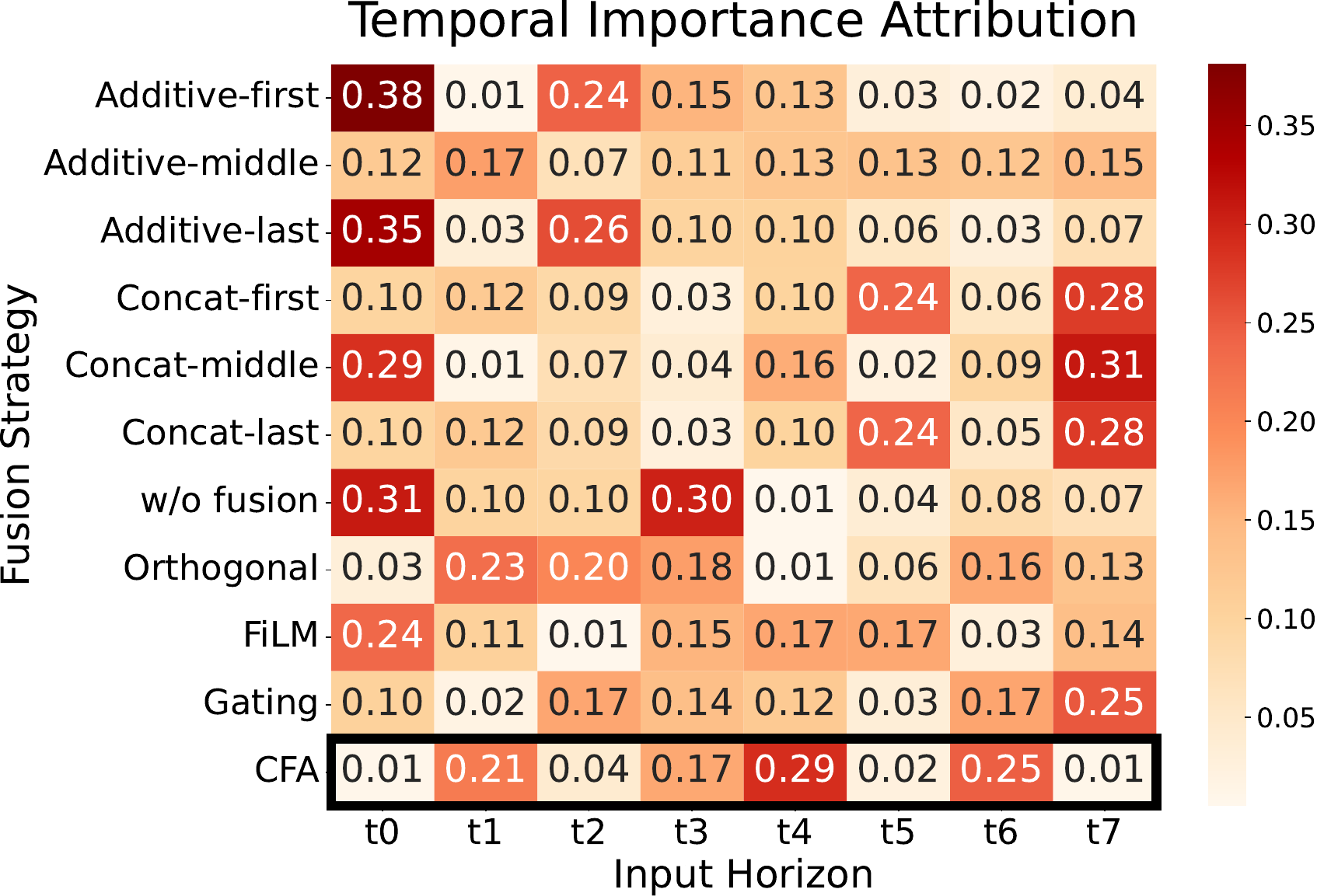}
  \end{adjustbox}
  \captionof{figure}{
  \textbf{Temporal importance.} CFA exhibits a distinct importance distribution over time steps.
  }
  \label{fig:analysis3_2}
\end{minipage}
\hfill
\begin{minipage}[b]{0.69\textwidth}
  \vspace{-6pt}
  \centering
  \begin{adjustbox}{max width=\linewidth}
    \begin{NiceTabular}{c|c|ccc|ccc|ccc>{\columncolor{LightYellow}}c}
    \toprule
    \multirow{4}{*}{Average }
     &
     \multirow{4}{*}{\shortstack{\textbf{Unimodal}\\\textbf{(w/o text)}}}
     & \multicolumn{6}{c}{\textbf{Naive}}
     & \multicolumn{4}{c}{\multirow{2.5}{*}{\textbf{Constrained}}}
     \\
     \cmidrule(lr){3-8}
      &  & \multicolumn{3}{c}{\textbf{Additive}}
      & \multicolumn{3}{c}{\textbf{Concat}}
      &
     \\
    \cmidrule(lr){3-5}\cmidrule(lr){6-8}\cmidrule(lr){9-12}
     & & First & Middle & Last
     & First & Middle & Last
     & Orthogonal & FiLM & Gating &
    \multicolumn{1}{c}{\cellcolor{white}\textbf{CFA}}
    \\
    \midrule
    \multirow{2}{*}{
    \shortstack[c]{\textbf{Params.}\\ (1e+06)}
    }
     & \multirow{2}{*}{9.382} &  9.384 & 9.408 & 9.384 & 9.384 & 10.539 & 9.384 & \cellcolor{LightGreen1} 9.403 & \cellcolor{LightGreen1} 9.423 & \cellcolor{LightGreen1} 12.067 & \textbf{9.439} \\
     &  &
     + 0.02\% &  + 0.28\% &  + 0.02\% &
     + 0.02\% &  + 12.33\% &  + 0.02\% &
     \cellcolor{LightGreen1} + 0.23\% & \cellcolor{LightGreen1} + 0.46\% & \cellcolor{LightGreen1} + 28.62\% & \textbf{+ 0.61\%} \\
     \midrule
    \multirow{2}{*}{
    \shortstack[c]{\textbf{FLOPs}\\ (1e+08)}
    }
    &
    \multirow{2}{*}{1.436} & 1.436 & 1.436 & 1.436 & 1.436 & 1.616 & 1.436 & \cellcolor{LightGreen1} 1.436 & \cellcolor{LightGreen1} 1.436 & \cellcolor{LightGreen1} 1.885 & \textbf{1.436}  \\
    &  &
     + 0.00\% &  + 0.02\% &  + 0.00\% &
     + 0.00\% &  + 12.53\% &  + 0.00\% &
     \cellcolor{LightGreen1} + 0.02\% & \cellcolor{LightGreen1} + 0.03\% & \cellcolor{LightGreen1} + 31.26\% & \textbf{+ 0.04\%} \\
    \bottomrule
    \end{NiceTabular}
  \end{adjustbox}
  \vspace{6pt}
  \captionof{table}{\textbf{Efficiency analysis of fusion strategies.} The table shows the average 1) training parameters and 2) FLOPs, 
  where CFA achieves negligible overhead compared to the unimodal baseline, maintaining comparable efficiency.
  }
  \label{tab:efficiency_analysis}
\end{minipage}
\end{figure*}

\textbf{5) Temporal attribution analysis (Figure~\ref{fig:analysis3_2}).}
Under the same experimental setting as \textbf{Analysis 4)}, we further analyze the importance scores that quantify \textit{how much each input TS step contributes to the prediction}. 
Specifically, we compute a gradient $\times$ input attribution with respect to the encoder input and aggregate over channels to obtain one importance value per time step.
As shown in Figure~\ref{fig:analysis3_2}, CFA assigns a different importance distribution over input TS steps compared to w/o fusion and other methods, indicating that textual information influences which temporal regions are referenced during prediction.
See Appendix~\ref{app:temporal_attribution} for details on the computation of temporal attribution.

\textbf{6) Efficiency analysis (Table~\ref{tab:efficiency_analysis}).}
To demonstrate the efficiency of CFA, we compare the average number of training parameters and FLOPs across 10 fusion methods, aggregated over 14 TS backbones and 9 datasets under prediction lengths of 8, 36, and 96 for monthly, weekly, and daily frequencies.
While certain strategies 
substantially increase computational overhead,
most designs introduce 
marginal overhead. 
Notably, CFA increases parameters by only 0.61\% and FLOPs by 0.04\% relative to the unimodal model, 
indicating that it maintains \textit{efficiency comparable to the unimodal setting}. 


\textbf{7) Information analysis (Figure~\ref{fig:analysis1}).}
We analyze 
the fusion 
using the \textit{effective rank} of layer-wise representations. 
Given a hidden representation $\mathbf{H}$, we compute its singular values $\{\sigma_i\}_{i=1}^{r}$ and define $p_i = \frac{\sigma_i}{\sum_{j=1}^{r} \sigma_j}$ and 
$\mathrm{erank}(\mathbf{H}) = \exp \left( - \sum_{i=1}^{r} p_i \log p_i \right)$,
where 
higher 
rank indicates more distributed representations. 
To assess its relevance, we compute the rank correlation between MAE and effective rank across fusion strategies under diverse settings (four TS backbones~\cite{Liu2022NonstationaryTransformer, Nie2023PatchTST, Liu2023iTransformer, Zhou2022FiLM}, nine datasets, four $H$s). 
Figure~\ref{fig:analysis1_1} shows that methods with higher effective rank tend to achieve lower MAE, as confirmed by a positive Spearman rank correlation,
with example shown in Table~\ref{fig:analysis1_2}.

\section{Conclusion}
In this paper, we show that naive fusion often underperforms unimodal baselines, suggesting that indiscriminate cross-modal fusion fails to preserve temporal representations.
We demonstrate that constrained fusion consistently improves performance by filtering irrelevant auxiliary signals while preserving temporal representations effectively.
Furthermore, we propose CFA, a constrained fusion method 
that suppresses irrelevant textual information in the low-rank subspace.
Extensive experiments across various 
datasets and backbones 
validate the effectiveness of our approach.

\textbf{Limitations and Future Work.}
As our study focuses on textual modalities, extending constrained fusion to other modalities (e.g. vision and tabular) remains a promising direction for future research.
In addition, although CFA shows consistent improvements across diverse settings, our analysis remains largely empirical, and deeper theoretical understanding would strengthen the framework.

\FloatBarrier   

\setlength{\emergencystretch}{3em}
\setlength{\bibsep}{9pt plus 2pt minus 1pt}
\bibliographystyle{ACM-Reference-Format}
\bibliography{references}


\begin{thebibliography}{47}


\ifx \showCODEN    \undefined \def \showCODEN     #1{\unskip}     \fi
\ifx \showISBNx    \undefined \def \showISBNx     #1{\unskip}     \fi
\ifx \showISBNxiii \undefined \def \showISBNxiii  #1{\unskip}     \fi
\ifx \showISSN     \undefined \def \showISSN      #1{\unskip}     \fi
\ifx \showLCCN     \undefined \def \showLCCN      #1{\unskip}     \fi
\ifx \shownote     \undefined \def \shownote      #1{#1}          \fi
\ifx \showarticletitle \undefined \def \showarticletitle #1{#1}   \fi
\ifx \showURL      \undefined \def \showURL       {\relax}        \fi
\providecommand\bibfield[2]{#2}
\providecommand\bibinfo[2]{#2}
\providecommand\natexlab[1]{#1}
\providecommand\showeprint[2][]{arXiv:#2}

\bibitem[Angryk et~al\mbox{.}(2020)]%
        {angryk2020multivariate}
\bibfield{author}{\bibinfo{person}{Rafal~A Angryk}, \bibinfo{person}{Petrus~C
  Martens}, \bibinfo{person}{Berkay Aydin}, \bibinfo{person}{Dustin Kempton},
  \bibinfo{person}{Sushant~S Mahajan}, \bibinfo{person}{Sunitha Basodi},
  \bibinfo{person}{Azim Ahmadzadeh}, \bibinfo{person}{Xumin Cai},
  \bibinfo{person}{Soukaina Filali~Boubrahimi}, \bibinfo{person}{Shah~Muhammad
  Hamdi}, {et~al\mbox{.}}} \bibinfo{year}{2020}\natexlab{}.
\newblock \showarticletitle{Multivariate time series dataset for space weather
  data analytics}.
\newblock \bibinfo{journal}{\emph{Scientific data}} \bibinfo{volume}{7},
  \bibinfo{number}{1} (\bibinfo{year}{2020}), \bibinfo{pages}{227}.
\newblock


\bibitem[Chattopadhyay et~al\mbox{.}(2024)]%
        {chattopadhyay2024context}
\bibfield{author}{\bibinfo{person}{Sameep Chattopadhyay},
  \bibinfo{person}{Pulkit Paliwal}, \bibinfo{person}{Sai~Shankar Narasimhan},
  \bibinfo{person}{Shubhankar Agarwal}, {and} \bibinfo{person}{Sandeep~P
  Chinchali}.} \bibinfo{year}{2024}\natexlab{}.
\newblock \showarticletitle{Context matters: Leveraging contextual features for
  time series forecasting}.
\newblock \bibinfo{journal}{\emph{arXiv preprint arXiv:2410.12672}}
  (\bibinfo{year}{2024}).
\newblock


\bibitem[Chen et~al\mbox{.}(2023)]%
        {Chen2023TSMixer}
\bibfield{author}{\bibinfo{person}{Si-An Chen}, \bibinfo{person}{Chun-Liang
  Li}, \bibinfo{person}{Nate Yoder}, \bibinfo{person}{Sercan~O Arik}, {and}
  \bibinfo{person}{Tomas Pfister}.} \bibinfo{year}{2023}\natexlab{}.
\newblock \showarticletitle{Tsmixer: An all-mlp architecture for time series
  forecasting}.
\newblock \bibinfo{journal}{\emph{TMLR}} (\bibinfo{year}{2023}).
\newblock


\bibitem[Chowdhury et~al\mbox{.}(2025)]%
        {chowdhury2025t3time}
\bibfield{author}{\bibinfo{person}{Abdul~Monaf Chowdhury},
  \bibinfo{person}{Rabeya Akter}, {and} \bibinfo{person}{Safaeid~Hossain
  Arib}.} \bibinfo{year}{2025}\natexlab{}.
\newblock \showarticletitle{T3time: Tri-modal time series forecasting via
  adaptive multi-head alignment and residual fusion}.
\newblock \bibinfo{journal}{\emph{arXiv preprint arXiv:2508.04251}}
  (\bibinfo{year}{2025}).
\newblock


\bibitem[Cirstea et~al\mbox{.}(2022)]%
        {cirstea2022towards}
\bibfield{author}{\bibinfo{person}{Razvan-Gabriel Cirstea},
  \bibinfo{person}{Bin Yang}, \bibinfo{person}{Chenjuan Guo},
  \bibinfo{person}{Tung Kieu}, {and} \bibinfo{person}{Shirui Pan}.}
  \bibinfo{year}{2022}\natexlab{}.
\newblock \showarticletitle{Towards spatio-temporal aware traffic time series
  forecasting}. In \bibinfo{booktitle}{\emph{2022 IEEE 38th International
  Conference on Data Engineering (ICDE)}}. IEEE, \bibinfo{pages}{2900--2913}.
\newblock


\bibitem[Das et~al\mbox{.}(2023)]%
        {Das2023TiDE}
\bibfield{author}{\bibinfo{person}{A. Das}, \bibinfo{person}{W. Kong},
  \bibinfo{person}{A.~B. Leach}, \bibinfo{person}{S. Mathur},
  \bibinfo{person}{R. Sen}, {and} \bibinfo{person}{R. Yu}.}
  \bibinfo{year}{2023}\natexlab{}.
\newblock \showarticletitle{Long-Term Forecasting with TiDE: Time-Series Dense
  Encoder}.
\newblock \bibinfo{journal}{\emph{arXiv Preprint arXiv:2304.08424}}
  (\bibinfo{year}{2023}).
\newblock


\bibitem[Devlin et~al\mbox{.}(2019)]%
        {devlin2018bert}
\bibfield{author}{\bibinfo{person}{Jacob Devlin}, \bibinfo{person}{Ming-Wei
  Chang}, \bibinfo{person}{Kenton Lee}, {and} \bibinfo{person}{Kristina
  Toutanova}.} \bibinfo{year}{2019}\natexlab{}.
\newblock \showarticletitle{BERT: Pre-training of Deep Bidirectional
  Transformers for Language Understanding}. In
  \bibinfo{booktitle}{\emph{NAACL}}.
\newblock


\bibitem[Dubey et~al\mbox{.}(2024)]%
        {dubey2024llama3}
\bibfield{author}{\bibinfo{person}{Abhimanyu Dubey}, \bibinfo{person}{Abhinav
  Jauhri}, \bibinfo{person}{Abhinav Pandey}, \bibinfo{person}{Abhishek Kadian},
  \bibinfo{person}{Ahmad Al-Dahle}, \bibinfo{person}{Aiesha Letman},
  \bibinfo{person}{Akhil Mathur}, \bibinfo{person}{Alan Schelten},
  \bibinfo{person}{Amy Yang}, \bibinfo{person}{Angela Fan}, {et~al\mbox{.}}}
  \bibinfo{year}{2024}\natexlab{}.
\newblock \showarticletitle{The {Llama} 3 Herd of Models}.
\newblock \bibinfo{journal}{\emph{arXiv preprint arXiv:2407.21783}}
  (\bibinfo{year}{2024}).
\newblock


\bibitem[Ekambaram et~al\mbox{.}(2020)]%
        {ekambaram2020attention}
\bibfield{author}{\bibinfo{person}{Vijay Ekambaram}, \bibinfo{person}{Kushagra
  Manglik}, \bibinfo{person}{Sumanta Mukherjee}, \bibinfo{person}{Surya
  Shravan~Kumar Sajja}, \bibinfo{person}{Satyam Dwivedi}, {and}
  \bibinfo{person}{Vikas Raykar}.} \bibinfo{year}{2020}\natexlab{}.
\newblock \showarticletitle{Attention based multi-modal new product sales
  time-series forecasting}. In \bibinfo{booktitle}{\emph{Proceedings of the
  26th ACM SIGKDD international conference on knowledge discovery \& data
  mining}}. \bibinfo{pages}{3110--3118}.
\newblock


\bibitem[Hu et~al\mbox{.}(2022)]%
        {hu2022lora}
\bibfield{author}{\bibinfo{person}{Edward~J Hu}, \bibinfo{person}{Yelong Shen},
  \bibinfo{person}{Phillip Wallis}, \bibinfo{person}{Zeyuan Allen-Zhu},
  \bibinfo{person}{Yuanzhi Li}, \bibinfo{person}{Shean Wang},
  \bibinfo{person}{Lu Wang}, {and} \bibinfo{person}{Weizhu Chen}.}
  \bibinfo{year}{2022}\natexlab{}.
\newblock \showarticletitle{Lora: Low-rank adaptation of large language
  models.}
\newblock \bibinfo{journal}{\emph{ICLR}} \bibinfo{volume}{1},
  \bibinfo{number}{2} (\bibinfo{year}{2022}), \bibinfo{pages}{3}.
\newblock


\bibitem[Jia et~al\mbox{.}(2024)]%
        {jia2024gpt4mts}
\bibfield{author}{\bibinfo{person}{Furong Jia}, \bibinfo{person}{Kevin Wang},
  \bibinfo{person}{Yixiang Zheng}, \bibinfo{person}{Defu Cao}, {and}
  \bibinfo{person}{Yan Liu}.} \bibinfo{year}{2024}\natexlab{}.
\newblock \showarticletitle{Gpt4mts: Prompt-based large language model for
  multimodal time-series forecasting}. In \bibinfo{booktitle}{\emph{AAAI}}.
\newblock


\bibitem[Jiang et~al\mbox{.}(2025)]%
        {jiang2025multi}
\bibfield{author}{\bibinfo{person}{Yushan Jiang}, \bibinfo{person}{Kanghui
  Ning}, \bibinfo{person}{Zijie Pan}, \bibinfo{person}{Xuyang Shen},
  \bibinfo{person}{Jingchao Ni}, \bibinfo{person}{Wenchao Yu},
  \bibinfo{person}{Anderson Schneider}, \bibinfo{person}{Haifeng Chen},
  \bibinfo{person}{Yuriy Nevmyvaka}, {and} \bibinfo{person}{Dongjin Song}.}
  \bibinfo{year}{2025}\natexlab{}.
\newblock \showarticletitle{Multi-modal time series analysis: A tutorial and
  survey}. In \bibinfo{booktitle}{\emph{Proceedings of the 31st ACM SIGKDD
  Conference on Knowledge Discovery and Data Mining V. 2}}.
  \bibinfo{pages}{6043--6053}.
\newblock


\bibitem[Jin et~al\mbox{.}(2024)]%
        {jin2023time}
\bibfield{author}{\bibinfo{person}{Ming Jin}, \bibinfo{person}{Shiyu Wang},
  \bibinfo{person}{Lintao Ma}, \bibinfo{person}{Zhixuan Chu},
  \bibinfo{person}{James~Y Zhang}, \bibinfo{person}{Xiaoming Shi},
  \bibinfo{person}{Pin-Yu Chen}, \bibinfo{person}{Yuxuan Liang},
  \bibinfo{person}{Yuan-Fang Li}, \bibinfo{person}{Shirui Pan},
  {et~al\mbox{.}}} \bibinfo{year}{2024}\natexlab{}.
\newblock \showarticletitle{{Time-LLM}: Time series forecasting by
  reprogramming large language models}. In \bibinfo{booktitle}{\emph{ICLR}}.
\newblock


\bibitem[Kim et~al\mbox{.}(2024)]%
        {MultiModalForecaster2411}
\bibfield{author}{\bibinfo{person}{Kai Kim}, \bibinfo{person}{Howard Tsai},
  \bibinfo{person}{Rajat Sen}, \bibinfo{person}{Abhimanyu Das},
  \bibinfo{person}{Zihao Zhou}, \bibinfo{person}{Abhishek Tanpure},
  \bibinfo{person}{Mathew Luo}, {and} \bibinfo{person}{Rose Yu}.}
  \bibinfo{year}{2024}\natexlab{}.
\newblock \bibinfo{title}{Multi-Modal Forecaster: Jointly Predicting Time
  Series and Textual Data}.
\newblock \bibinfo{howpublished}{arXiv:2411.06735}.
\newblock
\newblock
\shownote{Preprint}.


\bibitem[Kingma and Ba(2014)]%
        {kingma2014adam}
\bibfield{author}{\bibinfo{person}{Diederik~P Kingma} {and}
  \bibinfo{person}{Jimmy Ba}.} \bibinfo{year}{2014}\natexlab{}.
\newblock \showarticletitle{Adam: A method for stochastic optimization}.
\newblock \bibinfo{journal}{\emph{arXiv preprint arXiv:1412.6980}}
  (\bibinfo{year}{2014}).
\newblock


\bibitem[Kitaev et~al\mbox{.}(2020)]%
        {kitaev2020reformer}
\bibfield{author}{\bibinfo{person}{Nikita Kitaev}, \bibinfo{person}{{\L}ukasz
  Kaiser}, {and} \bibinfo{person}{Anselm Levskaya}.}
  \bibinfo{year}{2020}\natexlab{}.
\newblock \showarticletitle{Reformer: The efficient transformer}. In
  \bibinfo{booktitle}{\emph{ICLR}}.
\newblock


\bibitem[Le and Mikolov(2014)]%
        {le2014doc2vec}
\bibfield{author}{\bibinfo{person}{Quoc~V. Le} {and} \bibinfo{person}{Tomas
  Mikolov}.} \bibinfo{year}{2014}\natexlab{}.
\newblock \showarticletitle{Distributed Representations of Sentences and
  Documents}. In \bibinfo{booktitle}{\emph{ICML}}.
\newblock
\urldef\tempurl%
\url{https://proceedings.mlr.press/v32/le14.html}
\showURL{%
\tempurl}


\bibitem[Lee et~al\mbox{.}(2024)]%
        {lee2024learning}
\bibfield{author}{\bibinfo{person}{Seunghan Lee}, \bibinfo{person}{Taeyoung
  Park}, {and} \bibinfo{person}{Kibok Lee}.} \bibinfo{year}{2024}\natexlab{}.
\newblock \showarticletitle{Learning to embed time series patches
  independently}. In \bibinfo{booktitle}{\emph{International Conference on
  Learning Representations}}, Vol.~\bibinfo{volume}{2024}.
  \bibinfo{pages}{27599--27620}.
\newblock


\bibitem[Lee et~al\mbox{.}(2026)]%
        {lee2026finstar}
\bibfield{author}{\bibinfo{person}{Seunghan Lee}, \bibinfo{person}{Jun Seo},
  \bibinfo{person}{Jaehoon Lee}, \bibinfo{person}{Sungdong Yoo},
  \bibinfo{person}{Minjae Kim}, \bibinfo{person}{Tae~Yoon Lim},
  \bibinfo{person}{Dongwan Kang}, \bibinfo{person}{Hwanil Choi},
  \bibinfo{person}{Soonyoung Lee}, {and} \bibinfo{person}{Wonbin Ahn}.}
  \bibinfo{year}{2026}\natexlab{}.
\newblock \showarticletitle{FinSTaR: Towards Financial Reasoning with Time
  Series Reasoning Models}.
\newblock \bibinfo{journal}{\emph{arXiv preprint arXiv:2605.03460}}
  (\bibinfo{year}{2026}).
\newblock


\bibitem[Li et~al\mbox{.}(2025)]%
        {li2025language}
\bibfield{author}{\bibinfo{person}{Zihao Li}, \bibinfo{person}{Xiao Lin},
  \bibinfo{person}{Zhining Liu}, \bibinfo{person}{Jiaru Zou},
  \bibinfo{person}{Ziwei Wu}, \bibinfo{person}{Lecheng Zheng},
  \bibinfo{person}{Dongqi Fu}, \bibinfo{person}{Yada Zhu},
  \bibinfo{person}{Hendrik Hamann}, \bibinfo{person}{Hanghang Tong},
  {et~al\mbox{.}}} \bibinfo{year}{2025}\natexlab{}.
\newblock \showarticletitle{Language in the flow of time: Time-series-paired
  texts weaved into a unified temporal narrative}.
\newblock \bibinfo{journal}{\emph{arXiv preprint arXiv:2502.08942}}
  (\bibinfo{year}{2025}).
\newblock


\bibitem[Lin et~al\mbox{.}(2026)]%
        {lin2026timi}
\bibfield{author}{\bibinfo{person}{Jiafeng Lin}, \bibinfo{person}{Yuxuan Wang},
  \bibinfo{person}{Huakun Luo}, \bibinfo{person}{Zhongyi Pei}, {and}
  \bibinfo{person}{Jianmin Wang}.} \bibinfo{year}{2026}\natexlab{}.
\newblock \showarticletitle{TiMi: Empower Time Series Transformers with
  Multimodal Mixture of Experts}.
\newblock \bibinfo{journal}{\emph{arXiv preprint arXiv:2602.21693}}
  (\bibinfo{year}{2026}).
\newblock


\bibitem[Liu et~al\mbox{.}(2025)]%
        {liu2024timecma}
\bibfield{author}{\bibinfo{person}{Chenxi Liu}, \bibinfo{person}{Qianxiong Xu},
  \bibinfo{person}{Hao Miao}, \bibinfo{person}{Sun Yang},
  \bibinfo{person}{Lingzheng Zhang}, \bibinfo{person}{Cheng Long},
  \bibinfo{person}{Ziyue Li}, {and} \bibinfo{person}{Rui Zhao}.}
  \bibinfo{year}{2025}\natexlab{}.
\newblock \showarticletitle{Timecma: Towards {LLM}-empowered multivariate time
  series forecasting via cross-modality alignment}. In
  \bibinfo{booktitle}{\emph{AAAI}}.
\newblock


\bibitem[Liu et~al\mbox{.}(2024b)]%
        {liu2024time}
\bibfield{author}{\bibinfo{person}{Haoxin Liu}, \bibinfo{person}{Shangqing Xu},
  \bibinfo{person}{Zhiyuan Zhao}, \bibinfo{person}{Lingkai Kong},
  \bibinfo{person}{Harshavardhan Prabhakar~Kamarthi}, \bibinfo{person}{Aditya
  Sasanur}, \bibinfo{person}{Megha Sharma}, \bibinfo{person}{Jiaming Cui},
  \bibinfo{person}{Qingsong Wen}, \bibinfo{person}{Chao Zhang},
  {et~al\mbox{.}}} \bibinfo{year}{2024}\natexlab{b}.
\newblock \showarticletitle{Time-mmd: Multi-domain multimodal dataset for time
  series analysis}.
\newblock \bibinfo{journal}{\emph{Advances in Neural Information Processing
  Systems}}  \bibinfo{volume}{37} (\bibinfo{year}{2024}),
  \bibinfo{pages}{77888--77933}.
\newblock


\bibitem[Liu et~al\mbox{.}(2022)]%
        {Liu2022NonstationaryTransformer}
\bibfield{author}{\bibinfo{person}{Y. Liu} {et~al\mbox{.}}}
  \bibinfo{year}{2022}\natexlab{}.
\newblock \showarticletitle{Non-stationary Transformers: Exploring the
  Stationarity in Time Series Forecasting}. In
  \bibinfo{booktitle}{\emph{Advances in Neural Information Processing Systems
  (NeurIPS)}}.
\newblock


\bibitem[Liu et~al\mbox{.}(2024a)]%
        {Liu2023iTransformer}
\bibfield{author}{\bibinfo{person}{Yong Liu}, \bibinfo{person}{Tengge Hu},
  \bibinfo{person}{Haoran Zhang}, \bibinfo{person}{Haixu Wu},
  \bibinfo{person}{Shiyu Wang}, \bibinfo{person}{Lintao Ma}, {and}
  \bibinfo{person}{Mingsheng Long}.} \bibinfo{year}{2024}\natexlab{a}.
\newblock \showarticletitle{itransformer: Inverted transformers are effective
  for time series forecasting}. In \bibinfo{booktitle}{\emph{ICLR}}.
\newblock


\bibitem[Liu et~al\mbox{.}(2023)]%
        {Koopa2023}
\bibfield{author}{\bibinfo{person}{Yong Liu}, \bibinfo{person}{Chenyu Li},
  \bibinfo{person}{Jianmin Wang}, {and} \bibinfo{person}{Mingsheng Long}.}
  \bibinfo{year}{2023}\natexlab{}.
\newblock \showarticletitle{Koopa: Learning non-stationary time series dynamics
  with koopman predictors}.
\newblock \bibinfo{journal}{\emph{Advances in neural information processing
  systems}}  \bibinfo{volume}{36} (\bibinfo{year}{2023}),
  \bibinfo{pages}{12271--12290}.
\newblock


\bibitem[Nguyen et~al\mbox{.}(2026)]%
        {SpecTF2602}
\bibfield{author}{\bibinfo{person}{Huu~Hiep Nguyen},
  \bibinfo{person}{Minh~Hoang Nguyen}, \bibinfo{person}{Dung Nguyen}, {and}
  \bibinfo{person}{Hung Le}.} \bibinfo{year}{2026}\natexlab{}.
\newblock \bibinfo{title}{Spectral Text Fusion: A Frequency-Aware Approach to
  Multimodal Time-Series Forecasting}.
\newblock \bibinfo{howpublished}{arXiv:2602.01588}.
\newblock
\newblock
\shownote{Preprint}.


\bibitem[Nie et~al\mbox{.}(2023)]%
        {Nie2023PatchTST}
\bibfield{author}{\bibinfo{person}{Y. Nie}, \bibinfo{person}{N.~H. Nguyen},
  \bibinfo{person}{P. Sinthong}, {and} \bibinfo{person}{J. Kalagnanam}.}
  \bibinfo{year}{2023}\natexlab{}.
\newblock \showarticletitle{A Time Series is Worth 64 Words: Long-term
  Forecasting with Transformers}.
\newblock \bibinfo{journal}{\emph{arXiv Preprint arXiv:2211.14730}}
  (\bibinfo{year}{2023}).
\newblock


\bibitem[Park et~al\mbox{.}(2025)]%
        {UniCast2508}
\bibfield{author}{\bibinfo{person}{Sehyuk Park}, \bibinfo{person}{Soyeon~Caren
  Han}, {and} \bibinfo{person}{Eduard Hovy}.} \bibinfo{year}{2025}\natexlab{}.
\newblock \bibinfo{title}{UniCast: A Unified Framework for Instance-Conditioned
  Multimodal Time-Series Forecasting}.
\newblock \bibinfo{howpublished}{arXiv:2508.11954}.
\newblock
\newblock
\shownote{Preprint}.


\bibitem[Perez et~al\mbox{.}(2018)]%
        {perez2018film}
\bibfield{author}{\bibinfo{person}{Ethan Perez}, \bibinfo{person}{Florian
  Strub}, \bibinfo{person}{Harm De~Vries}, \bibinfo{person}{Vincent Dumoulin},
  {and} \bibinfo{person}{Aaron Courville}.} \bibinfo{year}{2018}\natexlab{}.
\newblock \showarticletitle{Film: Visual reasoning with a general conditioning
  layer}. In \bibinfo{booktitle}{\emph{AAAI}}.
\newblock


\bibitem[Radford et~al\mbox{.}(2019)]%
        {radford2019gpt2}
\bibfield{author}{\bibinfo{person}{Alec Radford}, \bibinfo{person}{Jeff Wu},
  \bibinfo{person}{Rewon Child}, \bibinfo{person}{David Luan},
  \bibinfo{person}{Dario Amodei}, {and} \bibinfo{person}{Ilya Sutskever}.}
  \bibinfo{year}{2019}\natexlab{}.
\newblock \showarticletitle{Language Models are Unsupervised Multitask
  Learners}.
\newblock \bibinfo{journal}{\emph{OpenAI}} (\bibinfo{year}{2019}).
\newblock
\urldef\tempurl%
\url{https://cdn.openai.com/better-language-models/language_models_are_unsupervised_multitask_learners.pdf}
\showURL{%
\tempurl}
\newblock
\shownote{Technical report}.


\bibitem[Su et~al\mbox{.}(2025)]%
        {su2025multimodal}
\bibfield{author}{\bibinfo{person}{Chen Su}, \bibinfo{person}{Yuanhe Tian},
  {and} \bibinfo{person}{Yan Song}.} \bibinfo{year}{2025}\natexlab{}.
\newblock \showarticletitle{Multimodal conditioned diffusive time series
  forecasting}.
\newblock \bibinfo{journal}{\emph{arXiv preprint arXiv:2504.19669}}
  (\bibinfo{year}{2025}).
\newblock


\bibitem[Vaswani et~al\mbox{.}(2017)]%
        {vaswani2017attention}
\bibfield{author}{\bibinfo{person}{Ashish Vaswani}, \bibinfo{person}{Noam
  Shazeer}, \bibinfo{person}{Niki Parmar}, \bibinfo{person}{Jakob Uszkoreit},
  \bibinfo{person}{Llion Jones}, \bibinfo{person}{Aidan~N Gomez},
  \bibinfo{person}{{\L}ukasz Kaiser}, {and} \bibinfo{person}{Illia
  Polosukhin}.} \bibinfo{year}{2017}\natexlab{}.
\newblock \showarticletitle{Attention is all you need}. In
  \bibinfo{booktitle}{\emph{NeurIPS}}.
\newblock


\bibitem[Wang et~al\mbox{.}(2025)]%
        {wang2025chattime}
\bibfield{author}{\bibinfo{person}{Chengsen Wang}, \bibinfo{person}{Qi Qi},
  \bibinfo{person}{Jingyu Wang}, \bibinfo{person}{Haifeng Sun},
  \bibinfo{person}{Zirui Zhuang}, \bibinfo{person}{Jinming Wu},
  \bibinfo{person}{Lei Zhang}, {and} \bibinfo{person}{Jianxin Liao}.}
  \bibinfo{year}{2025}\natexlab{}.
\newblock \showarticletitle{Chattime: A unified multimodal time series
  foundation model bridging numerical and textual data}. In
  \bibinfo{booktitle}{\emph{AAAI}}.
\newblock


\bibitem[Wang et~al\mbox{.}(2024)]%
        {wang2024timexer}
\bibfield{author}{\bibinfo{person}{Yuxuan Wang}, \bibinfo{person}{Haixu Wu},
  \bibinfo{person}{Jiaxiang Dong}, \bibinfo{person}{Guo Qin},
  \bibinfo{person}{Haoran Zhang}, \bibinfo{person}{Yong Liu},
  \bibinfo{person}{Yunzhong Qiu}, \bibinfo{person}{Jianmin Wang}, {and}
  \bibinfo{person}{Mingsheng Long}.} \bibinfo{year}{2024}\natexlab{}.
\newblock \showarticletitle{Timexer: Empowering transformers for time series
  forecasting with exogenous variables}.
\newblock \bibinfo{journal}{\emph{Advances in Neural Information Processing
  Systems}}  \bibinfo{volume}{37} (\bibinfo{year}{2024}),
  \bibinfo{pages}{469--498}.
\newblock


\bibitem[Wu et~al\mbox{.}(2021)]%
        {wu2021autoformer}
\bibfield{author}{\bibinfo{person}{Haixu Wu}, \bibinfo{person}{Jiehui Xu},
  \bibinfo{person}{Jianmin Wang}, {and} \bibinfo{person}{Mingsheng Long}.}
  \bibinfo{year}{2021}\natexlab{}.
\newblock \showarticletitle{Autoformer: Decomposition transformers with
  auto-correlation for long-term series forecasting}. In
  \bibinfo{booktitle}{\emph{NeurIPS}}.
\newblock


\bibitem[Xue and Salim(2023)]%
        {xue2023promptcast}
\bibfield{author}{\bibinfo{person}{Hao Xue} {and} \bibinfo{person}{Flora~D
  Salim}.} \bibinfo{year}{2023}\natexlab{}.
\newblock \showarticletitle{Promptcast: A new prompt-based learning paradigm
  for time series forecasting}.
\newblock \bibinfo{journal}{\emph{IEEE Transactions on Knowledge and Data
  Engineering}} \bibinfo{volume}{36}, \bibinfo{number}{11}
  (\bibinfo{year}{2023}), \bibinfo{pages}{6851--6864}.
\newblock


\bibitem[Yao et~al\mbox{.}(2025)]%
        {CAPTime2505}
\bibfield{author}{\bibinfo{person}{Yueyang Yao}, \bibinfo{person}{Jiajun Li},
  \bibinfo{person}{Xingyuan Dai}, \bibinfo{person}{MengMeng Zhang},
  \bibinfo{person}{Xiaoyan Gong}, \bibinfo{person}{Fei-Yue Wang}, {and}
  \bibinfo{person}{Yisheng Lv}.} \bibinfo{year}{2025}\natexlab{}.
\newblock \bibinfo{title}{Context-Aware Probabilistic Modeling with LLM for
  Multimodal Time Series Forecasting}.
\newblock \bibinfo{howpublished}{arXiv:2505.10774}.
\newblock
\newblock
\shownote{Preprint}.


\bibitem[Yi et~al\mbox{.}(2023)]%
        {yi2023frequency}
\bibfield{author}{\bibinfo{person}{Kun Yi}, \bibinfo{person}{Qi Zhang},
  \bibinfo{person}{Wei Fan}, \bibinfo{person}{Shoujin Wang},
  \bibinfo{person}{Pengyang Wang}, \bibinfo{person}{Hui He},
  \bibinfo{person}{Ning An}, \bibinfo{person}{Defu Lian},
  \bibinfo{person}{Longbing Cao}, {and} \bibinfo{person}{Zhendong Niu}.}
  \bibinfo{year}{2023}\natexlab{}.
\newblock \showarticletitle{Frequency-domain mlps are more effective learners
  in time series forecasting}.
\newblock \bibinfo{journal}{\emph{Advances in Neural Information Processing
  Systems}}  \bibinfo{volume}{36} (\bibinfo{year}{2023}),
  \bibinfo{pages}{76656--76679}.
\newblock


\bibitem[Zeng et~al\mbox{.}(2023)]%
        {Zeng2022DLinear}
\bibfield{author}{\bibinfo{person}{A. Zeng}, \bibinfo{person}{M. Chen},
  \bibinfo{person}{L. Zhang}, {and} \bibinfo{person}{Q. Xu}.}
  \bibinfo{year}{2023}\natexlab{}.
\newblock \showarticletitle{Are Transformers Effective for Time Series
  Forecasting?}. In \bibinfo{booktitle}{\emph{AAAI}}.
\newblock


\bibitem[Zhang and Yan(2023)]%
        {zhang2023crossformer}
\bibfield{author}{\bibinfo{person}{Yunhao Zhang} {and} \bibinfo{person}{Junchi
  Yan}.} \bibinfo{year}{2023}\natexlab{}.
\newblock \showarticletitle{Crossformer: Transformer utilizing cross-dimension
  dependency for multivariate time series forecasting}. In
  \bibinfo{booktitle}{\emph{ICLR}}.
\newblock


\bibitem[Zhong et~al\mbox{.}(2025)]%
        {TimeVLM2502}
\bibfield{author}{\bibinfo{person}{Siru Zhong}, \bibinfo{person}{Weilin Ruan},
  \bibinfo{person}{Ming Jin}, \bibinfo{person}{Huan Li},
  \bibinfo{person}{Qingsong Wen}, {and} \bibinfo{person}{Yuxuan Liang}.}
  \bibinfo{year}{2025}\natexlab{}.
\newblock \showarticletitle{{Time-VLM}: Exploring Multimodal Vision-Language
  Models for Augmented Time Series Forecasting}. In
  \bibinfo{booktitle}{\emph{Proceedings of the 42nd International Conference on
  Machine Learning (ICML)}} \emph{(\bibinfo{series}{PMLR},
  Vol.~\bibinfo{volume}{267})}. \bibinfo{pages}{78478--78497}.
\newblock


\bibitem[Zhou et~al\mbox{.}(2021)]%
        {zhou2021informer}
\bibfield{author}{\bibinfo{person}{Haoyi Zhou}, \bibinfo{person}{Shanghang
  Zhang}, \bibinfo{person}{Jieqi Peng}, \bibinfo{person}{Shuai Zhang},
  \bibinfo{person}{Jianxin Li}, \bibinfo{person}{Hui Xiong}, {and}
  \bibinfo{person}{Wancai Zhang}.} \bibinfo{year}{2021}\natexlab{}.
\newblock \showarticletitle{Informer: Beyond efficient transformer for long
  sequence time-series forecasting}. In \bibinfo{booktitle}{\emph{AAAI}}.
\newblock


\bibitem[Zhou et~al\mbox{.}(2025)]%
        {BALMTSF2509}
\bibfield{author}{\bibinfo{person}{Shiqiao Zhou}, \bibinfo{person}{Holger
  Schöner}, \bibinfo{person}{Huanbo Lyu}, \bibinfo{person}{Edouard Fouché},
  {and} \bibinfo{person}{Shuo Wang}.} \bibinfo{year}{2025}\natexlab{}.
\newblock \bibinfo{title}{{BALM-TSF}: Balanced Multimodal Alignment for
  {LLM}-Based Time Series Forecasting}.
\newblock \bibinfo{howpublished}{arXiv:2509.00622}.
\newblock
\newblock
\shownote{Preprint}.


\bibitem[Zhou et~al\mbox{.}(2022a)]%
        {zhou2022fedformer}
\bibfield{author}{\bibinfo{person}{Tian Zhou}, \bibinfo{person}{Ziqing Ma},
  \bibinfo{person}{Qingsong Wen}, \bibinfo{person}{Xue Wang},
  \bibinfo{person}{Liang Sun}, {and} \bibinfo{person}{Rong Jin}.}
  \bibinfo{year}{2022}\natexlab{a}.
\newblock \showarticletitle{Fedformer: Frequency enhanced decomposed
  transformer for long-term series forecasting}. In
  \bibinfo{booktitle}{\emph{ICML}}.
\newblock


\bibitem[Zhou et~al\mbox{.}(2022b)]%
        {Zhou2022FiLM}
\bibfield{author}{\bibinfo{person}{T. Zhou}, \bibinfo{person}{Z. Ma},
  \bibinfo{person}{Q. Wen}, \bibinfo{person}{X. Wang}, \bibinfo{person}{L.
  Sun}, {and} \bibinfo{person}{R. Jin}.} \bibinfo{year}{2022}\natexlab{b}.
\newblock \showarticletitle{FiLM: Frequency Improved Legendre Memory Model for
  Long-Term Time Series Forecasting}.
\newblock \bibinfo{journal}{\emph{Advances in Neural Information Processing
  Systems}} (\bibinfo{year}{2022}).
\newblock


\bibitem[Zhu et~al\mbox{.}(2025)]%
        {liu2025pa}
\bibfield{author}{\bibinfo{person}{Enqiang Zhu}, \bibinfo{person}{Zhenbin
  Deng}, \bibinfo{person}{Shengzhi Wang}, \bibinfo{person}{Yi-Kun Tang}, {and}
  \bibinfo{person}{Chanjuan Liu}.} \bibinfo{year}{2025}\natexlab{}.
\newblock \showarticletitle{PA-RNet: Perturbation-Aware Residual Network for
  Robust Multimodal Time Series Forecasting}.
\newblock \bibinfo{journal}{\emph{arXiv preprint arXiv:2508.04750}}
  (\bibinfo{year}{2025}).
\newblock


\end{thebibliography}
\clearpage
\FloatBarrier
\appendix
\renewcommand{\thefigure}{\thesection.\arabic{figure}}
\renewcommand{\thetable}{\thesection.\arabic{table}}
\counterwithin{figure}{section}
\counterwithin{table}{section}

\section{Related Work}
\label{sec:related_works}
\textbf{TS forecasting models.}
Recent TS forecasting methods 
employ Transformers~\cite{vaswani2017attention}
to capture temporal and 
channel
dependencies. PatchTST~\cite{Nie2023PatchTST} segments TS into patches and adopts channel-independent modeling. PITS~\cite{lee2024learning} embeds patches independently without inter-patch interaction. Crossformer~\cite{zhang2023crossformer} captures 
channel
interactions via hierarchical attention. iTransformer~\cite{Liu2023iTransformer} applies 
attention across 
features 
to model 
channel
dependencies. Nonstationary Transformer~\cite{Liu2022NonstationaryTransformer} 
models non-stationarity within attention to address distribution shifts.
Several works adopt lightweight architectures without attention,
where DLinear \cite{Zeng2022DLinear} uses linear 
decomposition to capture trend and seasonal components. 
TSMixer \cite{Chen2023TSMixer} 
employs
MLP-based mixing across temporal and feature dimensions. 
TiDE \cite{Das2023TiDE} employs an MLP-based encoder-decoder with temporal embeddings. 
FiLM \cite{Zhou2022FiLM} 
introduces a Frequency-improved Legendre Memory model 
with Legendre projection and Fourier denoising.
Koopa \cite{Koopa2023} learns latent 
dynamics through Koopman operator.

\textbf{Multimodal TS forecasting models.}
Recent work integrates external modalities 
with TS to enrich forecasting with contextual information~\cite{jin2023time,xue2023promptcast, jiang2025multi}. UniCast~\cite{UniCast2508} combines pretrained vision and text encoders with a frozen TS foundation model.
 CAPTime~\cite{CAPTime2505} aligns TS representations with LLM-derived textual context within a probabilistic forecasting framework. BALM-TSF~\cite{BALMTSF2509} mitigates modality imbalance by aligning TS and 
 textual embeddings before fusion. SpecTF~\cite{SpecTF2602} performs frequency-domain fusion by projecting textual embeddings into the spectral space and integrating them with TS components. 
 Multi-Modal Forecaster~\cite{MultiModalForecaster2411} jointly models TS and text through shared embeddings.
  Time-VLM~\cite{TimeVLM2502} leverages pretrained vision-language models to construct multimodal representations. 
TimeCMA~\cite{liu2024timecma}
 performs
 cross-modality alignment 
 with dual 
 branches.
GPT4MTS~\cite{jia2024gpt4mts} employs LLMs to generate task-aware 
textual prompts to guide TS forecasting.
T3Time~\cite{chowdhury2025t3time} employs temporal, spectral, and prompt features with 
a
gating mechanism.

\setlength{\columnsep}{8pt}  
As shown in Table~\ref{tab:fusion_plugin_summary}, these 
works either adopt naive fusion strategies
or focus on designing architecture-specific models for multimodal forecasting,
which are not generally applicable to existing unimodal TS models.
While TaTS~\cite{li2025language} can be plugged into existing unimodal TS models, it adopts a naive first-layer additive fusion scheme 
included
as a baseline in our experiments.
Additionally, although ContextFormer~\cite{chattopadhyay2024context} can be applied to arbitrary 
TS encoders, its module accounts for over 75\% of the total parameters due to multiple cross-attention layers,
limiting general applicability.




\section{Details of Models}
\label{app:model}

We use the \texttt{mmtslib} package provided by Time-MMD~\cite{liu2024time}. We adopt 14 TS models and 4 text models implemented in the library.
\textbf{TS Models.} We use 14 TS models from three categories as follows:

\setlist[itemize]{leftmargin=0.3cm}
\begin{itemize}

    \item \textbf{Transformer-based Models}
    \begin{itemize}
        \item \textbf{Transformer}~\cite{vaswani2017attention}: A standard sequence-to-sequence model based on multi-head self-attention.
        
        \item \textbf{Informer}~\cite{zhou2021informer}: A sparse-attention Transformer designed for efficient long-term TS forecasting.
        
        \item \textbf{Reformer}~\cite{kitaev2020reformer}: An efficient Transformer variant using locality-sensitive hashing and reversible layers.
        
        \item \textbf{Autoformer}~\cite{wu2021autoformer}: A decomposition-based Transformer that replaces attention with an auto-correlation mechanism.
        
        \item \textbf{FEDformer}~\cite{zhou2022fedformer}: A frequency-enhanced Transformer leveraging Fourier decomposition for long-horizon prediction.
        
        \item \textbf{Crossformer}~\cite{zhang2023crossformer}: A multivariate Transformer that 
        models cross-dimension dependencies.
        
        \item \textbf{iTransformer}~\cite{Liu2023iTransformer}: An inverted Transformer that treats variates as tokens to model inter-variable relationships.
        
        \item \textbf{Nonstationary Transformer}~\cite{Liu2022NonstationaryTransformer}: A Transformer tailored to capture non-stationary temporal patterns.
        
        \item \textbf{PatchTST}~\cite{Nie2023PatchTST}: A patch-based Transformer that segments TS into subseries tokens for scalable forecasting.
    \end{itemize}

    \item \textbf{Linear / MLP-based Models}
    \begin{itemize}
        \item \textbf{DLinear}~\cite{Zeng2022DLinear}: A linear model that performs forecasting via direct regression on historical inputs.
        
        \item \textbf{TiDE}~\cite{Das2023TiDE}: An MLP-based encoder-decoder model designed for long-term TS forecasting.
        
        \item \textbf{TSMixer}~\cite{Chen2023TSMixer}: A fully-MLP architecture that mixes temporal and feature information for prediction.
    \end{itemize}

    \item \textbf{Other Architectures}
    \begin{itemize}
        \item \textbf{Koopa}~\cite{Koopa2023}: A Koopman-operator-inspired model that decomposes TS into stable and dynamic components.
        
        \item \textbf{FiLM}~\cite{Zhou2022FiLM}: A frequency-based method that applies Fourier and Legendre projections for denoising and trend modeling.
    \end{itemize}

\end{itemize}

\textbf{Text Models.} We use 4 text models, including three LLMs as follows:
\begin{itemize}
    \item \textbf{BERT}~\cite{devlin2018bert}: A bidirectional Transformer pre-trained with masked language modeling.
    
    \item \textbf{GPT-2}~\cite{radford2019gpt2}: An autoregressive Transformer trained to predict next-token distributions.
    
    \item \textbf{Llama-3}~\cite{dubey2024llama3}: A recent large language model with improved text generation and comprehension capabilities.
    
    \item \textbf{Doc2Vec}~\cite{le2014doc2vec}: A document-level embedding model that learns fixed-length vector representations of text.
\end{itemize}

\section{Details of Datasets}
\label{app:dataset}

We use the nine multimodal datasets from various domains proposed in Time-MMD~\cite{liu2024time}.
The input length ($L$) and forecasting horizons ($H$) are determined according to the frequency of each dataset (daily, weekly, or monthly). 
Detailed information about the datasets is provided in Table~\ref{tab:dataset_full_meta}.


The qualitative descriptions of each dataset are as follows:

\setlist[itemize]{leftmargin=0.3cm}
\begin{itemize}
    \item \textbf{Agriculture}: Tracks U.S. retail broiler (chicken) composite prices, reflecting supply–demand dynamics and seasonal patterns in the agricultural market.

    \item \textbf{Climate}: Measures drought severity levels across regions, capturing long-term climate variability and extreme weather trends.

    \item \textbf{Economy}: Represents the U.S. international trade balance, indicating macroeconomic conditions and global trade fluctuations.

    \item \textbf{Energy}: Records U.S. gasoline prices, a key indicator of energy market volatility and consumer economic burden.

    \item \textbf{Environment}: Monitors daily air quality index (AQI), reflecting pollution dynamics and environmental risk levels.

    \item \textbf{Health}: Tracks weekly influenza-like illness (ILI) proportions, serving as a proxy for epidemic spread and public health trends.

    \item \textbf{Security}: Captures disaster and emergency grant allocations, reflecting the temporal impact of large-scale natural and societal crises.

    \item \textbf{Social Good}: Measures unemployment rates, highlighting labor market disparities and socioeconomic stability.

    \item \textbf{Traffic}: Represents travel volume statistics, indicating mobility trends and transportation demand dynamics.

\end{itemize}

\begin{table}[!t]
\centering
\small
\caption{\textbf{Meta information of the nine Time-MMD~\cite{liu2024time} datasets.}
We report the domain, dimensionality, data frequency, number of samples, timespan, input length ($L$), and forecasting horizons ($H$) for all nine multimodal datasets.}
\begin{adjustbox}{width=\linewidth}
\begin{NiceTabular}{l c c c c c l}
\toprule
Domain & Dim. & Freq. & \#Samples & Timespan & $L$ & $H$ \\
\midrule

Agriculture
& 1
& M
& 496
& 1983--Present
& 24
& \{6, 8, 10, 12\} \\

Climate
& 5
& M
& 496
& 1983--Present
& 24
& \{6, 8, 10, 12\} \\

Economy
& 3
& M
& 423
& 1989--Present
& 24
& \{6, 8, 10, 12\} \\

Energy
& 9
& W
& 1479
& 1996--Present
& 48
& \{12, 24, 36, 48\} \\

Environment
& 4
& D
& 11102
& 1982--2023
& 336
& \{48, 96, 192, 336\} \\

Health
& 11
& W
& 1389
& 1997--Present
& 48
& \{12, 24, 36, 48\} \\

Security
& 1
& M
& 297
& 1999--Present
& 24
& \{6, 8, 10, 12\} \\

Social Good
& 1
& M
& 900
& 1950--Present
& 24
& \{6, 8, 10, 12\} \\

Traffic
& 1
& M
& 531
& 1980--Present
& 24
& \{6, 8, 10, 12\} \\

\bottomrule
\end{NiceTabular}
\end{adjustbox}
\label{tab:dataset_full_meta}
\end{table}

\section{Experimental Setup}
\label{app:exp_setup}

\textbf{(1) Evaluation metrics.}
We evaluate forecasting performance using Mean Squared Error (MSE) and Mean Absolute Error (MAE), which are standard metrics in TS forecasting.

\textbf{(2) Input and output horizons.}
We consider different forecasting horizon settings depending on the reporting frequency of each dataset, as shown in Table~\ref{tbl:window_size}.
\begin{table}[!t]
\centering
\caption{Input and output horizon settings.}
\begin{NiceTabular}{lcc}
\toprule
\textbf{Frequency} & \textbf{Lookback window ($L$)} & \textbf{Forecast horizons ($H$)}  \\
\midrule
Daily   & 96 & [48, 96, 192, 336] \\
Weekly  & 36 & [12, 24, 36, 48] \\
Monthly & 8& [6, 8, 10, 12]  \\
\bottomrule
\end{NiceTabular}
\label{tbl:window_size}
\end{table}

\textbf{(3) Optimizer.} We use the Adam optimizer~\cite{kingma2014adam} for all trainable modules with a batch size of 32.

\textbf{(4) Learning rate.}
We assign separate optimizers and learning rates to different components. The default learning rates are defined as follows:
\begin{table}[!t]
\centering
\caption{Learning rate configuration.}
\begin{adjustbox}{max width=\linewidth}
\begin{tabular}{lccc}
\toprule
\textbf{Optimizer} & Target module & Learning rate & Default value \\
\midrule
Model Optimizer        & Time Series Model     & $\eta_{\mathrm{TS}}$        & $1\times10^{-4}$ \\
MLP Optimizer          & Text Embedding MLP    & $\eta_{\mathrm{MLP}}$       & $1\times10^{-2}$ \\
Projection Optimizer   & Projection Layer      & $\eta_{\mathrm{Proj}}$      & $1\times10^{-3}$ \\
\bottomrule
\end{tabular}
\end{adjustbox}
\end{table}

To identify the best configuration, we multiply each default learning rate by the following scaling factors:
\{0.05,\ 0.1,\ 0.5,\ 1.0,\ 2.0,\ 5.0,\ 10.0,\ 20.0,\ 50.0,\ 100.0\},
resulting in 10 different learning rate configurations per component.

\textbf{(5) Epochs. }
We train the model for a maximum of 10 epochs with patience 5.

\textbf{(6) Data split.}
All datasets are divided chronologically into train:validation:test = 7:1:2.

\textbf{(7) Compute resource.}
All experiments are conducted on a single NVIDIA L40-48G GPU.

\section{Model Hyperparameters}
\label{app:model_hyperparams}
Table~\ref{tbl:model_hyperparams} presents the hyperparameter settings of the 14 TS models used in our experiments. 
We follow the default configurations adopted in Time-MMD~\cite{liu2024time}. 
Here, $d_{\text{model}}$ denotes the model (hidden) dimension, 
$n_{\text{heads}}$ the number of attention heads, 
$L_{\text{enc}}$/$L_{\text{dec}}$ the number of encoder/decoder layers, 
$d_{\text{ff}}$ the feed-forward dimension, 
and $p_{\text{drop}}$ the dropout rate.

Note that since DLinear~\cite{Zeng2022DLinear} has no hidden representation, 
all three injection positions (first, middle, and last) operate directly on the channel dimension 
$C_{\text{in}}$ rather than a latent dimension $d_{\text{model}}$. 
Specifically, first fusion adds or concatenates the projected text embedding 
$\mathbf{W}_{\text{proj}}\mathbf{e}_t \in \mathbb{R}^{C_{\text{in}}}$ 
to the raw input $\mathbf{x} \in \mathbb{R}^{L \times C_{\text{in}}}$ before decomposition. 
Middle fusion injects the text signal into the seasonal and trend components 
$\in \mathbb{R}^{C_{\text{in}} \times H}$ independently after the 
$\texttt{seq\_len} \rightarrow \texttt{pred\_len}$ linear projection but before their summation. 
Last fusion modifies the final output $\in \mathbb{R}^{H \times C_{\text{in}}}$ after the linear mapping. 
For CFA, the adapter bottleneck dimension is computed as 
$\lfloor C_{\text{in}} / r \rfloor$ rather than 
$\lfloor d_{\text{model}} / r \rfloor$. 
Thus, the low-rank text residual is injected into the variate space instead of a token-level or hidden-level space.

\begin{table}[!t]
\centering
\caption{Default hyperparameters of the 14 TS models.}
\label{tbl:model_hyperparams}
\begin{adjustbox}{width=\linewidth}
\begin{NiceTabular}{l c c c c c c}
\toprule
\textbf{Model}
  & $\bm{d}_{\textbf{model}}$
  & $\bm{n}_{\textbf{heads}}$
  & $\bm{L}_{\textbf{enc}}$
  & $\bm{L}_{\textbf{dec}}$
  & $\bm{d}_{\textbf{ff}}$
  & $\bm{p}_{\textbf{drop}}$ \\
\midrule

\multicolumn{7}{l}{\textit{\textbf{Transformer}}-based methods} \\
\midrule
Transformer~\cite{vaswani2017attention}
  & 512 & 8 & 2 & 1 & 2048 & 0.1 \\
Informer~\cite{zhou2021informer}
  & 512 & 8 & 2 & 1 & 2048 & 0.1 \\
Autoformer~\cite{wu2021autoformer}
  & 512 & 8 & 2 & 1 & 2048 & 0.1 \\
FEDformer~\cite{zhou2022fedformer}
  & 512 & 8 & 2 & 1 & 2048 & 0.1 \\
Nonstationary Transformer~\cite{Liu2022NonstationaryTransformer}
  & 512 & 8 & 2 & 1 & 2048 & 0.1 \\
Reformer~\cite{kitaev2020reformer}
  & 512 & 8 & 2 & --- & 2048 & 0.1 \\
iTransformer~\cite{Liu2023iTransformer}
  & 512 & 8 & 2 & --- & 2048 & 0.1 \\
PatchTST~\cite{Nie2023PatchTST}
  & 512 & 8 & 2 & --- & 2048 & 0.1 \\
Crossformer~\cite{zhang2023crossformer}
  & 512 & 8 & 2 & --- & 2048 & 0.1 \\

\midrule

\multicolumn{7}{l}{\textit{\textbf{Linear/MLP}}-based methods} \\
\midrule
DLinear~\cite{Zeng2022DLinear}
  & --- & --- & --- & --- & --- & 0.1 \\
TSMixer~\cite{Chen2023TSMixer}
  & 512 & --- & 2 & --- & --- & 0.1 \\
TiDE~\cite{Das2023TiDE}
  & 512 & --- & 2 & 1 & 2048 & 0.1 \\

\midrule

\multicolumn{7}{l}{\textit{\textbf{Others}}} \\
\midrule
FiLM~\cite{Zhou2022FiLM}
  & 512 & --- & 2 & --- & --- & 0.1 \\
Koopa~\cite{Koopa2023}
  & --- & --- & --- & --- & --- & 0.1 \\

\bottomrule
\end{NiceTabular}
\end{adjustbox}
\end{table}

\section{Toy Experiment with Multimodal Fusion}
\label{sec:appendix_toy}

\begin{figure*}[t]
\centering
\begin{adjustbox}{max width=\linewidth}
\includegraphics[width=.999\textwidth]{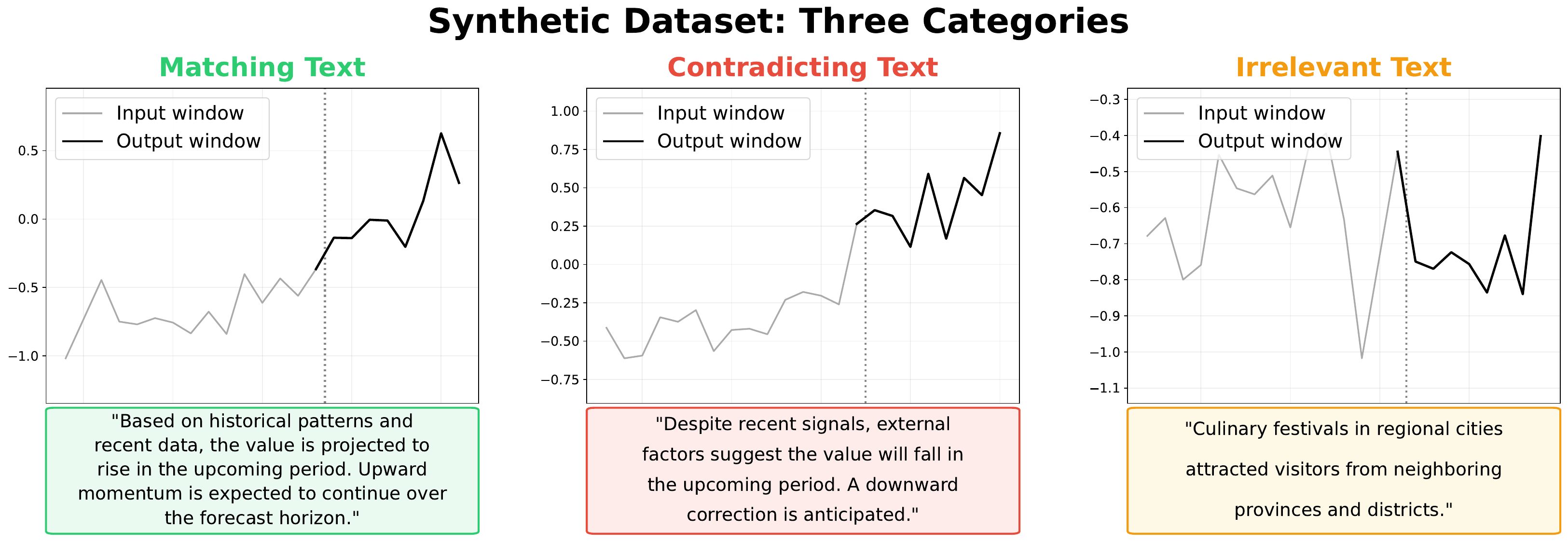}
\end{adjustbox}
\caption{
\textbf{Illustration of the toy dataset}. Each time step of the TS is paired with a text description, categorized as 1) matching, 2) contradicting, or 3) irrelevant. Matching descriptions align with the TS trend, contradicting ones oppose it, and irrelevant ones provide no useful signal.
}
\label{fig:toy_dataset}
\end{figure*}

\begin{figure*}[t]
\centering
\begin{minipage}[b]{0.38\textwidth}
  \centering
  \begin{adjustbox}{max width=\linewidth}
  \begin{tabular}{lcc}
  \toprule
   & \textbf{Matching} & \textbf{Contradicting} \\
  \midrule
  Mean         & 33.50 & 31.56 \\
  Std          & 3.56  & 3.07 \\
  $t$, $p$     & \multicolumn{2}{c}{$t=4.03$, $p<0.001$} \\
  Cohen's $d$  & \multicolumn{2}{c}{0.58} \\
  \bottomrule
  \end{tabular}
  \end{adjustbox}
  \captionof{table}{\textbf{Text-contribution ratio.} The table shows how much of the text signal survives the low-rank bottleneck for each text type. Matching text is injected more strongly than contradicting text, indicating that the adapter selectively suppresses conflicting information.}
  \label{tab:toy_repr}
\end{minipage}
\hfill
\begin{minipage}[b]{0.60\textwidth}
  \centering
  \begin{adjustbox}{max width=\linewidth}
    \includegraphics[width=0.99\linewidth]{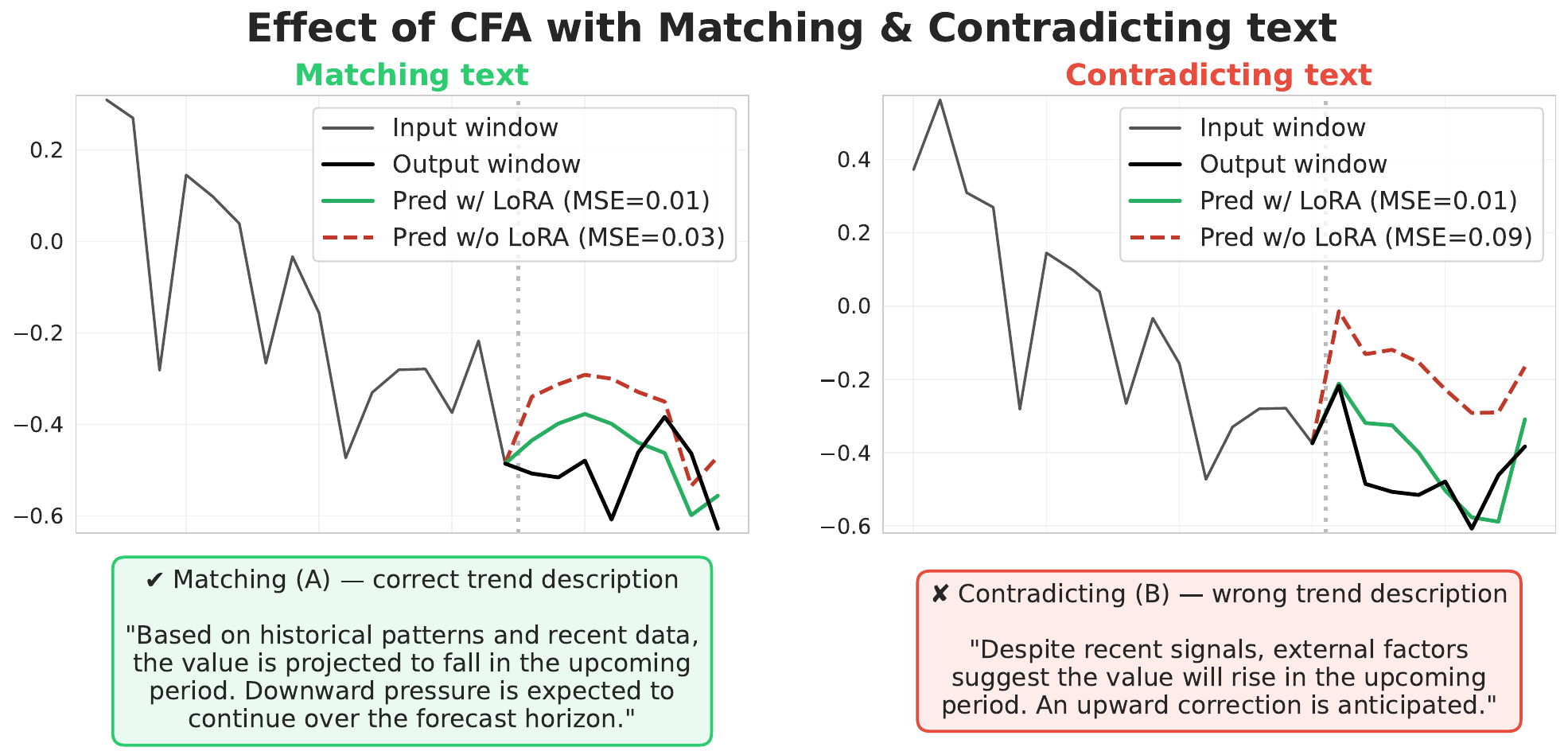}
  \end{adjustbox}
  \captionof{figure}{Forecast comparison for a matched (left) and a contradicting (right) text sample.
  \textcolor{green!60!black}{\textbf{Green solid}}: w/~bottleneck.
  \textcolor{red!80!black}{\textbf{Red dashed}}: w/o~bottleneck.
  \textbf{Black solid}: ground truth.
  The vertical dotted line separates the input window from the prediction horizon.}
  \label{fig:toy_viz}
\end{minipage}
\end{figure*}

We conduct a controlled toy experiment to analyze \textit{the effect of the
low-rank bottleneck (low-rank adapter)} in CFA. The low-rank adapter constrains the projection of text embeddings through a low-rank bottleneck before they are injected into the TS backbone, which is expected to \textit{suppress irrelevant or contradicting textual signals
while preserving useful guidance}.

To examine this effect, we compare CFA with and without the
low-rank constraint.
Throughout this section, \textbf{w/~bottleneck} denotes CFA with the
low-rank bottleneck, while \textbf{w/o~bottleneck} denotes the
middle-additive baseline that directly injects text embeddings
without the low-rank constraint.


\textbf{A. Toy dataset construction.}
We synthesize a univariate TS and pair each time step with a short
natural-language description.
Each description belongs to one of three categories:

\setlist[itemize]{leftmargin=0.3cm}
\begin{itemize}
  \item 1) \textbf{Matching:} The description correctly reflects the
    current trend of the TS (\emph{e.g.}, ``The value is rising
    steadily''). This text provides useful information and should
    be incorporated by the model.

  \item 2) \textbf{Contradicting:} The description
    contradicts the actual trend (\emph{e.g.}, ``The value is declining''
    when the series is rising). Injecting such text without filtering
    is expected to degrade forecast quality.

  \item 3) \textbf{Irrelevant:} The description is unrelated to the TS
    (\emph{e.g.}, a sentence about an unrelated domain). This text
    contains no useful signal and should ideally be ignored.
\end{itemize}

We construct a dataset ($N = 1{,}000$) in which all three text types coexist, with 70\% used for training, 10\% for validation, and 20\% for testing.
To illustrate the dataset, Figure~\ref{fig:toy_dataset} shows example TS segments paired with each type of textual description.


\textbf{B. Model and horizon.}
We use the Nonstationary Transformer~\citep{Liu2022NonstationaryTransformer} as the TS backbone with an input length of $L=8$ and a prediction horizon of $T=8$.

\textbf{C. Analyses.}
We examine the effect of the low-rank bottleneck from three perspectives:
\setlist[itemize]{leftmargin=0.3cm}
\begin{itemize}
  \item \textbf{Performance comparison} (Section~\ref{sec:appendix_toy_perf}): MSE per text type.
  \item \textbf{Representation analysis} (Section~\ref{sec:appendix_toy_repr}): Bottleneck activation norms and
    text-contribution ratios.
  \item \textbf{TS visualization} (Section~\ref{sec:appendix_toy_viz}): Predictions for matched vs.\ contradicting text.
\end{itemize}

\subsection{Analysis 1: Performance Comparison}
\label{sec:appendix_toy_perf}

Table~\ref{tab:toy_perf} reports the MSE on the
test split for case w/ bottleneck (CFA)
and w/o bottleneck (middle-additive) across three categories of text descriptions.

\begin{table}[!t]
\centering
\small
\caption{\textbf{Performance on the synthetic dataset.}
Results are reported for CFA w/ and w/o bottleneck across three categories of text descriptions.
}
\begin{NiceTabular}{l c c c}
\toprule
Text type & w/o bottleneck & w/ bottleneck  & Improv. (\%) \\
\midrule
\textbf{Matching}      & 0.1683 & \textbf{0.1477} & + 12.19 \\
\textbf{Contradicting} & 0.1635 &\textbf{0.1560} & + 4.59  \\
\textbf{Irrelevant}    & 0.1851 & \textbf{0.1480} & + 20.04 \\

\bottomrule
\end{NiceTabular}
\label{tab:toy_perf}
\end{table}

From Table~\ref{tab:toy_perf}, three observations follow as below:

\textbf{(i) CFA improves performance across all text types.}
CFA (w/ bottleneck) achieves lower MSE than the middle-additive baseline (w/o bottleneck)
for \textbf{matching}, \textbf{contradicting}, and \textbf{irrelevant}
descriptions, \textit{indicating that the low-rank bottleneck consistently
improves multimodal fusion}.

\textbf{(ii) CFA effectively suppresses irrelevant textual signals.}
For \textbf{irrelevant} descriptions, which contain no useful forecasting
information, the middle-additive baseline still injects the text into
the TS backbone, introducing noise into the representations.
In contrast, the \textit{low-rank bottleneck largely filters out such inputs},
yielding the largest MSE reduction ($+20.04\%$).

\textbf{(iii) CFA better distinguishes helpful and harmful text.}
Without the low-rank adapter, \textbf{contradicting} text surprisingly yields lower MSE
than \textbf{matching} text, suggesting that \textit{the baseline fails to
properly utilize the semantic content of the descriptions}.
In contrast, With the low-rank adapter, the expected ordering emerges,
where
\textbf{matching} descriptions are most beneficial, while
\textbf{contradicting} descriptions remain the most challenging
($\text{MSE}_{\text{Matching}} < \text{MSE}_{\text{Irrelevant}} < \text{MSE}_{\text{Contradicting}}$).

\subsection{Analysis 2: Representation Analysis}
\label{sec:appendix_toy_repr}

To investigate how CFA selectively filters text, we measure the \textbf{text-contribution ratio} at the output of the low-rank adapter. For each test sample, we compute $r = \frac{\|\mathbf{o}\|_2}{\|\bar{\mathbf{e}}\|_2}$,
where $\mathbf{o}\in\mathbb{R}^{d_{\text{model}}}$ is the adapter output added to the TS representation, and $\bar{\mathbf{e}}$ is the pooled text embedding before the adapter. Intuitively, $r$ quantifies \textit{how much of the original text signal survives the bottleneck}: higher $r$ indicates stronger injection, while lower $r$ indicates suppression.
In this analysis, we focus on the first encoder layer of the Nonstationary Transformer and report $r$ by text category (\textit{Matching} vs.\ \textit{Contradicting}).


\begin{figure}[!t]
  \centering
  \begin{adjustbox}{max width=\linewidth}
    \includegraphics[width=\textwidth]{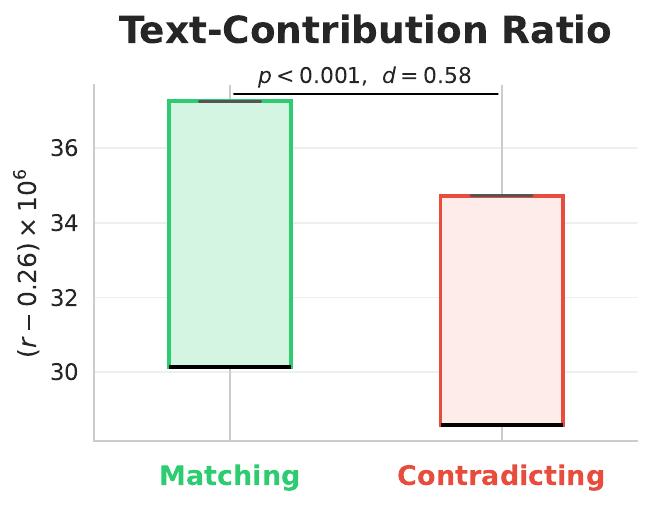}
  \end{adjustbox}
  \caption{Text-contribution ratio.}
  \label{fig:toy_repr}
\end{figure}


Table~\ref{tab:toy_repr} and Figure~\ref{fig:toy_repr} summarize the distribution of $r$ for \textbf{matching} and \textbf{contradicting} text,
where values are $(r - 0.25745)\times 10^{6}$ for readability.
 Matching text yields a higher mean value ($33.50$) than contradicting text ($31.56$), and the difference is statistically significant ($t=4.03$, $p<0.001$, Cohen's $d=0.58$). This suggests that the low-rank bottleneck \textit{suppresses contradictory text while preserving matching information}, providing controlled guidance to the TS backbone.

\subsection{Analysis 3: TS Visualization}
\label{sec:appendix_toy_viz}

Figure~\ref{fig:toy_viz} provides a qualitative comparison of the forecast
trajectories for a representative \text{matching} (left) and \text{contradicting} (right) test
sample from the AB dataset.


\textbf{Matching text.}
When the injected text correctly describes the trend, both w/~bottleneck and
w/o~bottleneck produce reasonable forecasts that follow the ground truth.
The CFA prediction (w/~bottleneck) tracks the ground truth slightly more closely,
suggesting that \textit{the low-rank bottleneck preserves useful information while
absorbing the helpful signal}.

\textbf{Contradicting text.}
The contrast is clearer for the contradicting-text sample,
where
the ground truth descends during the forecast horizon, while the injected text suggests an upward movement.
The baseline (w/o~bottleneck) is misled by this conflicting signal and its prediction
rises in the early forecast steps, deviating from the ground truth.
CFA (w/~bottleneck), in contrast, disregards the misleading text and predicts a
downward trajectory that closely follows the ground truth.
This qualitative contrast visualizes the filtering capability of the low-rank bottleneck, as the low-rank constraint \textit{prevents harmful text from corrupting the forecast even when the text signal is strongly contradictory}.

\section{Details of Constrained Fusion Strategies}
\label{app:constrained}

In this section, we provide formal descriptions and mathematical formulations for the three constrained fusion strategies introduced in Section~\ref{sec:constrained_fusion}.  
We adopt the notation defined in Section~\ref{sec:problem_formulation}: TS embeddings $\mathbf{Z}_\text{TS} = [\mathbf{z}_{\text{TS},1}, \ldots, \mathbf{z}_{\text{TS},L}]$, text embeddings $\mathbf{Z}_\text{Text} = [\mathbf{z}_{\text{Text},1}, \ldots, \mathbf{z}_{\text{Text},L}]$.
\subsection{Gating Mechanism}
The gating mechanism determines the relevance of textual information at each time step using a learned gate.  
Formally, let $\mathbf{z}_{\text{TS},t} \in \mathbb{R}^D$ denote the TS embedding at time step $t$, and $\mathbf{z}_{\text{Text},t} \in \mathbb{R}^D$ the text embedding.  
A gate $\mathbf{g}_t \in [0,1]^D$ is computed as:
\begin{equation*}
    \mathbf{g}_t = \sigma(\mathbf{W}_g [\mathbf{z}_{\text{TS},t}; \mathbf{z}_{\text{Text},t}] + \mathbf{b}_g),
\end{equation*}
where $\sigma$ is the sigmoid function, $[\cdot;\cdot]$ denotes concatenation, and $\mathbf{W}_g, \mathbf{b}_g$ are learnable parameters.  
The fused embedding is then:
\begin{equation*}
    \mathbf{z}_{\text{fused},t} = \mathbf{z}_{\text{TS},t} + \mathbf{g}_t \odot \mathbf{z}_{\text{Text},t}.
\end{equation*}

\subsection{FiLM (Feature-wise Linear Modulation)}
FiLM modulates TS embeddings based on text embeddings by applying feature-wise scaling and shifting as:
\begin{equation*}
    \mathbf{z}_{\text{fused},t} = \boldsymbol{\gamma}(\mathbf{z}_{\text{Text},t}) \odot \mathbf{z}_{\text{TS},t} + \boldsymbol{\beta}(\mathbf{z}_{\text{Text},t}),
\end{equation*}
where $\boldsymbol{\gamma}(\cdot), \boldsymbol{\beta}(\cdot)$ are learned functions (e.g., MLPs) that generate scaling and shifting parameters from the text embedding.  
This preserves the temporal structure of TS embeddings while modulating each feature dimension according to textual context.

\subsection{Orthogonal Fusion}
Orthogonal fusion projects the text embedding onto the orthogonal complement of the TS embedding to avoid overwriting temporal information:
\begin{equation*}
    \mathbf{z}_{\text{Text},t}^\perp = \mathbf{z}_{\text{Text},t} - \frac{\mathbf{z}_{\text{TS},t}^\top \mathbf{z}_{\text{Text},t}}{\|\mathbf{z}_{\text{TS},t}\|^2} \mathbf{z}_{\text{TS},t},
\end{equation*}
\begin{equation*}
    \mathbf{z}_{\text{fused},t} = \mathbf{z}_{\text{TS},t} + \mathbf{z}_{\text{Text},t}^\perp.
\end{equation*}
This ensures that only components of the text embedding orthogonal to the TS embedding are integrated, preserving the original temporal information.

\subsection{Controlled Fusion Adapter (CFA)}
Controlled Fusion Adapter (CFA) integrates textual information into TS embeddings through a lightweight residual adapter with a low-dimensional bottleneck.
Unlike direct additive or multiplicative fusion, CFA first transforms the text embedding into a compact latent space and then projects it back to the original dimension before residual addition.
This design preserves the TS backbone representation while allowing text to provide directional guidance with minimal parameter overhead.

Formally, for each time step $t$, the text embedding $\mathbf{z}_{\text{Text},t} \in \mathbb{R}^D$ is first projected to a reduced bottleneck space:

\begin{equation*}
\mathbf{z}_{\text{Adapter},t}
= \mathbf{W}_{\text{down}} \mathbf{z}_{\text{Text},t}
\in \mathbb{R}^{D/r},
\end{equation*}
where $r$ is the reduction ratio controlling parameter efficiency.
The intermediate representation is then normalized and activated:

\begin{equation*}
\mathbf{z}_{\text{Adapter},t}^{\prime}
= \phi\big(\text{LayerNorm}(\mathbf{z}_{\text{Adapter},t})\big),
\end{equation*}

where $\phi(\cdot)$ denotes a non-linear activation function (e.g., ReLU).
Finally, the representation is projected back to the original dimension and added as a residual to the TS embedding:

\begin{equation*}
\mathbf{z}_{\text{fused},t}
= \mathbf{z}_{\text{TS},t}
+ \mathbf{W}_{\text{up}} \mathbf{z}_{\text{Adapter},t}^{\prime}.
\end{equation*}

Here, $\mathbf{W}_{\text{down}} \in \mathbb{R}^{\frac{D}{r} \times D}$ and $\mathbf{W}_{\text{up}} \in \mathbb{R}^{D \times \frac{D}{r}}$ are learnable parameters.
The residual structure ensures that the original TS embedding is preserved, while the bottleneck controls the magnitude and complexity of textual influence.
In practice, initializing $\mathbf{W}_{\text{up}}$ near zero helps stabilize early training by preventing excessive perturbation of the TS backbone.


\section{Theoretical Perspective on Low-Rank Text Fusion}
\label{app:theory}
CFA injects textual signals into TS embeddings through a low-rank adapter.
Following the formulation in Section~\ref{sec:CFA}, the injected signal is given by
\begin{align}
\mathbf{z}_{\text{Adapter},t}'' =
\mathbf{W}_{\text{up}}\,\phi\!\bigl(\mathrm{LayerNorm}(\mathbf{W}_{\text{down}}\mathbf{z}_{\text{Text},t})\bigr), 
\end{align}
which is added to the TS embedding as
\begin{align}
\tilde{\mathbf{z}}_{\text{TS},t}
=
\mathbf{z}_{\text{TS},t}
+
\mathbf{z}_{\text{Adapter},t}'' .
\end{align}


Since $\mathbf{W}_{\text{up}} \in \mathbb{R}^{D \times D/r}$,
its column space has dimension at most $D/r$.
For any intermediate vector $\mathbf{v} \in \mathbb{R}^{D/r}$
obtained by applying the down-projection, LayerNorm, and non-linear activation $\phi$,
the final output $\mathbf{W}_{\text{up}}\mathbf{v}$ must lie in the column space of $\mathbf{W}_{\text{up}}$.
Therefore the injected textual signal lies in a subspace of
$\mathbb{R}^{D}$ whose dimension is at most $D/r$.

This low-rank constraint \textit{restricts the directions in which textual signals influence TS embeddings}. As a result, the textual information modifies TS representations only \textit{within a low-dimensional subspace}. 
This 
reduces the capacity of textual signals and suppresses irrelevant information during fusion.

\section{Temporal Attribution for Multimodal TS Forecasting}
\label{app:temporal_attribution}

To analyze which past time steps the model relies on when generating forecasts, we assign an importance score to each of the $L$ input time steps. This enables direct comparison of temporal focus across different fusion strategies.

\textbf{Problem setup.}
Let the encoder input be $\mathbf{x} \in \mathbb{R}^{B \times L \times D}$, where $B$ denotes the batch size, $L$ the input horizon, and $D$ the number of channels. 
The forecasting model predicts $\hat{\mathbf{y}}$ over the output horizon, and we define the loss as mean squared error
\[
\mathcal{L} = \mathrm{MSE}(\hat{\mathbf{y}}, \mathbf{y}).
\]
Our goal is to quantify the contribution of each input time step $t \in \{0, \ldots, L-1\}$ to this loss.

\textbf{Gradient-based temporal importance.}
We adopt a Gradient $\times$ Input attribution scheme. 
For a single sample (omitting the batch index), let $x_t^{(d)}$ denote the input value at time step $t$ and channel $d$. 
The gradient of the loss with respect to the input is
\[
g_t^{(d)} = \frac{\partial \mathcal{L}}{\partial x_t^{(d)}}.
\]
The element-wise attribution is defined as
\[
a_t^{(d)} = g_t^{(d)} \cdot x_t^{(d)}.
\]
We aggregate over channels to obtain one importance value per time step:
\[
I_t = \sum_{d=1}^{D} \left| a_t^{(d)} \right|
= \sum_{d=1}^{D} \left| \frac{\partial \mathcal{L}}{\partial x_t^{(d)}} \cdot x_t^{(d)} \right|.
\]
This yields an importance vector $\mathbf{I} = (I_0, \ldots, I_{L-1}) \in \mathbb{R}^{L}$. 
When the input horizon is $L=8$, we therefore obtain exactly 8 importance values, one for each input time step.

For visualization and comparison across samples, we optionally normalize the scores:
\[
\tilde{I}_t = \frac{I_t}{\sum_{t'=0}^{L-1} I_{t'} + \epsilon},
\]
where $\epsilon > 0$ ensures numerical stability.

\textbf{Implementation details}
We compute gradients with respect to the encoder input by cloning the input tensor and enabling gradient tracking on the clone. 
After a forward pass and backpropagation of the MSE loss, we extract $\partial \mathcal{L} / \partial \mathbf{x}$ and compute the element-wise product with the original input values. 
All attribution scores are computed under the same forecasting setting used for evaluation.

For models that do not employ attention mechanisms, such as TiDE, the temporal attribution is derived solely from the gradient-based formulation above. 
Thus, the number of importance values always equals the input horizon $L$, ensuring consistent interpretation across backbones.

\refstepcounter{section}     
\begin{table}[!t]
\centering
\caption{Average performance (MAE, MSE) by CFA reduction rank $r$.}
\label{tab:ablation_cfa_overall}
\begin{adjustbox}{max width=0.95\linewidth}
\begin{NiceTabular}{l c c c c c}
\toprule
Metric & $r=2$ & $r=4$ & $r=8$ & $r=16$ & $r=32$ \\
\midrule
MAE & 1.031 & \bfseries 1.025 & 1.029 & 1.035 & 1.030 \\
SE & 13.326 & \bfseries 13.290 & 13.334 & 13.374 & 13.339 \\
\bottomrule
\end{NiceTabular}
\end{adjustbox}
\label{tbl:cfa_red_avg}
\end{table}
\addtocounter{section}{-1}   
\section{Sensitivity Analysis}
\label{app:sensitivity_analysis}
We examine the robustness of the CFA to the reduction ratio $r$ across datasets and TS models.
The results in Table~\ref{tbl:cfa_red_avg} show that performance remains stable for different values of $r$. 
Although $r=4$ achieves the best average performance in several cases, we adopt $r=8$ in our experiments considering both performance and computational efficiency.

\section{Comparison with Other Methods}
\label{app:compare_others}

Table~\ref{app_tbl:others} presents a comparison of our method against other TS models, covering both unimodal TS models and architecture-specific multimodal TS models. 
We use eight Time-MMD multimodal datasets~\cite{liu2024time}, excluding the Environment dataset due to reproducibility issues with the baseline methods.\footnote{The reproducibility issue with Environment arises in MM-TSF~\cite{liu2024time}, which corresponds to the last-additive setting in our experiments; the reported results differ from those obtained in our runs.}
The results are based on SpecTF~\cite{SpecTF2602}, which is concurrent with our work (2026.02). 
We apply our approach to three representative TS backbones: a Transformer-based model (Nonstationary Transformer~\cite{Liu2022NonstationaryTransformer}), a Linear/MLP-based model (TiDE~\cite{Das2023TiDE}), and other architectures (Koopa~\cite{Koopa2023}).

\section{Full Results of Main Experiments}
\label{app:full_results}

We report the full results of our main experiments across all 14 TS backbones, paired with each of the 4 text models (BERT, GPT-2, Llama-3, Doc2Vec).

\clearpage
\setcounter{section}{9}      
\refstepcounter{section}     
\setcounter{table}{1}        


\begin{table*}[!t]
\centering
\label{tab:ablation_cfa_model}
\begin{minipage}{\linewidth}
\centering
\caption{Average MAE per model by CFA reduction ratio $r$.}
\begin{adjustbox}{max width=\linewidth}
\begin{NiceTabular}{l c c c c c}
\toprule
Model & $r=2$ & $r=4$ & $r=8$ & $r=16$ & $r=32$ \\
\midrule
Autoformer~\cite{wu2021autoformer}& 1.026 & 1.021 & 1.031 & 1.039 & \bfseries 1.018 \\
Crossformer~\cite{zhang2023crossformer}& 1.125 & 1.134 & 1.127 & 1.137 & \bfseries 1.118 \\
DLinear~\cite{Zeng2022DLinear} &  \bfseries 0.969 & \bfseries 0.969 & \bfseries 0.969 & \bfseries 0.969 & 0.970 \\
FEDformer~\cite{zhou2022fedformer} &  0.967 & \bfseries 0.963 & 0.970 & 0.978 & 0.981 \\
FiLM~\cite{Zhou2022FiLM} & \bfseries 0.963 & \bfseries 0.963 & \bfseries 0.963 & \bfseries 0.963 & 0.963 \\
Informer~\cite{zhou2021informer} & 1.217 & \bfseries 1.190 & 1.211 & 1.217 & 1.208 \\
Koopa~\cite{Koopa2023} &  \bfseries 0.944 & \bfseries 0.944 & \bfseries 0.944 & \bfseries 0.944 & 0.945 \\
Nonstationary Transformer~\cite{Liu2022NonstationaryTransformer}& 0.904 & 0.909 & \bfseries 0.899 & 0.916 & 0.910 \\
PatchTST~\cite{Nie2023PatchTST} & 0.948 & \bfseries 0.943 & 0.948 & 0.948 & 0.951 \\
Reformer~\cite{kitaev2020reformer} & 1.151 & \bfseries 1.136 & 1.167 & 1.179 & 1.162 \\
TSMixer~\cite{Chen2023TSMixer} & \bfseries 1.123 & \bfseries 1.123 & \bfseries 1.123 & \bfseries 1.123 & \bfseries 1.123 \\
TiDE~\cite{Das2023TiDE} &0.988 & 0.979 & \bfseries 0.974 & 0.981 & 0.986 \\
Transformer~\cite{vaswani2017attention}& 1.149 & 1.134 & \bfseries 1.128 & 1.162 & 1.145 \\
iTransformer~\cite{Liu2023iTransformer} & 0.953 & 0.943 & 0.950 & \bfseries 0.940 & 0.944 \\
\bottomrule
\end{NiceTabular}
\end{adjustbox}
\end{minipage}

\vspace{20pt}

\begin{minipage}{\linewidth}
\centering
\caption{Average MSE per model by CFA reduction ratio $r$.}
\begin{adjustbox}{max width=\linewidth}
\begin{NiceTabular}{l c c c c c}
\toprule
Model & $r=2$ & $r=4$ & $r=8$ & $r=16$ & $r=32$ \\
\midrule
Autoformer~\cite{wu2021autoformer}& 12.947 & \bfseries 12.744 & 13.107 & 13.203 & 12.805 \\
Crossformer~\cite{zhang2023crossformer}& 14.302 & 14.345 & 14.135 & 14.291 & \bfseries 14.088 \\
DLinear~\cite{Zeng2022DLinear} &  \bfseries 12.613 & \bfseries 12.613 & \bfseries 12.613 & \bfseries 12.613 & \bfseries 12.613 \\
FEDformer~\cite{zhou2022fedformer} &  \bfseries 12.597 & 12.732 & 12.748 & 12.955 & 12.887 \\
FiLM~\cite{Zhou2022FiLM} & \bfseries 12.766 & \bfseries 12.766 & \bfseries 12.766 & \bfseries 12.766 & 12.766 \\
Informer~\cite{zhou2021informer} & 14.829 & 14.851 & 14.861 & 14.897 & \bfseries 14.813 \\
Koopa~\cite{Koopa2023} &  \bfseries 12.420 & \bfseries 12.420 & \bfseries 12.420 & \bfseries 12.420 & 12.421 \\
Nonstationary Transformer~\cite{Liu2022NonstationaryTransformer}& 12.247 & 12.208 & \bfseries 12.185 & 12.276 & 12.274 \\
PatchTST~\cite{Nie2023PatchTST} & 12.615 & \bfseries 12.580 & 12.632 & 12.613 & 12.584 \\
Reformer~\cite{kitaev2020reformer} & \bfseries 14.038 & 14.165 & 14.387 & 14.447 & 14.394 \\
TSMixer~\cite{Chen2023TSMixer} & \bfseries 14.693 & \bfseries 14.693 & \bfseries 14.693 & \bfseries 14.693 & \bfseries 14.693 \\
TiDE~\cite{Das2023TiDE} &12.974 & \bfseries 12.756 & 12.779 & 12.797 & 13.051 \\
Transformer~\cite{vaswani2017attention}& 14.797 & 14.714 & 14.733 & \bfseries 14.691 & 14.698 \\
iTransformer~\cite{Liu2023iTransformer} & 12.727 & \bfseries 12.471 & 12.621 & 12.568 & 12.654 \\
\bottomrule
\end{NiceTabular}
\end{adjustbox}
\end{minipage}
\end{table*}

\begin{table*}[!t]
\centering
\label{tab:ablation_cfa_dataset}
\begin{minipage}{\linewidth}
\centering
\caption{Average MAE per dataset by CFA reduction ratio $r$.}
\begin{adjustbox}{max width=\linewidth}
\begin{NiceTabular}{l c c c c c}
\toprule
Dataset & $r=2$ & $r=4$ & $r=8$ & $r=16$ & $r=32$ \\
\midrule
Agriculture & 0.2534 & \bfseries 0.2452 & 0.2508 & 0.2515 & 0.2560 \\
Climate & 0.8680 & 0.8655 & \bfseries 0.8648 & 0.8669 & 0.8666 \\
Economy & 0.2679 & \bfseries 0.2364 & 0.2610 & 0.2825 & 0.2518 \\
Energy & 0.3927 & \bfseries 0.3894 & 0.3908 & 0.3896 & 0.3939 \\
Environment & \bfseries 0.5156 & 0.5192 & 0.5190 & 0.5188 & 0.5190 \\
Public & 0.7976 & 0.7941 & \bfseries 0.7931 & 0.8001 & 0.7946 \\
Security & 5.4168 & \bfseries 5.4129 & 5.4218 & 5.4441 & 5.4309 \\
SocialGood & 0.4584 & 0.4577 & \bfseries 0.4532 & 0.4642 & 0.4545 \\
Traffic & 0.3046 & 0.3054 & 0.3041 & \bfseries 0.3019 & 0.3054 \\
\bottomrule
\end{NiceTabular}
\end{adjustbox}
\end{minipage}

\vspace{20pt}

\begin{minipage}{\linewidth}
\centering
\caption{Average MSE per dataset by CFA reduction ratio $r$.}
\begin{adjustbox}{max width=\linewidth}
\begin{NiceTabular}{l c c c c c}
\toprule
Dataset & $r=2$ & $r=4$ & $r=8$ & $r=16$ & $r=32$ \\
\midrule
Agriculture & 0.1483 & \bfseries 0.1446 & 0.1473 & 0.1501 & 0.1526 \\
Climate & 1.1596 & \bfseries 1.1520 & 1.1537 & 1.1576 & 1.1540 \\
Economy & 0.1602 & \bfseries 0.1278 & 0.1540 & 0.1790 & 0.1390 \\
Energy & 0.2813 & \bfseries 0.2781 & 0.2800 & 0.2794 & 0.2838 \\
Environment & \bfseries 0.4836 & 0.4877 & 0.4875 & 0.4856 & 0.4848 \\
Public & 1.4135 & 1.4161 & \bfseries 1.4055 & 1.4173 & 1.4062 \\
Security & 115.1568 & \bfseries 114.8712 & 115.2570 & 115.5611 & 115.3047 \\
SocialGood & 0.9195 & 0.9183 & 0.9121 & 0.9215 & \bfseries 0.9115 \\
Traffic & 0.2114 & 0.2125 & 0.2112 & \bfseries 0.2108 & 0.2113 \\
\bottomrule
\end{NiceTabular}
\end{adjustbox}
\end{minipage}

\end{table*}

\refstepcounter{section}     
\newcommand{\best}[1]{\textbf{\textcolor{red}{#1}}}
\newcommand{\second}[1]{\underline{\textcolor{blue}{#1}}}

\begin{table*}[!t]
\caption{\textbf{Comparison with other methods, including architecture-specific multimodal TS forecasting models.} \best{Red bold}: best, \second{blue underline}: second best.}
  \centering
  \begin{adjustbox}{width=1.00\linewidth}
\begin{tabular}{cc|c|c|c|c|c|c|c||c|c|c}
    \toprule 
    \multicolumn{1}{c}{\multirow{2.5}{*}{Dataset}} &
    \multicolumn{1}{c}{\multirow{2.5}{*}{$H$}} &
    \multirow{2.5}{*}{SpecTF~\cite{SpecTF2602}} &
    \multirow{2.5}{*}{TaTS~\cite{li2025language}} &
    \multirow{2.5}{*}{MM-TSF~\cite{liu2024time}} &
    \multirow{2.5}{*}{TimeXer~\cite{wang2024timexer}} &
    \multirow{2.5}{*}{FreTS~\cite{yi2023frequency}} &
    \multirow{2.5}{*}{Time-LLM~\cite{jin2023time}} &
    \multirow{2.5}{*}{ChatTime~\cite{wang2025chattime}} &
    \multicolumn{3}{c}{\textbf{CFA} (Ours)} \\
    \cmidrule(lr){10-12}
    \multicolumn{2}{c|}{} & & & & & & & & Transformer~\cite{Liu2022NonstationaryTransformer} & + Linear/MLP~\cite{Das2023TiDE} & Others~\cite{Koopa2023} \\
    \midrule
    \multirow{5.5}{*}{\scalebox{1.0}{Agriculture}}
    & 6  & 0.064 & 0.075 & 0.084 & 0.088 & 0.082 & 0.092 & 0.193 & \best{0.053} & 0.060 & \second{0.056} \\
    & 8  & 0.084 & 0.097 & 0.112 & 0.109 & 0.101 & 0.151 & 0.264 & \best{0.069} & 0.082 & \second{0.076} \\
    & 10 & 0.110 & 0.127 & 0.133 & 0.144 & 0.125 & 0.176 & 0.345 & \best{0.091} & 0.103 & \second{0.100} \\
    & 12 & 0.152 & 0.167 & 0.165 & 0.187 & 0.172 & 0.210 & 0.418 & \best{0.116} & 0.135 & \second{0.128} \\
    \cmidrule(lr){2-12}
    & Avg & 0.103 & 0.112 & 0.123 & 0.132 & 0.120 & 0.157 & 0.305 & \best{0.082} & 0.095 & \second{0.090} \\
    \midrule
    \multirow{5.5}{*}{\scalebox{1.0}{Climate}}
    & 6  & \second{0.904} & 0.990 & 1.074 & 1.027 & \best{0.286} & 1.126 & 1.490 & 1.073 & 1.258 & 1.253 \\
\end{tabular}
  \end{adjustbox}
  \label{app_tbl:others}
  \end{table*}

\refstepcounter{section}     
\begin{table*}[!htbp]
\caption{Comparison of multimodal fusion strategies with \textbf{Nonstationary Transformer} (+ \textbf{BERT}).}
\centering
\begin{adjustbox}{width=1.00\linewidth}
\begin{tabular}{c|ll|c|ccc|ccc|cccc}
\toprule
\mr{2.5}{\shortstack{TS \\ Model}} & \mr{2.5}{Dataset} & \mr{2.5}{$H$} & \mr{2.5}{Unimodal} & \multicolumn{3}{c}{\textbf{Additive}} & \multicolumn{3}{c}{\textbf{Concat}} & \multicolumn{4}{c}{\textbf{Constrained}} \\
\cmidrule(lr){5-7}\cmidrule(lr){8-10}\cmidrule(lr){11-14}
& & & & First & Middle & Last & First & Middle & Last & Orthogonal & FiLM & Gating & \textbf{CFA } \\
\midrule
\mr{40}{\rotatebox{90}{\textbf{Nonstationary Transformer}}} & \mr{4}{Agriculture} & \hz{6} & 0.051 & \blue{0.053} & \blue{0.052} & \blue{0.052} & \blue{0.060} & \blue{0.055} & \blue{3.611} & \blue{0.051} & \blue{0.054} & \blue{0.056} & \blue{0.053} \\
 &  & \hz{8} & 0.071 & \textbf{\red{0.069}} & \textbf{\red{0.071}} & \textbf{\red{0.067}} & \blue{0.072} & \textbf{\red{0.071}} & \blue{1.243} & \blue{0.072} & \blue{0.071} & \blue{0.073} & \textbf{\red{0.069}} \\
 &  & \hz{10} & 0.092 & \blue{0.100} & \textbf{\red{0.089}} & \textbf{\red{0.088}} & \blue{0.098} & \textbf{\red{0.090}} & \blue{0.398} & \textbf{\red{0.091}} & \textbf{\red{0.090}} & \textbf{\red{0.091}} & \textbf{\red{0.091}} \\
 &  & \hz{12} & 0.116 & \blue{0.121} & \blue{0.119} & \textbf{\red{0.109}} & \blue{0.121} & \blue{0.120} & \blue{0.353} & \blue{0.118} & \blue{0.128} & \blue{0.119} & \blue{0.116} \\
\cmidrule(lr){2-14}
 & \mr{4}{Climate} & \hz{6} & 1.205 & \blue{1.305} & \blue{1.310} & \textbf{\red{1.172}} & \textbf{\red{1.124}} & \blue{1.235} & \blue{1.233} & \blue{1.289} & \textbf{\red{1.185}} & \blue{1.272} & \textbf{\red{1.073}} \\
 &  & \hz{8} & 1.240 & \textbf{\red{1.225}} & \textbf{\red{1.124}} & \textbf{\red{1.219}} & \textbf{\red{1.231}} & \blue{1.300} & \textbf{\red{1.064}} & \blue{1.289} & \textbf{\red{1.238}} & \textbf{\red{1.202}} & \blue{1.250} \\
\bottomrule
\end{tabular}
\end{adjustbox}
\end{table*}

\begin{table*}[!htbp]
\caption{Comparison of multimodal fusion strategies with \textbf{Koopa} (+ \textbf{BERT}).}
\centering
\begin{adjustbox}{width=1.00\linewidth}
\begin{tabular}{c|ll|c|ccc|ccc|cccc}
\toprule
\mr{2.5}{\shortstack{TS \\ Model}} & \mr{2.5}{Dataset} & \mr{2.5}{$H$} & \mr{2.5}{Unimodal} & \multicolumn{3}{c}{\textbf{Additive}} & \multicolumn{3}{c}{\textbf{Concat}} & \multicolumn{4}{c}{\textbf{Constrained}} \\
\cmidrule(lr){5-7}\cmidrule(lr){8-10}\cmidrule(lr){11-14}
& & & & First & Middle & Last & First & Middle & Last & Orthogonal & FiLM & Gating & \textbf{CFA } \\
\midrule
\mr{40}{\rotatebox{90}{\textbf{Koopa}}} & \mr{4}{Agriculture} & \hz{6} & 0.059 & \textbf{\red{0.056}} & \textbf{\red{0.058}} & \textbf{\red{0.058}} & \blue{5.693} & \blue{3.044} & \blue{3.044} & \blue{0.060} & \blue{0.067} & \textbf{\red{0.055}} & \textbf{\red{0.056}} \\
 &  & \hz{8} & 0.077 & \blue{0.078} & \textbf{\red{0.072}} & \textbf{\red{0.072}} & \blue{0.173} & \blue{0.108} & \blue{0.108} & \textbf{\red{0.076}} & \blue{0.092} & \textbf{\red{0.073}} & \textbf{\red{0.076}} \\
 &  & \hz{10} & 0.099 & \textbf{\red{0.091}} & \textbf{\red{0.090}} & \textbf{\red{0.090}} & \blue{0.426} & \blue{0.624} & \blue{0.624} & \blue{0.100} & \blue{0.127} & \textbf{\red{0.091}} & \blue{0.100} \\
 &  & \hz{12} & 0.129 & \textbf{\red{0.119}} & \textbf{\red{0.117}} & \textbf{\red{0.117}} & \blue{1.032} & \blue{1.233} & \blue{1.233} & \blue{0.131} & \blue{0.155} & \textbf{\red{0.117}} & \textbf{\red{0.128}} \\
\cmidrule(lr){2-14}
 & \mr{4}{Climate} & \hz{6} & 1.251 & \blue{6.593} & \blue{1.258} & \blue{1.258} & \blue{4.140} & \textbf{\red{1.219}} & \textbf{\red{1.219}} & \blue{1.277} & \textbf{\red{1.069}} & \blue{1.253} & \blue{1.253} \\
 &  & \hz{8} & 1.264 & \blue{1.499} & \textbf{\red{1.261}} & \textbf{\red{1.261}} & \textbf{\red{1.192}} & \textbf{\red{1.122}} & \textbf{\red{1.122}} & \blue{1.291} & \textbf{\red{1.057}} & \textbf{\red{1.262}} & \textbf{\red{1.258}} \\
\bottomrule
\end{tabular}
\end{adjustbox}
\end{table*}

\begin{table*}[!htbp]
\caption{Comparison of multimodal fusion strategies with \textbf{DLinear} (+ \textbf{BERT}).}
\centering
\begin{adjustbox}{width=1.00\linewidth}
\begin{tabular}{c|ll|c|ccc|ccc|cccc}
\toprule
\mr{2.5}{\shortstack{TS \\ Model}} & \mr{2.5}{Dataset} & \mr{2.5}{$H$} & \mr{2.5}{Unimodal} & \multicolumn{3}{c}{\textbf{Additive}} & \multicolumn{3}{c}{\textbf{Concat}} & \multicolumn{4}{c}{\textbf{Constrained}} \\
\cmidrule(lr){5-7}\cmidrule(lr){8-10}\cmidrule(lr){11-14}
& & & & First & Middle & Last & First & Middle & Last & Orthogonal & FiLM & Gating & \textbf{CFA } \\
\midrule
\mr{40}{\rotatebox{90}{\textbf{DLinear}}} & \mr{4}{Agriculture} & \hz{6} & 0.114 & \blue{0.127} & \blue{0.134} & \blue{0.132} & \blue{0.575} & \blue{0.320} & \blue{0.454} & \blue{0.135} & \blue{0.233} & \blue{0.133} & \textbf{\red{0.085}} \\
 &  & \hz{8} & 0.324 & \textbf{\red{0.182}} & \textbf{\red{0.185}} & \textbf{\red{0.181}} & \blue{0.973} & \blue{2.019} & \blue{1.390} & \textbf{\red{0.206}} & \textbf{\red{0.217}} & \textbf{\red{0.223}} & \textbf{\red{0.099}} \\
 &  & \hz{10} & 0.163 & \blue{0.203} & \blue{0.204} & \blue{0.204} & \blue{0.481} & \blue{0.419} & \blue{0.199} & \blue{0.178} & \blue{0.208} & \blue{0.205} & \textbf{\red{0.127}} \\
 &  & \hz{12} & 0.217 & \blue{0.239} & \blue{0.240} & \blue{0.242} & \blue{0.428} & \blue{0.350} & \blue{0.326} & \textbf{\red{0.162}} & \blue{0.410} & \textbf{\red{0.180}} & \textbf{\red{0.183}} \\
\cmidrule(lr){2-14}
 & \mr{4}{Climate} & \hz{6} & 1.198 & \blue{1.277} & \blue{1.229} & \blue{1.203} & \blue{1.308} & \blue{1.252} & \blue{1.219} & \blue{1.306} & \textbf{\red{1.084}} & \blue{1.241} & \textbf{\red{1.090}} \\
 &  & \hz{8} & 1.364 & \textbf{\red{1.318}} & \textbf{\red{1.267}} & \textbf{\red{1.247}} & \textbf{\red{1.250}} & \textbf{\red{1.064}} & \textbf{\red{1.162}} & \textbf{\red{1.328}} & \textbf{\red{1.071}} & \textbf{\red{1.241}} & \textbf{\red{1.090}} \\
\bottomrule
\end{tabular}
\end{adjustbox}
\end{table*}

\begin{table*}[!htbp]
\caption{Comparison of multimodal fusion strategies with \textbf{iTransformer} (+ \textbf{BERT}).}
\centering
\begin{adjustbox}{width=1.00\linewidth}
\begin{tabular}{c|ll|c|ccc|ccc|cccc}
\toprule
\mr{2.5}{\shortstack{TS \\ Model}} & \mr{2.5}{Dataset} & \mr{2.5}{$H$} & \mr{2.5}{Unimodal} & \multicolumn{3}{c}{\textbf{Additive}} & \multicolumn{3}{c}{\textbf{Concat}} & \multicolumn{4}{c}{\textbf{Constrained}} \\
\cmidrule(lr){5-7}\cmidrule(lr){8-10}\cmidrule(lr){11-14}
& & & & First & Middle & Last & First & Middle & Last & Orthogonal & FiLM & Gating & \textbf{CFA } \\
\midrule
\mr{40}{\rotatebox{90}{\textbf{iTransformer}}} & \mr{4}{Agriculture} & \hz{6} & 0.059 & \blue{0.060} & \blue{0.061} & \textbf{\red{0.058}} & \blue{0.062} & \textbf{\red{0.058}} & \blue{1.273} & \textbf{\red{0.059}} & \blue{0.060} & \blue{0.059} & \textbf{\red{0.059}} \\
 &  & \hz{8} & 0.076 & \blue{0.078} & \blue{0.081} & \textbf{\red{0.072}} & \textbf{\red{0.076}} & \blue{0.077} & \blue{1.097} & \blue{0.081} & \blue{0.080} & \blue{0.079} & \textbf{\red{0.076}} \\
 &  & \hz{10} & 0.097 & \blue{0.103} & \blue{0.101} & \textbf{\red{0.090}} & \blue{0.097} & \blue{0.099} & \blue{0.347} & \blue{0.101} & \blue{0.103} & \blue{0.098} & \blue{0.098} \\
 &  & \hz{12} & 0.128 & \blue{0.129} & \blue{0.129} & \textbf{\red{0.118}} & \blue{0.128} & \blue{0.131} & \blue{0.390} & \blue{0.128} & \blue{0.131} & \blue{0.129} & \textbf{\red{0.127}} \\
\cmidrule(lr){2-14}
 & \mr{4}{Climate} & \hz{6} & 1.100 & \blue{1.189} & \blue{1.140} & \blue{1.128} & \blue{1.130} & \blue{1.141} & \blue{1.326} & \blue{1.141} & \blue{1.130} & \blue{1.150} & \blue{1.115} \\
 &  & \hz{8} & 1.178 & \blue{1.204} & \blue{1.203} & \blue{1.193} & \textbf{\red{1.162}} & \blue{1.193} & \textbf{\red{1.158}} & \blue{1.202} & \blue{1.231} & \blue{1.197} & \blue{1.227} \\
\bottomrule
\end{tabular}
\end{adjustbox}
\end{table*}
\FloatBarrier
\begin{table*}[!htbp]
\caption{Comparison of multimodal fusion strategies with \textbf{PatchTST} (+ \textbf{BERT}).}
\centering
\begin{adjustbox}{width=1.00\linewidth}
\begin{tabular}{c|ll|c|ccc|ccc|cccc}
\toprule
\mr{2.5}{\shortstack{TS \\ Model}} & \mr{2.5}{Dataset} & \mr{2.5}{$H$} & \mr{2.5}{Unimodal} & \multicolumn{3}{c}{\textbf{Additive}} & \multicolumn{3}{c}{\textbf{Concat}} & \multicolumn{4}{c}{\textbf{Constrained}} \\
\cmidrule(lr){5-7}\cmidrule(lr){8-10}\cmidrule(lr){11-14}
& & & & First & Middle & Last & First & Middle & Last & Orthogonal & FiLM & Gating & \textbf{CFA } \\
\midrule
\mr{40}{\rotatebox{90}{\textbf{PatchTST}}} & \mr{4}{Agriculture} & \hz{6} & 0.058 & \blue{0.059} & \blue{0.060} & \textbf{\red{0.056}} & \blue{0.059} & \blue{0.060} & \blue{1.357} & \blue{0.059} & \blue{0.060} & \blue{0.060} & \blue{0.058} \\
 &  & \hz{8} & 0.078 & \blue{0.078} & \blue{0.080} & \textbf{\red{0.074}} & \blue{0.078} & \blue{0.079} & \blue{0.816} & \textbf{\red{0.078}} & \blue{0.079} & \blue{0.079} & \textbf{\red{0.077}} \\
 &  & \hz{10} & 0.100 & \textbf{\red{0.098}} & \blue{0.101} & \textbf{\red{0.093}} & \textbf{\red{0.098}} & \blue{0.101} & \blue{0.382} & \blue{0.101} & \blue{0.101} & \blue{0.101} & \blue{0.100} \\
 &  & \hz{12} & 0.132 & \textbf{\red{0.129}} & \textbf{\red{0.129}} & \textbf{\red{0.122}} & \textbf{\red{0.129}} & \textbf{\red{0.128}} & \blue{0.371} & \textbf{\red{0.129}} & \textbf{\red{0.131}} & \textbf{\red{0.127}} & \textbf{\red{0.129}} \\
\cmidrule(lr){2-14}
 & \mr{4}{Climate} & \hz{6} & 1.257 & \textbf{\red{1.234}} & \textbf{\red{1.238}} & \textbf{\red{1.235}} & \textbf{\red{1.234}} & \textbf{\red{1.253}} & \blue{1.332} & \textbf{\red{1.238}} & \textbf{\red{1.248}} & \textbf{\red{1.215}} & \textbf{\red{1.230}} \\
 &  & \hz{8} & 1.265 & \textbf{\red{1.255}} & \textbf{\red{1.224}} & \textbf{\red{1.257}} & \textbf{\red{1.255}} & \textbf{\red{1.249}} & \textbf{\red{1.183}} & \textbf{\red{1.224}} & \textbf{\red{1.252}} & \textbf{\red{1.233}} & \textbf{\red{1.236}} \\
\bottomrule
\end{tabular}
\end{adjustbox}
\end{table*}

\begin{table*}[!htbp]
\caption{Comparison of multimodal fusion strategies with \textbf{FEDformer} (+ \textbf{BERT}).}
\centering
\begin{adjustbox}{width=1.00\linewidth}
\begin{tabular}{c|ll|c|ccc|ccc|cccc}
\toprule
\mr{2.5}{\shortstack{TS \\ Model}} & \mr{2.5}{Dataset} & \mr{2.5}{$H$} & \mr{2.5}{Unimodal} & \multicolumn{3}{c}{\textbf{Additive}} & \multicolumn{3}{c}{\textbf{Concat}} & \multicolumn{4}{c}{\textbf{Constrained}} \\
\cmidrule(lr){5-7}\cmidrule(lr){8-10}\cmidrule(lr){11-14}
& & & & First & Middle & Last & First & Middle & Last & Orthogonal & FiLM & Gating & \textbf{CFA } \\
\midrule
\mr{40}{\rotatebox{90}{\textbf{FEDformer}}} & \mr{4}{Agriculture} & \hz{6} & 0.060 & \blue{0.063} & \blue{0.064} & \textbf{\red{0.058}} & \blue{0.062} & \textbf{\red{0.058}} & \blue{1.314} & \blue{0.063} & \textbf{\red{0.057}} & \blue{0.061} & \textbf{\red{0.054}} \\
 &  & \hz{8} & 0.072 & \blue{0.075} & \blue{0.074} & \blue{0.076} & \blue{0.079} & \textbf{\red{0.068}} & \blue{1.026} & \blue{0.072} & \blue{0.073} & \textbf{\red{0.071}} & \textbf{\red{0.069}} \\
 &  & \hz{10} & 0.099 & \textbf{\red{0.096}} & \textbf{\red{0.090}} & \textbf{\red{0.091}} & \blue{0.104} & \textbf{\red{0.096}} & \blue{0.278} & \textbf{\red{0.089}} & \textbf{\red{0.089}} & \textbf{\red{0.089}} & \textbf{\red{0.090}} \\
 &  & \hz{12} & 0.117 & \textbf{\red{0.113}} & \blue{0.120} & \blue{0.118} & \blue{0.124} & \blue{0.119} & \blue{0.435} & \blue{0.118} & \blue{0.120} & \blue{0.123} & \blue{0.117} \\
\cmidrule(lr){2-14}
 & \mr{4}{Climate} & \hz{6} & 1.116 & \blue{1.155} & \textbf{\red{1.093}} & \blue{1.141} & \textbf{\red{1.071}} & \textbf{\red{1.085}} & \blue{1.242} & \textbf{\red{1.093}} & \textbf{\red{1.066}} & \textbf{\red{1.055}} & \textbf{\red{1.058}} \\
 &  & \hz{8} & 1.082 & \textbf{\red{1.071}} & \textbf{\red{1.057}} & \textbf{\red{1.078}} & \textbf{\red{1.077}} & \textbf{\red{1.050}} & \textbf{\red{1.052}} & \textbf{\red{1.057}} & \textbf{\red{1.049}} & \textbf{\red{1.076}} & \textbf{\red{1.044}} \\
\bottomrule
\end{tabular}
\end{adjustbox}
\end{table*}

\begin{table*}[!htbp]
\caption{Comparison of multimodal fusion strategies with \textbf{FiLM} (+ \textbf{BERT}).}
\centering
\begin{adjustbox}{width=1.00\linewidth}
\begin{tabular}{c|ll|c|ccc|ccc|cccc}
\toprule
\mr{2.5}{\shortstack{TS \\ Model}} & \mr{2.5}{Dataset} & \mr{2.5}{$H$} & \mr{2.5}{Unimodal} & \multicolumn{3}{c}{\textbf{Additive}} & \multicolumn{3}{c}{\textbf{Concat}} & \multicolumn{4}{c}{\textbf{Constrained}} \\
\cmidrule(lr){5-7}\cmidrule(lr){8-10}\cmidrule(lr){11-14}
& & & & First & Middle & Last & First & Middle & Last & Orthogonal & FiLM & Gating & \textbf{CFA } \\
\midrule
\mr{40}{\rotatebox{90}{\textbf{FiLM}}} & \mr{4}{Agriculture} & \hz{6} & 0.061 & \blue{0.103} & \blue{0.062} & \blue{0.064} & \blue{2.776} & \blue{0.062} & \blue{4.712} & \blue{0.064} & \blue{0.064} & \blue{0.061} & \blue{0.063} \\
 &  & \hz{8} & 0.082 & \blue{0.200} & \textbf{\red{0.081}} & \textbf{\red{0.080}} & \blue{2.550} & \blue{0.087} & \blue{1.179} & \textbf{\red{0.080}} & \textbf{\red{0.080}} & \textbf{\red{0.080}} & \textbf{\red{0.078}} \\
 &  & \hz{10} & 0.105 & \blue{0.182} & \textbf{\red{0.105}} & \textbf{\red{0.100}} & \blue{0.149} & \blue{0.109} & \blue{0.337} & \blue{0.105} & \blue{0.107} & \textbf{\red{0.103}} & \textbf{\red{0.103}} \\
 &  & \hz{12} & 0.143 & \textbf{\red{0.142}} & \textbf{\red{0.133}} & \textbf{\red{0.123}} & \blue{1.022} & \textbf{\red{0.136}} & \blue{0.465} & \textbf{\red{0.133}} & \textbf{\red{0.136}} & \textbf{\red{0.136}} & \textbf{\red{0.132}} \\
\cmidrule(lr){2-14}
 & \mr{4}{Climate} & \hz{6} & 1.305 & \textbf{\red{1.274}} & \textbf{\red{1.282}} & \textbf{\red{1.268}} & \blue{1.427} & \textbf{\red{1.266}} & \blue{1.320} & \textbf{\red{1.270}} & \textbf{\red{1.287}} & \textbf{\red{1.233}} & \textbf{\red{1.254}} \\
 &  & \hz{8} & 1.282 & \textbf{\red{1.273}} & \textbf{\red{1.268}} & \textbf{\red{1.268}} & \textbf{\red{1.069}} & \blue{1.292} & \textbf{\red{1.099}} & \blue{1.313} & \textbf{\red{1.263}} & \textbf{\red{1.255}} & \textbf{\red{1.264}} \\
\bottomrule
\end{tabular}
\end{adjustbox}
\end{table*}

\begin{table*}[!htbp]
\caption{Comparison of multimodal fusion strategies with \textbf{TiDE} (+ \textbf{BERT}).}
\centering
\begin{adjustbox}{width=1.00\linewidth}
\begin{tabular}{c|ll|c|ccc|ccc|cccc}
\toprule
\mr{2.5}{\shortstack{TS \\ Model}} & \mr{2.5}{Dataset} & \mr{2.5}{$H$} & \mr{2.5}{Unimodal} & \multicolumn{3}{c}{\textbf{Additive}} & \multicolumn{3}{c}{\textbf{Concat}} & \multicolumn{4}{c}{\textbf{Constrained}} \\
\cmidrule(lr){5-7}\cmidrule(lr){8-10}\cmidrule(lr){11-14}
& & & & First & Middle & Last & First & Middle & Last & Orthogonal & FiLM & Gating & \textbf{CFA } \\
\midrule
\mr{40}{\rotatebox{90}{\textbf{TiDE}}} & \mr{4}{Agriculture} & \hz{6} & 0.071 & \textbf{\red{0.069}} & \blue{0.082} & \textbf{\red{0.069}} & \blue{1.152} & \textbf{\red{0.065}} & \blue{1.159} & \textbf{\red{0.063}} & \textbf{\red{0.063}} & \textbf{\red{0.062}} & \textbf{\red{0.060}} \\
 &  & \hz{8} & 0.091 & \textbf{\red{0.090}} & \blue{0.092} & \textbf{\red{0.091}} & \blue{1.226} & \blue{0.093} & \blue{1.179} & \textbf{\red{0.082}} & \textbf{\red{0.082}} & \textbf{\red{0.080}} & \textbf{\red{0.082}} \\
 &  & \hz{10} & 0.112 & \blue{7.192} & \textbf{\red{0.110}} & \blue{0.112} & \blue{11.723} & \blue{0.120} & \blue{0.435} & \textbf{\red{0.103}} & \textbf{\red{0.104}} & \textbf{\red{0.104}} & \textbf{\red{0.103}} \\
 &  & \hz{12} & 0.160 & \blue{8.039} & \textbf{\red{0.144}} & \textbf{\red{0.135}} & \blue{14.564} & \textbf{\red{0.150}} & \blue{0.405} & \textbf{\red{0.135}} & \textbf{\red{0.134}} & \textbf{\red{0.134}} & \textbf{\red{0.135}} \\
\cmidrule(lr){2-14}
 & \mr{4}{Climate} & \hz{6} & 1.456 & \blue{1.577} & \blue{1.606} & \blue{1.576} & \blue{1.491} & \textbf{\red{1.418}} & \textbf{\red{1.363}} & \textbf{\red{1.252}} & \textbf{\red{1.244}} & \textbf{\red{1.239}} & \textbf{\red{1.258}} \\
 &  & \hz{8} & 1.469 & \blue{1.555} & \blue{1.481} & \textbf{\red{1.446}} & \textbf{\red{1.048}} & \blue{1.532} & \textbf{\red{1.172}} & \textbf{\red{1.264}} & \textbf{\red{1.274}} & \textbf{\red{1.258}} & \textbf{\red{1.267}} \\
\bottomrule
\end{tabular}
\end{adjustbox}
\end{table*}
\FloatBarrier
\begin{table*}[!htbp]
\caption{Comparison of multimodal fusion strategies with \textbf{Autoformer} (+ \textbf{BERT}).}
\centering
\begin{adjustbox}{width=1.00\linewidth}
\begin{tabular}{c|ll|c|ccc|ccc|cccc}
\toprule
\mr{2.5}{\shortstack{TS \\ Model}} & \mr{2.5}{Dataset} & \mr{2.5}{$H$} & \mr{2.5}{Unimodal} & \multicolumn{3}{c}{\textbf{Additive}} & \multicolumn{3}{c}{\textbf{Concat}} & \multicolumn{4}{c}{\textbf{Constrained}} \\
\cmidrule(lr){5-7}\cmidrule(lr){8-10}\cmidrule(lr){11-14}
& & & & First & Middle & Last & First & Middle & Last & Orthogonal & FiLM & Gating & \textbf{CFA } \\
\midrule
\mr{40}{\rotatebox{90}{\textbf{Autoformer}}} & \mr{4}{Agriculture} & \hz{6} & 0.072 & \textbf{\red{0.072}} & \textbf{\red{0.065}} & \textbf{\red{0.063}} & \textbf{\red{0.071}} & \textbf{\red{0.062}} & \blue{1.291} & \textbf{\red{0.065}} & \textbf{\red{0.068}} & \textbf{\red{0.066}} & \textbf{\red{0.061}} \\
 &  & \hz{8} & 0.083 & \textbf{\red{0.082}} & \blue{0.094} & \textbf{\red{0.080}} & \blue{0.089} & \textbf{\red{0.070}} & \blue{1.244} & \textbf{\red{0.082}} & \blue{0.094} & \textbf{\red{0.079}} & \blue{0.089} \\
 &  & \hz{10} & 0.112 & \textbf{\red{0.106}} & \textbf{\red{0.105}} & \textbf{\red{0.097}} & \textbf{\red{0.109}} & \textbf{\red{0.096}} & \blue{0.393} & \textbf{\red{0.104}} & \textbf{\red{0.110}} & \textbf{\red{0.100}} & \textbf{\red{0.097}} \\
 &  & \hz{12} & 0.136 & \blue{0.140} & \blue{0.139} & \textbf{\red{0.129}} & \blue{0.141} & \textbf{\red{0.134}} & \blue{0.505} & \textbf{\red{0.125}} & \textbf{\red{0.132}} & \textbf{\red{0.128}} & \textbf{\red{0.130}} \\
\cmidrule(lr){2-14}
 & \mr{4}{Climate} & \hz{6} & 1.240 & \blue{1.278} & \blue{1.248} & \textbf{\red{1.222}} & \textbf{\red{1.168}} & \textbf{\red{1.192}} & \blue{1.266} & \blue{1.244} & \blue{1.248} & \textbf{\red{1.206}} & \textbf{\red{1.225}} \\
 &  & \hz{8} & 1.200 & \blue{1.234} & \blue{1.265} & \textbf{\red{1.199}} & \textbf{\red{1.180}} & \textbf{\red{1.168}} & \textbf{\red{1.122}} & \blue{1.241} & \textbf{\red{1.162}} & \textbf{\red{1.132}} & \textbf{\red{1.179}} \\
\bottomrule
\end{tabular}
\end{adjustbox}
\end{table*}

\begin{table*}[!htbp]
\caption{Comparison of multimodal fusion strategies with \textbf{Crossformer} (+ \textbf{BERT}).}
\centering
\begin{adjustbox}{width=1.00\linewidth}
\begin{tabular}{c|ll|c|ccc|ccc|cccc}
\toprule
\mr{2.5}{\shortstack{TS \\ Model}} & \mr{2.5}{Dataset} & \mr{2.5}{$H$} & \mr{2.5}{Unimodal} & \multicolumn{3}{c}{\textbf{Additive}} & \multicolumn{3}{c}{\textbf{Concat}} & \multicolumn{4}{c}{\textbf{Constrained}} \\
\cmidrule(lr){5-7}\cmidrule(lr){8-10}\cmidrule(lr){11-14}
& & & & First & Middle & Last & First & Middle & Last & Orthogonal & FiLM & Gating & \textbf{CFA } \\
\midrule
\mr{40}{\rotatebox{90}{\textbf{Crossformer}}} & \mr{4}{Agriculture} & \hz{6} & 0.088 & \blue{0.132} & \blue{0.146} & \blue{0.115} & \textbf{\red{0.088}} & \blue{0.152} & \blue{0.210} & \blue{0.159} & \blue{0.184} & \blue{0.213} & \blue{0.166} \\
 &  & \hz{8} & 0.207 & \textbf{\red{0.194}} & \blue{0.227} & \blue{0.235} & \blue{1.212} & \blue{0.265} & \blue{0.318} & \textbf{\red{0.171}} & \blue{0.238} & \textbf{\red{0.167}} & \blue{0.247} \\
 &  & \hz{10} & 0.252 & \blue{0.301} & \textbf{\red{0.215}} & \blue{0.376} & \textbf{\red{0.186}} & \blue{0.300} & \blue{0.470} & \textbf{\red{0.229}} & \textbf{\red{0.186}} & \blue{0.255} & \blue{0.260} \\
 &  & \hz{12} & 0.311 & \blue{0.325} & \textbf{\red{0.225}} & \blue{0.366} & \blue{0.404} & \blue{0.411} & \blue{0.365} & \blue{0.387} & \textbf{\red{0.301}} & \blue{0.318} & \blue{0.393} \\
\cmidrule(lr){2-14}
 & \mr{4}{Climate} & \hz{6} & 1.039 & \blue{1.054} & \blue{1.083} & \blue{1.065} & \blue{1.057} & \blue{1.042} & \blue{1.051} & \blue{1.062} & \blue{1.063} & \textbf{\red{1.012}} & \blue{1.119} \\
 &  & \hz{8} & 1.080 & \textbf{\red{1.070}} & \textbf{\red{1.047}} & \textbf{\red{1.066}} & \blue{1.112} & \textbf{\red{1.072}} & \blue{1.102} & \blue{1.095} & \textbf{\red{1.074}} & \blue{1.080} & \blue{1.090} \\
\bottomrule
\end{tabular}
\end{adjustbox}
\end{table*}

\begin{table*}[!htbp]
\caption{Comparison of multimodal fusion strategies with \textbf{Reformer} (+ \textbf{BERT}).}
\centering
\begin{adjustbox}{width=1.00\linewidth}
\begin{tabular}{c|ll|c|ccc|ccc|cccc}
\toprule
\mr{2.5}{\shortstack{TS \\ Model}} & \mr{2.5}{Dataset} & \mr{2.5}{$H$} & \mr{2.5}{Unimodal} & \multicolumn{3}{c}{\textbf{Additive}} & \multicolumn{3}{c}{\textbf{Concat}} & \multicolumn{4}{c}{\textbf{Constrained}} \\
\cmidrule(lr){5-7}\cmidrule(lr){8-10}\cmidrule(lr){11-14}
& & & & First & Middle & Last & First & Middle & Last & Orthogonal & FiLM & Gating & \textbf{CFA } \\
\midrule
\mr{40}{\rotatebox{90}{\textbf{Reformer}}} & \mr{4}{Agriculture} & \hz{6} & 0.194 & \blue{0.222} & \blue{0.319} & \blue{0.203} & \blue{0.231} & \textbf{\red{0.182}} & \blue{0.308} & \textbf{\red{0.132}} & \textbf{\red{0.142}} & \blue{0.198} & \textbf{\red{0.183}} \\
 &  & \hz{8} & 0.254 & \blue{0.383} & \blue{0.323} & \textbf{\red{0.244}} & \textbf{\red{0.234}} & \blue{0.375} & \blue{0.406} & \blue{0.325} & \textbf{\red{0.218}} & \blue{0.391} & \textbf{\red{0.231}} \\
 &  & \hz{10} & 0.215 & \blue{0.519} & \blue{0.323} & \blue{0.339} & \blue{0.529} & \blue{0.299} & \blue{0.701} & \blue{0.322} & \blue{0.363} & \blue{0.379} & \blue{0.295} \\
 &  & \hz{12} & 0.467 & \blue{0.490} & \blue{0.518} & \textbf{\red{0.435}} & \blue{0.644} & \blue{0.588} & \textbf{\red{0.384}} & \blue{0.551} & \textbf{\red{0.437}} & \blue{0.479} & \textbf{\red{0.242}} \\
\cmidrule(lr){2-14}
 & \mr{4}{Climate} & \hz{6} & 1.025 & \blue{1.036} & \textbf{\red{1.012}} & \blue{1.092} & \blue{1.056} & \blue{1.060} & \textbf{\red{0.967}} & \textbf{\red{1.010}} & \blue{1.047} & \blue{1.028} & \blue{1.074} \\
 &  & \hz{8} & 1.066 & \blue{1.070} & \textbf{\red{1.042}} & \textbf{\red{1.051}} & \textbf{\red{1.037}} & \blue{1.101} & \textbf{\red{1.053}} & \textbf{\red{1.041}} & \blue{1.070} & \blue{1.088} & \blue{1.079} \\
\bottomrule
\end{tabular}
\end{adjustbox}
\end{table*}

\begin{table*}[!htbp]
\caption{Comparison of multimodal fusion strategies with \textbf{Transformer} (+ \textbf{BERT}).}
\centering
\begin{adjustbox}{width=1.00\linewidth}
\begin{tabular}{c|ll|c|ccc|ccc|cccc}
\toprule
\mr{2.5}{\shortstack{TS \\ Model}} & \mr{2.5}{Dataset} & \mr{2.5}{$H$} & \mr{2.5}{Unimodal} & \multicolumn{3}{c}{\textbf{Additive}} & \multicolumn{3}{c}{\textbf{Concat}} & \multicolumn{4}{c}{\textbf{Constrained}} \\
\cmidrule(lr){5-7}\cmidrule(lr){8-10}\cmidrule(lr){11-14}
& & & & First & Middle & Last & First & Middle & Last & Orthogonal & FiLM & Gating & \textbf{CFA } \\
\midrule
\mr{40}{\rotatebox{90}{\textbf{Transformer}}} & \mr{4}{Agriculture} & \hz{6} & 0.133 & \textbf{\red{0.119}} & \blue{0.177} & \blue{0.137} & \blue{0.159} & \textbf{\red{0.119}} & \blue{0.217} & \blue{0.188} & \blue{0.165} & \blue{0.140} & \blue{0.153} \\
 &  & \hz{8} & 0.147 & \blue{0.334} & \textbf{\red{0.143}} & \blue{0.243} & \blue{0.210} & \blue{0.174} & \blue{0.433} & \blue{0.163} & \blue{0.161} & \blue{0.211} & \blue{0.182} \\
 &  & \hz{10} & 0.177 & \blue{0.524} & \blue{0.282} & \blue{0.273} & \textbf{\red{0.133}} & \blue{0.210} & \blue{0.666} & \blue{0.203} & \textbf{\red{0.172}} & \blue{0.178} & \textbf{\red{0.158}} \\
 &  & \hz{12} & 0.247 & \blue{0.312} & \blue{0.254} & \blue{0.325} & \textbf{\red{0.174}} & \blue{0.286} & \blue{0.485} & \blue{0.264} & \textbf{\red{0.240}} & \blue{0.254} & \textbf{\red{0.244}} \\
\cmidrule(lr){2-14}
 & \mr{4}{Climate} & \hz{6} & 0.985 & \blue{1.032} & \blue{0.989} & \blue{1.022} & \textbf{\red{0.875}} & \blue{0.990} & \textbf{\red{0.983}} & \blue{0.989} & \blue{1.024} & \blue{1.055} & \textbf{\red{0.984}} \\
 &  & \hz{8} & 1.058 & \textbf{\red{1.055}} & \textbf{\red{0.989}} & \textbf{\red{1.055}} & \blue{1.086} & \blue{1.091} & \textbf{\red{0.990}} & \textbf{\red{0.989}} & \textbf{\red{1.032}} & \textbf{\red{1.054}} & \textbf{\red{1.053}} \\
\bottomrule
\end{tabular}
\end{adjustbox}
\end{table*}
\FloatBarrier
\begin{table*}[!htbp]
\caption{Comparison of multimodal fusion strategies with \textbf{TSMixer} (+ \textbf{BERT}).}
\centering
\begin{adjustbox}{width=1.00\linewidth}
\begin{tabular}{c|ll|c|ccc|ccc|cccc}
\toprule
\mr{2.5}{\shortstack{TS \\ Model}} & \mr{2.5}{Dataset} & \mr{2.5}{$H$} & \mr{2.5}{Unimodal} & \multicolumn{3}{c}{\textbf{Additive}} & \multicolumn{3}{c}{\textbf{Concat}} & \multicolumn{4}{c}{\textbf{Constrained}} \\
\cmidrule(lr){5-7}\cmidrule(lr){8-10}\cmidrule(lr){11-14}
& & & & First & Middle & Last & First & Middle & Last & Orthogonal & FiLM & Gating & \textbf{CFA } \\
\midrule
\mr{40}{\rotatebox{90}{\textbf{TSMixer}}} & \mr{4}{Agriculture} & \hz{6} & 0.083 & \blue{0.105} & \blue{0.142} & \blue{0.117} & \blue{0.154} & \blue{0.142} & \blue{0.111} & \blue{0.103} & \blue{0.158} & \blue{0.113} & \blue{0.092} \\
 &  & \hz{8} & 0.125 & \blue{0.220} & \blue{0.217} & \blue{0.180} & \blue{0.278} & \blue{0.182} & \blue{0.165} & \blue{0.185} & \blue{0.217} & \blue{0.174} & \blue{0.136} \\
 &  & \hz{10} & 0.428 & \blue{0.515} & \blue{1.135} & \textbf{\red{0.390}} & \textbf{\red{0.380}} & \blue{1.077} & \textbf{\red{0.240}} & \textbf{\red{0.323}} & \blue{1.309} & \blue{0.619} & \textbf{\red{0.204}} \\
 &  & \hz{12} & 0.834 & \blue{1.352} & \blue{1.080} & \blue{1.080} & \blue{1.496} & \textbf{\red{0.458}} & \textbf{\red{0.386}} & \textbf{\red{0.397}} & \blue{1.398} & \blue{1.012} & \textbf{\red{0.281}} \\
\cmidrule(lr){2-14}
 & \mr{4}{Climate} & \hz{6} & 1.054 & \blue{1.061} & \blue{1.061} & \blue{1.080} & \blue{1.085} & \blue{1.083} & \blue{1.101} & \blue{1.065} & \textbf{\red{1.054}} & \blue{1.069} & \blue{1.070} \\
 &  & \hz{8} & 1.088 & \textbf{\red{1.073}} & \blue{1.097} & \textbf{\red{1.084}} & \blue{1.100} & \textbf{\red{1.085}} & \blue{1.096} & \blue{1.111} & \blue{1.108} & \blue{1.091} & \blue{1.095} \\
\bottomrule
\end{tabular}
\end{adjustbox}
\end{table*}

\begin{table*}[!htbp]
\caption{Comparison of multimodal fusion strategies with \textbf{Informer} (+ \textbf{BERT}).}
\centering
\begin{adjustbox}{width=1.00\linewidth}
\begin{tabular}{c|ll|c|ccc|ccc|cccc}
\toprule
\mr{2.5}{\shortstack{TS \\ Model}} & \mr{2.5}{Dataset} & \mr{2.5}{$H$} & \mr{2.5}{Unimodal} & \multicolumn{3}{c}{\textbf{Additive}} & \multicolumn{3}{c}{\textbf{Concat}} & \multicolumn{4}{c}{\textbf{Constrained}} \\
\cmidrule(lr){5-7}\cmidrule(lr){8-10}\cmidrule(lr){11-14}
& & & & First & Middle & Last & First & Middle & Last & Orthogonal & FiLM & Gating & \textbf{CFA } \\
\midrule
\mr{40}{\rotatebox{90}{\textbf{Informer}}} & \mr{4}{Agriculture} & \hz{6} & 0.212 & \blue{0.373} & \blue{0.283} & \blue{0.319} & \blue{0.308} & \textbf{\red{0.194}} & \blue{0.228} & \textbf{\red{0.189}} & \blue{0.395} & \blue{0.293} & \textbf{\red{0.198}} \\
 &  & \hz{8} & 0.315 & \blue{0.415} & \textbf{\red{0.251}} & \blue{0.346} & \textbf{\red{0.204}} & \textbf{\red{0.280}} & \blue{0.322} & \textbf{\red{0.193}} & \blue{0.392} & \blue{0.352} & \blue{0.348} \\
 &  & \hz{10} & 0.417 & \textbf{\red{0.236}} & \textbf{\red{0.386}} & \blue{0.440} & \blue{0.578} & \textbf{\red{0.353}} & \textbf{\red{0.381}} & \textbf{\red{0.321}} & \textbf{\red{0.352}} & \textbf{\red{0.311}} & \textbf{\red{0.403}} \\
 &  & \hz{12} & 0.565 & \textbf{\red{0.405}} & \blue{0.621} & \blue{0.593} & \textbf{\red{0.320}} & \blue{0.648} & \blue{0.597} & \blue{0.581} & \textbf{\red{0.436}} & \textbf{\red{0.365}} & \textbf{\red{0.453}} \\
\cmidrule(lr){2-14}
 & \mr{4}{Climate} & \hz{6} & 1.070 & \textbf{\red{1.066}} & \blue{1.111} & \textbf{\red{1.058}} & \textbf{\red{1.019}} & \textbf{\red{1.034}} & \textbf{\red{1.022}} & \textbf{\red{1.046}} & \textbf{\red{1.031}} & \textbf{\red{1.061}} & \textbf{\red{1.040}} \\
 &  & \hz{8} & 1.081 & \textbf{\red{1.065}} & \blue{1.083} & \textbf{\red{1.073}} & \textbf{\red{0.916}} & \blue{1.087} & \textbf{\red{1.073}} & \textbf{\red{1.047}} & \textbf{\red{1.073}} & \textbf{\red{1.049}} & \blue{1.108} \\
\bottomrule
\end{tabular}
\end{adjustbox}
\end{table*}

\begin{table*}[!htbp]
\caption{Comparison of multimodal fusion strategies with \textbf{Nonstationary Transformer} (+ \textbf{GPT}).}
\centering
\begin{adjustbox}{width=1.00\linewidth}
\begin{tabular}{c|ll|c|ccc|ccc|cccc}
\toprule
\mr{2.5}{\shortstack{TS \\ Model}} & \mr{2.5}{Dataset} & \mr{2.5}{$H$} & \mr{2.5}{Unimodal} & \multicolumn{3}{c}{\textbf{Additive}} & \multicolumn{3}{c}{\textbf{Concat}} & \multicolumn{4}{c}{\textbf{Constrained}} \\
\cmidrule(lr){5-7}\cmidrule(lr){8-10}\cmidrule(lr){11-14}
& & & & First & Middle & Last & First & Middle & Last & Orthogonal & FiLM & Gating & \textbf{CFA } \\
\midrule
\mr{40}{\rotatebox{90}{\textbf{Nonstationary Transformer}}} & \mr{4}{Agriculture} & \hz{6} & 0.055 & \blue{0.061} & \textbf{\red{0.050}} & \textbf{\red{0.052}} & \blue{0.062} & \textbf{\red{0.054}} & \blue{1.746} & \textbf{\red{0.051}} & \textbf{\red{0.053}} & \blue{0.058} & \textbf{\red{0.053}} \\
 &  & \hz{8} & 0.071 & \blue{0.073} & \textbf{\red{0.070}} & \textbf{\red{0.068}} & \blue{0.073} & \textbf{\red{0.070}} & \blue{1.026} & \textbf{\red{0.070}} & \textbf{\red{0.070}} & \blue{0.074} & \textbf{\red{0.067}} \\
 &  & \hz{10} & 0.092 & \blue{0.102} & \textbf{\red{0.090}} & \textbf{\red{0.088}} & \blue{0.100} & \textbf{\red{0.091}} & \blue{0.367} & \textbf{\red{0.092}} & \textbf{\red{0.092}} & \textbf{\red{0.090}} & \textbf{\red{0.092}} \\
 &  & \hz{12} & 0.116 & \blue{0.121} & \blue{0.119} & \textbf{\red{0.110}} & \blue{0.127} & \blue{0.120} & \blue{0.428} & \blue{0.121} & \blue{0.126} & \blue{0.118} & \blue{0.116} \\
\cmidrule(lr){2-14}
 & \mr{4}{Climate} & \hz{6} & 1.206 & \blue{1.316} & \blue{1.302} & \textbf{\red{1.192}} & \textbf{\red{1.106}} & \blue{1.237} & \blue{1.296} & \blue{1.304} & \textbf{\red{1.179}} & \blue{1.226} & \textbf{\red{1.084}} \\
 &  & \hz{8} & 1.240 & \blue{1.301} & \textbf{\red{1.128}} & \textbf{\red{1.232}} & \blue{1.248} & \blue{1.302} & \textbf{\red{1.154}} & \blue{1.286} & \textbf{\red{1.238}} & \textbf{\red{1.203}} & \blue{1.246} \\
\bottomrule
\end{tabular}
\end{adjustbox}
\end{table*}

\begin{table*}[!htbp]
\caption{Comparison of multimodal fusion strategies with \textbf{Koopa} (+ \textbf{GPT}).}
\centering
\begin{adjustbox}{width=1.00\linewidth}
\begin{tabular}{c|ll|c|ccc|ccc|cccc}
\toprule
\mr{2.5}{\shortstack{TS \\ Model}} & \mr{2.5}{Dataset} & \mr{2.5}{$H$} & \mr{2.5}{Unimodal} & \multicolumn{3}{c}{\textbf{Additive}} & \multicolumn{3}{c}{\textbf{Concat}} & \multicolumn{4}{c}{\textbf{Constrained}} \\
\cmidrule(lr){5-7}\cmidrule(lr){8-10}\cmidrule(lr){11-14}
& & & & First & Middle & Last & First & Middle & Last & Orthogonal & FiLM & Gating & \textbf{CFA } \\
\midrule
\mr{40}{\rotatebox{90}{\textbf{Koopa}}} & \mr{4}{Agriculture} & \hz{6} & 0.059 & \textbf{\red{0.057}} & \textbf{\red{0.058}} & \textbf{\red{0.058}} & \blue{3.581} & \blue{1.539} & \blue{1.539} & \blue{0.061} & \blue{0.072} & \textbf{\red{0.057}} & \textbf{\red{0.056}} \\
 &  & \hz{8} & 0.077 & \textbf{\red{0.070}} & \textbf{\red{0.071}} & \textbf{\red{0.071}} & \blue{0.188} & \blue{0.187} & \blue{0.187} & \blue{0.077} & \blue{0.129} & \textbf{\red{0.074}} & \textbf{\red{0.076}} \\
 &  & \hz{10} & 0.099 & \textbf{\red{0.092}} & \textbf{\red{0.090}} & \textbf{\red{0.090}} & \blue{1.280} & \blue{0.406} & \blue{0.406} & \blue{0.100} & \blue{0.118} & \textbf{\red{0.091}} & \blue{0.100} \\
 &  & \hz{12} & 0.129 & \textbf{\red{0.118}} & \textbf{\red{0.118}} & \textbf{\red{0.118}} & \blue{0.799} & \blue{0.662} & \blue{0.662} & \blue{0.130} & \blue{0.174} & \textbf{\red{0.117}} & \textbf{\red{0.129}} \\
\cmidrule(lr){2-14}
 & \mr{4}{Climate} & \hz{6} & 1.251 & \blue{14.469} & \blue{1.255} & \blue{1.255} & \blue{6.592} & \textbf{\red{1.219}} & \textbf{\red{1.219}} & \blue{1.268} & \textbf{\red{1.079}} & \textbf{\red{1.246}} & \blue{1.253} \\
 &  & \hz{8} & 1.264 & \blue{1.541} & \blue{1.267} & \blue{1.267} & \textbf{\red{1.195}} & \textbf{\red{1.114}} & \textbf{\red{1.114}} & \blue{1.320} & \textbf{\red{1.072}} & \blue{1.268} & \textbf{\red{1.258}} \\
\bottomrule
\end{tabular}
\end{adjustbox}
\end{table*}

\begin{table*}[!htbp]
\caption{Comparison of multimodal fusion strategies with \textbf{iTransformer} (+ \textbf{GPT}).}
\centering
\begin{adjustbox}{width=1.00\linewidth}
\begin{tabular}{c|ll|c|ccc|ccc|cccc}
\toprule
\mr{2.5}{\shortstack{TS \\ Model}} & \mr{2.5}{Dataset} & \mr{2.5}{$H$} & \mr{2.5}{Unimodal} & \multicolumn{3}{c}{\textbf{Additive}} & \multicolumn{3}{c}{\textbf{Concat}} & \multicolumn{4}{c}{\textbf{Constrained}} \\
\cmidrule(lr){5-7}\cmidrule(lr){8-10}\cmidrule(lr){11-14}
& & & & First & Middle & Last & First & Middle & Last & Orthogonal & FiLM & Gating & \textbf{CFA } \\
\midrule
\mr{40}{\rotatebox{90}{\textbf{iTransformer}}} & \mr{4}{Agriculture} & \hz{6} & 0.059 & \blue{0.060} & \blue{0.062} & \textbf{\red{0.057}} & \blue{0.060} & \textbf{\red{0.058}} & \blue{1.036} & \blue{0.060} & \blue{0.060} & \blue{0.060} & \textbf{\red{0.059}} \\
 &  & \hz{8} & 0.077 & \blue{0.079} & \blue{0.082} & \textbf{\red{0.072}} & \textbf{\red{0.077}} & \blue{0.080} & \blue{1.057} & \blue{0.081} & \blue{0.079} & \blue{0.081} & \textbf{\red{0.076}} \\
 &  & \hz{10} & 0.096 & \blue{0.104} & \blue{0.103} & \textbf{\red{0.090}} & \blue{0.101} & \blue{0.102} & \blue{0.301} & \blue{0.103} & \blue{0.101} & \blue{0.103} & \blue{0.099} \\
 &  & \hz{12} & 0.128 & \blue{0.131} & \blue{0.134} & \textbf{\red{0.117}} & \blue{0.131} & \blue{0.133} & \blue{0.303} & \textbf{\red{0.127}} & \blue{0.132} & \blue{0.129} & \textbf{\red{0.125}} \\
\cmidrule(lr){2-14}
 & \mr{4}{Climate} & \hz{6} & 1.096 & \blue{1.185} & \blue{1.174} & \blue{1.131} & \blue{1.122} & \blue{1.212} & \blue{1.326} & \blue{1.123} & \blue{1.168} & \blue{1.151} & \blue{1.121} \\
 &  & \hz{8} & 1.178 & \blue{1.207} & \blue{1.232} & \blue{1.202} & \textbf{\red{1.169}} & \blue{1.215} & \blue{1.187} & \blue{1.231} & \blue{1.258} & \blue{1.203} & \blue{1.236} \\
\bottomrule
\end{tabular}
\end{adjustbox}
\end{table*}

\begin{table*}[!htbp]
\caption{Comparison of multimodal fusion strategies with \textbf{DLinear} (+ \textbf{GPT}).}
\centering
\begin{adjustbox}{width=1.00\linewidth}
\begin{tabular}{c|ll|c|ccc|ccc|cccc}
\toprule
\mr{2.5}{\shortstack{TS \\ Model}} & \mr{2.5}{Dataset} & \mr{2.5}{$H$} & \mr{2.5}{Unimodal} & \multicolumn{3}{c}{\textbf{Additive}} & \multicolumn{3}{c}{\textbf{Concat}} & \multicolumn{4}{c}{\textbf{Constrained}} \\
\cmidrule(lr){5-7}\cmidrule(lr){8-10}\cmidrule(lr){11-14}
& & & & First & Middle & Last & First & Middle & Last & Orthogonal & FiLM & Gating & \textbf{CFA } \\
\midrule
\mr{40}{\rotatebox{90}{\textbf{DLinear}}} & \mr{4}{Agriculture} & \hz{6} & 0.114 & \blue{0.120} & \blue{0.132} & \blue{0.132} & \blue{0.455} & \blue{0.802} & \blue{0.392} & \textbf{\red{0.112}} & \blue{0.235} & \blue{0.133} & \textbf{\red{0.085}} \\
 &  & \hz{8} & 0.324 & \textbf{\red{0.173}} & \textbf{\red{0.175}} & \textbf{\red{0.183}} & \blue{1.168} & \blue{1.372} & \blue{1.182} & \textbf{\red{0.206}} & \textbf{\red{0.232}} & \textbf{\red{0.221}} & \textbf{\red{0.099}} \\
 &  & \hz{10} & 0.163 & \blue{0.198} & \blue{0.203} & \blue{0.209} & \blue{0.532} & \blue{0.445} & \blue{0.436} & \blue{0.197} & \blue{0.221} & \blue{0.216} & \textbf{\red{0.127}} \\
 &  & \hz{12} & 0.217 & \blue{0.241} & \blue{0.243} & \blue{0.242} & \blue{0.446} & \blue{0.364} & \blue{0.426} & \textbf{\red{0.162}} & \blue{0.348} & \textbf{\red{0.180}} & \textbf{\red{0.183}} \\
\cmidrule(lr){2-14}
 & \mr{4}{Climate} & \hz{6} & 1.198 & \blue{1.292} & \blue{1.255} & \blue{1.226} & \blue{1.312} & \blue{1.241} & \blue{1.226} & \blue{1.312} & \textbf{\red{1.080}} & \blue{1.285} & \textbf{\red{1.090}} \\
 &  & \hz{8} & 1.364 & \textbf{\red{1.329}} & \textbf{\red{1.288}} & \textbf{\red{1.264}} & \textbf{\red{1.240}} & \textbf{\red{1.172}} & \textbf{\red{1.160}} & \textbf{\red{1.318}} & \textbf{\red{1.084}} & \blue{1.368} & \textbf{\red{1.090}} \\
\bottomrule
\end{tabular}
\end{adjustbox}
\end{table*}
\FloatBarrier
\begin{table*}[!htbp]
\caption{Comparison of multimodal fusion strategies with \textbf{PatchTST} (+ \textbf{GPT}).}
\centering
\begin{adjustbox}{width=1.00\linewidth}
\begin{tabular}{c|ll|c|ccc|ccc|cccc}
\toprule
\mr{2.5}{\shortstack{TS \\ Model}} & \mr{2.5}{Dataset} & \mr{2.5}{$H$} & \mr{2.5}{Unimodal} & \multicolumn{3}{c}{\textbf{Additive}} & \multicolumn{3}{c}{\textbf{Concat}} & \multicolumn{4}{c}{\textbf{Constrained}} \\
\cmidrule(lr){5-7}\cmidrule(lr){8-10}\cmidrule(lr){11-14}
& & & & First & Middle & Last & First & Middle & Last & Orthogonal & FiLM & Gating & \textbf{CFA } \\
\midrule
\mr{40}{\rotatebox{90}{\textbf{PatchTST}}} & \mr{4}{Agriculture} & \hz{6} & 0.059 & \blue{0.060} & \blue{0.062} & \textbf{\red{0.057}} & \blue{0.060} & \blue{0.061} & \blue{0.639} & \blue{0.062} & \textbf{\red{0.059}} & \blue{0.060} & \textbf{\red{0.059}} \\
 &  & \hz{8} & 0.078 & \blue{0.079} & \blue{0.081} & \textbf{\red{0.076}} & \blue{0.079} & \blue{0.080} & \blue{0.814} & \blue{0.081} & \blue{0.079} & \blue{0.079} & \textbf{\red{0.077}} \\
 &  & \hz{10} & 0.100 & \textbf{\red{0.099}} & \blue{0.101} & \textbf{\red{0.097}} & \textbf{\red{0.099}} & \blue{0.102} & \blue{0.342} & \blue{0.101} & \blue{0.102} & \blue{0.101} & \blue{0.100} \\
 &  & \hz{12} & 0.129 & \textbf{\red{0.129}} & \textbf{\red{0.129}} & \textbf{\red{0.120}} & \textbf{\red{0.129}} & \blue{0.129} & \blue{0.394} & \textbf{\red{0.128}} & \textbf{\red{0.129}} & \blue{0.130} & \textbf{\red{0.129}} \\
\cmidrule(lr){2-14}
 & \mr{4}{Climate} & \hz{6} & 1.257 & \textbf{\red{1.234}} & \textbf{\red{1.240}} & \textbf{\red{1.237}} & \textbf{\red{1.234}} & \blue{1.259} & \blue{1.352} & \textbf{\red{1.240}} & \blue{1.263} & \textbf{\red{1.224}} & \textbf{\red{1.225}} \\
 &  & \hz{8} & 1.265 & \textbf{\red{1.255}} & \textbf{\red{1.226}} & \textbf{\red{1.263}} & \textbf{\red{1.255}} & \textbf{\red{1.250}} & \textbf{\red{1.135}} & \textbf{\red{1.226}} & \textbf{\red{1.257}} & \textbf{\red{1.230}} & \textbf{\red{1.253}} \\
\bottomrule
\end{tabular}
\end{adjustbox}
\end{table*}

\begin{table*}[!htbp]
\caption{Comparison of multimodal fusion strategies with \textbf{FiLM} (+ \textbf{GPT}).}
\centering
\begin{adjustbox}{width=1.00\linewidth}
\begin{tabular}{c|ll|c|ccc|ccc|cccc}
\toprule
\mr{2.5}{\shortstack{TS \\ Model}} & \mr{2.5}{Dataset} & \mr{2.5}{$H$} & \mr{2.5}{Unimodal} & \multicolumn{3}{c}{\textbf{Additive}} & \multicolumn{3}{c}{\textbf{Concat}} & \multicolumn{4}{c}{\textbf{Constrained}} \\
\cmidrule(lr){5-7}\cmidrule(lr){8-10}\cmidrule(lr){11-14}
& & & & First & Middle & Last & First & Middle & Last & Orthogonal & FiLM & Gating & \textbf{CFA } \\
\midrule
\mr{40}{\rotatebox{90}{\textbf{FiLM}}} & \mr{4}{Agriculture} & \hz{6} & 0.061 & \blue{0.083} & \blue{0.063} & \blue{0.066} & \blue{2.706} & \blue{0.065} & \blue{1.955} & \blue{0.063} & \blue{0.064} & \blue{0.061} & \blue{0.063} \\
 &  & \hz{8} & 0.082 & \blue{0.087} & \textbf{\red{0.081}} & \textbf{\red{0.082}} & \blue{1.423} & \blue{0.088} & \blue{0.816} & \textbf{\red{0.080}} & \blue{0.083} & \textbf{\red{0.081}} & \textbf{\red{0.078}} \\
 &  & \hz{10} & 0.105 & \blue{21.529} & \textbf{\red{0.104}} & \textbf{\red{0.099}} & \blue{0.149} & \blue{0.106} & \blue{0.362} & \blue{0.107} & \blue{0.106} & \textbf{\red{0.102}} & \textbf{\red{0.103}} \\
 &  & \hz{12} & 0.143 & \textbf{\red{0.139}} & \textbf{\red{0.133}} & \textbf{\red{0.122}} & \blue{0.973} & \textbf{\red{0.136}} & \blue{0.449} & \textbf{\red{0.133}} & \textbf{\red{0.137}} & \textbf{\red{0.137}} & \textbf{\red{0.132}} \\
\cmidrule(lr){2-14}
 & \mr{4}{Climate} & \hz{6} & 1.305 & \textbf{\red{1.263}} & \textbf{\red{1.280}} & \textbf{\red{1.260}} & \blue{1.559} & \textbf{\red{1.264}} & \textbf{\red{1.299}} & \textbf{\red{1.219}} & \textbf{\red{1.293}} & \textbf{\red{1.237}} & \textbf{\red{1.254}} \\
 &  & \hz{8} & 1.282 & \textbf{\red{1.273}} & \textbf{\red{1.277}} & \textbf{\red{1.269}} & \textbf{\red{1.034}} & \blue{1.317} & \textbf{\red{1.155}} & \blue{1.310} & \textbf{\red{1.269}} & \textbf{\red{1.270}} & \textbf{\red{1.264}} \\
\bottomrule
\end{tabular}
\end{adjustbox}
\end{table*}

\begin{table*}[!htbp]
\caption{Comparison of multimodal fusion strategies with \textbf{TiDE} (+ \textbf{GPT}).}
\centering
\begin{adjustbox}{width=1.00\linewidth}
\begin{tabular}{c|ll|c|ccc|ccc|cccc}
\toprule
\mr{2.5}{\shortstack{TS \\ Model}} & \mr{2.5}{Dataset} & \mr{2.5}{$H$} & \mr{2.5}{Unimodal} & \multicolumn{3}{c}{\textbf{Additive}} & \multicolumn{3}{c}{\textbf{Concat}} & \multicolumn{4}{c}{\textbf{Constrained}} \\
\cmidrule(lr){5-7}\cmidrule(lr){8-10}\cmidrule(lr){11-14}
& & & & First & Middle & Last & First & Middle & Last & Orthogonal & FiLM & Gating & \textbf{CFA } \\
\midrule
\mr{40}{\rotatebox{90}{\textbf{TiDE}}} & \mr{4}{Agriculture} & \hz{6} & 0.073 & \textbf{\red{0.068}} & \blue{0.082} & \textbf{\red{0.068}} & \blue{1.720} & \textbf{\red{0.065}} & \blue{1.699} & \textbf{\red{0.063}} & \textbf{\red{0.063}} & \textbf{\red{0.062}} & \textbf{\red{0.060}} \\
 &  & \hz{8} & 0.088 & \blue{0.091} & \blue{0.092} & \blue{0.092} & \blue{1.005} & \blue{0.093} & \blue{0.826} & \textbf{\red{0.082}} & \textbf{\red{0.082}} & \textbf{\red{0.080}} & \textbf{\red{0.082}} \\
 &  & \hz{10} & 0.112 & \blue{17.163} & \textbf{\red{0.110}} & \blue{0.112} & \blue{19.212} & \blue{0.120} & \blue{0.419} & \textbf{\red{0.103}} & \textbf{\red{0.104}} & \textbf{\red{0.104}} & \textbf{\red{0.103}} \\
 &  & \hz{12} & 0.160 & \blue{20.607} & \textbf{\red{0.144}} & \textbf{\red{0.124}} & \blue{29.733} & \textbf{\red{0.150}} & \blue{0.353} & \textbf{\red{0.135}} & \textbf{\red{0.134}} & \textbf{\red{0.134}} & \textbf{\red{0.135}} \\
\cmidrule(lr){2-14}
 & \mr{4}{Climate} & \hz{6} & 1.515 & \blue{1.601} & \blue{1.606} & \blue{1.600} & \textbf{\red{1.496}} & \textbf{\red{1.418}} & \textbf{\red{1.379}} & \textbf{\red{1.252}} & \textbf{\red{1.244}} & \textbf{\red{1.239}} & \textbf{\red{1.258}} \\
 &  & \hz{8} & 1.486 & \blue{1.567} & \textbf{\red{1.481}} & \textbf{\red{1.458}} & \textbf{\red{1.081}} & \blue{1.532} & \textbf{\red{1.184}} & \textbf{\red{1.264}} & \textbf{\red{1.274}} & \textbf{\red{1.258}} & \textbf{\red{1.267}} \\
\bottomrule
\end{tabular}
\end{adjustbox}
\end{table*}

\begin{table*}[!htbp]
\caption{Comparison of multimodal fusion strategies with \textbf{FEDformer} (+ \textbf{GPT}).}
\centering
\begin{adjustbox}{width=1.00\linewidth}
\begin{tabular}{c|ll|c|ccc|ccc|cccc}
\toprule
\mr{2.5}{\shortstack{TS \\ Model}} & \mr{2.5}{Dataset} & \mr{2.5}{$H$} & \mr{2.5}{Unimodal} & \multicolumn{3}{c}{\textbf{Additive}} & \multicolumn{3}{c}{\textbf{Concat}} & \multicolumn{4}{c}{\textbf{Constrained}} \\
\cmidrule(lr){5-7}\cmidrule(lr){8-10}\cmidrule(lr){11-14}
& & & & First & Middle & Last & First & Middle & Last & Orthogonal & FiLM & Gating & \textbf{CFA } \\
\midrule
\mr{40}{\rotatebox{90}{\textbf{FEDformer}}} & \mr{4}{Agriculture} & \hz{6} & 0.062 & \blue{0.063} & \blue{0.062} & \textbf{\red{0.057}} & \textbf{\red{0.061}} & \textbf{\red{0.062}} & \blue{0.604} & \blue{0.064} & \textbf{\red{0.062}} & \textbf{\red{0.055}} & \textbf{\red{0.059}} \\
 &  & \hz{8} & 0.071 & \blue{0.076} & \blue{0.074} & \textbf{\red{0.071}} & \blue{0.083} & \blue{0.073} & \blue{0.726} & \blue{0.075} & \blue{0.078} & \textbf{\red{0.069}} & \textbf{\red{0.069}} \\
 &  & \hz{10} & 0.092 & \blue{0.103} & \textbf{\red{0.090}} & \blue{0.094} & \blue{0.104} & \blue{0.096} & \blue{0.362} & \blue{0.095} & \blue{0.093} & \blue{0.092} & \blue{0.093} \\
 &  & \hz{12} & 0.115 & \blue{0.118} & \blue{0.116} & \blue{0.123} & \blue{0.131} & \blue{0.122} & \blue{0.316} & \blue{0.116} & \blue{0.122} & \blue{0.122} & \blue{0.117} \\
\cmidrule(lr){2-14}
 & \mr{4}{Climate} & \hz{6} & 1.116 & \textbf{\red{1.061}} & \textbf{\red{1.098}} & \blue{1.182} & \textbf{\red{1.071}} & \textbf{\red{1.086}} & \blue{1.261} & \textbf{\red{1.098}} & \textbf{\red{1.095}} & \textbf{\red{1.077}} & \textbf{\red{1.060}} \\
 &  & \hz{8} & 1.082 & \blue{1.091} & \textbf{\red{1.060}} & \blue{1.091} & \textbf{\red{1.030}} & \textbf{\red{1.063}} & \textbf{\red{1.051}} & \textbf{\red{1.055}} & \blue{1.093} & \textbf{\red{1.075}} & \blue{1.101} \\
\bottomrule
\end{tabular}
\end{adjustbox}
\end{table*}
\FloatBarrier
\begin{table*}[!htbp]
\caption{Comparison of multimodal fusion strategies with \textbf{Autoformer} (+ \textbf{GPT}).}
\centering
\begin{adjustbox}{width=1.00\linewidth}
\begin{tabular}{c|ll|c|ccc|ccc|cccc}
\toprule
\mr{2.5}{\shortstack{TS \\ Model}} & \mr{2.5}{Dataset} & \mr{2.5}{$H$} & \mr{2.5}{Unimodal} & \multicolumn{3}{c}{\textbf{Additive}} & \multicolumn{3}{c}{\textbf{Concat}} & \multicolumn{4}{c}{\textbf{Constrained}} \\
\cmidrule(lr){5-7}\cmidrule(lr){8-10}\cmidrule(lr){11-14}
& & & & First & Middle & Last & First & Middle & Last & Orthogonal & FiLM & Gating & \textbf{CFA } \\
\midrule
\mr{40}{\rotatebox{90}{\textbf{Autoformer}}} & \mr{4}{Agriculture} & \hz{6} & 0.070 & \blue{0.072} & \textbf{\red{0.066}} & \blue{0.075} & \textbf{\red{0.066}} & \textbf{\red{0.066}} & \blue{1.226} & \textbf{\red{0.069}} & \textbf{\red{0.069}} & \textbf{\red{0.062}} & \textbf{\red{0.068}} \\
 &  & \hz{8} & 0.090 & \textbf{\red{0.086}} & \textbf{\red{0.080}} & \textbf{\red{0.086}} & \blue{0.091} & \textbf{\red{0.078}} & \blue{0.483} & \blue{0.094} & \blue{0.097} & \textbf{\red{0.077}} & \blue{0.093} \\
 &  & \hz{10} & 0.110 & \blue{0.117} & \textbf{\red{0.106}} & \textbf{\red{0.107}} & \blue{0.115} & \blue{0.112} & \blue{0.348} & \blue{0.111} & \textbf{\red{0.102}} & \blue{0.113} & \textbf{\red{0.094}} \\
 &  & \hz{12} & 0.128 & \blue{0.144} & \blue{0.135} & \blue{0.139} & \blue{0.141} & \blue{0.134} & \blue{0.446} & \blue{0.131} & \blue{0.139} & \textbf{\red{0.126}} & \blue{0.129} \\
\cmidrule(lr){2-14}
 & \mr{4}{Climate} & \hz{6} & 1.240 & \blue{1.292} & \blue{1.243} & \textbf{\red{1.222}} & \textbf{\red{1.182}} & \textbf{\red{1.206}} & \blue{1.319} & \textbf{\red{1.228}} & \textbf{\red{1.239}} & \textbf{\red{1.206}} & \textbf{\red{1.204}} \\
 &  & \hz{8} & 1.200 & \blue{1.224} & \blue{1.252} & \blue{1.203} & \blue{1.293} & \textbf{\red{1.176}} & \textbf{\red{1.103}} & \blue{1.296} & \textbf{\red{1.178}} & \blue{1.221} & \textbf{\red{1.106}} \\
\bottomrule
\end{tabular}
\end{adjustbox}
\end{table*}

\begin{table*}[!htbp]
\caption{Comparison of multimodal fusion strategies with \textbf{Crossformer} (+ \textbf{GPT}).}
\centering
\begin{adjustbox}{width=1.00\linewidth}
\begin{tabular}{c|ll|c|ccc|ccc|cccc}
\toprule
\mr{2.5}{\shortstack{TS \\ Model}} & \mr{2.5}{Dataset} & \mr{2.5}{$H$} & \mr{2.5}{Unimodal} & \multicolumn{3}{c}{\textbf{Additive}} & \multicolumn{3}{c}{\textbf{Concat}} & \multicolumn{4}{c}{\textbf{Constrained}} \\
\cmidrule(lr){5-7}\cmidrule(lr){8-10}\cmidrule(lr){11-14}
& & & & First & Middle & Last & First & Middle & Last & Orthogonal & FiLM & Gating & \textbf{CFA } \\
\midrule
\mr{40}{\rotatebox{90}{\textbf{Crossformer}}} & \mr{4}{Agriculture} & \hz{6} & 0.088 & \blue{0.136} & \blue{0.218} & \blue{0.152} & \blue{0.153} & \blue{0.125} & \blue{0.221} & \blue{0.114} & \blue{0.179} & \blue{0.130} & \blue{0.176} \\
 &  & \hz{8} & 0.196 & \blue{0.219} & \blue{0.223} & \blue{0.206} & \blue{0.891} & \textbf{\red{0.176}} & \blue{0.347} & \blue{0.248} & \textbf{\red{0.103}} & \blue{0.207} & \blue{0.224} \\
 &  & \hz{10} & 0.248 & \blue{0.318} & \textbf{\red{0.218}} & \blue{0.385} & \blue{0.266} & \blue{0.413} & \blue{0.412} & \blue{0.300} & \blue{0.272} & \blue{0.276} & \blue{0.265} \\
 &  & \hz{12} & 0.311 & \blue{0.366} & \textbf{\red{0.253}} & \blue{0.411} & \blue{0.602} & \blue{0.410} & \blue{0.484} & \textbf{\red{0.242}} & \blue{0.315} & \blue{0.364} & \blue{0.323} \\
\cmidrule(lr){2-14}
 & \mr{4}{Climate} & \hz{6} & 1.039 & \textbf{\red{1.018}} & \blue{1.087} & \blue{1.091} & \blue{1.056} & \blue{1.061} & \blue{1.055} & \blue{1.080} & \blue{1.063} & \blue{1.061} & \blue{1.089} \\
 &  & \hz{8} & 1.080 & \textbf{\red{1.070}} & \textbf{\red{1.031}} & \blue{1.097} & \blue{1.113} & \blue{1.083} & \blue{1.084} & \blue{1.111} & \blue{1.092} & \blue{1.092} & \blue{1.120} \\
\bottomrule
\end{tabular}
\end{adjustbox}
\end{table*}

\begin{table*}[!htbp]
\caption{Comparison of multimodal fusion strategies with \textbf{Reformer} (+ \textbf{GPT}).}
\centering
\begin{adjustbox}{width=1.00\linewidth}
\begin{tabular}{c|ll|c|ccc|ccc|cccc}
\toprule
\mr{2.5}{\shortstack{TS \\ Model}} & \mr{2.5}{Dataset} & \mr{2.5}{$H$} & \mr{2.5}{Unimodal} & \multicolumn{3}{c}{\textbf{Additive}} & \multicolumn{3}{c}{\textbf{Concat}} & \multicolumn{4}{c}{\textbf{Constrained}} \\
\cmidrule(lr){5-7}\cmidrule(lr){8-10}\cmidrule(lr){11-14}
& & & & First & Middle & Last & First & Middle & Last & Orthogonal & FiLM & Gating & \textbf{CFA } \\
\midrule
\mr{40}{\rotatebox{90}{\textbf{Reformer}}} & \mr{4}{Agriculture} & \hz{6} & 0.202 & \blue{0.231} & \textbf{\red{0.147}} & \textbf{\red{0.152}} & \blue{0.255} & \textbf{\red{0.174}} & \blue{0.317} & \textbf{\red{0.150}} & \blue{0.231} & \blue{0.220} & \textbf{\red{0.163}} \\
 &  & \hz{8} & 0.221 & \blue{0.411} & \blue{0.302} & \blue{0.239} & \blue{0.242} & \blue{0.369} & \blue{0.416} & \blue{0.304} & \textbf{\red{0.187}} & \textbf{\red{0.191}} & \blue{0.233} \\
 &  & \hz{10} & 0.319 & \blue{0.525} & \blue{0.452} & \blue{0.416} & \blue{0.523} & \textbf{\red{0.314}} & \blue{0.714} & \blue{0.328} & \textbf{\red{0.251}} & \blue{0.457} & \textbf{\red{0.246}} \\
 &  & \hz{12} & 0.533 & \textbf{\red{0.514}} & \textbf{\red{0.392}} & \textbf{\red{0.496}} & \blue{0.615} & \blue{0.721} & \textbf{\red{0.377}} & \textbf{\red{0.316}} & \textbf{\red{0.277}} & \textbf{\red{0.212}} & \textbf{\red{0.241}} \\
\cmidrule(lr){2-14}
 & \mr{4}{Climate} & \hz{6} & 1.025 & \blue{1.102} & \blue{1.028} & \blue{1.091} & \blue{1.064} & \blue{1.114} & \blue{1.080} & \blue{1.038} & \blue{1.076} & \textbf{\red{1.011}} & \blue{1.082} \\
 &  & \hz{8} & 1.066 & \blue{1.089} & \blue{1.080} & \blue{1.069} & \textbf{\red{1.042}} & \blue{1.099} & \blue{1.075} & \textbf{\red{1.040}} & \textbf{\red{1.047}} & \blue{1.085} & \blue{1.087} \\
\bottomrule
\end{tabular}
\end{adjustbox}
\end{table*}

\begin{table*}[!htbp]
\caption{Comparison of multimodal fusion strategies with \textbf{TSMixer} (+ \textbf{GPT}).}
\centering
\begin{adjustbox}{width=1.00\linewidth}
%
\end{adjustbox}
\end{table*}
\FloatBarrier
\begin{table*}[!htbp]
\caption{Comparison of multimodal fusion strategies with \textbf{Transformer} (+ \textbf{GPT}).}
\centering
\begin{adjustbox}{width=1.00\linewidth}
%
\end{adjustbox}
\end{table*}

\begin{table*}[!htbp]
\caption{Comparison of multimodal fusion strategies with \textbf{Informer} (+ \textbf{GPT}).}
\centering
\begin{adjustbox}{width=1.00\linewidth}
%
\end{adjustbox}
\end{table*}

\begin{table*}[!htbp]
\caption{Comparison of multimodal fusion strategies with \textbf{Nonstationary Transformer} (+ \textbf{LLAMA3}).}
\centering
\begin{adjustbox}{width=1.00\linewidth}
%
\end{adjustbox}
\end{table*}

\begin{table*}[!htbp]
\caption{Comparison of multimodal fusion strategies with \textbf{PatchTST} (+ \textbf{LLAMA3}).}
\centering
\begin{adjustbox}{width=1.00\linewidth}
%
\end{adjustbox}
\end{table*}

\begin{table*}[!htbp]
\caption{Comparison of multimodal fusion strategies with \textbf{iTransformer} (+ \textbf{LLAMA3}).}
\centering
\begin{adjustbox}{width=1.00\linewidth}
%
\end{adjustbox}
\end{table*}

\begin{table*}[!htbp]
\caption{Comparison of multimodal fusion strategies with \textbf{Koopa} (+ \textbf{LLAMA3}).}
\centering
\begin{adjustbox}{width=1.00\linewidth}
%
\end{adjustbox}
\end{table*}
\FloatBarrier
\begin{table*}[!htbp]
\caption{Comparison of multimodal fusion strategies with \textbf{DLinear} (+ \textbf{LLAMA3}).}
\centering
\begin{adjustbox}{width=1.00\linewidth}
%
\end{adjustbox}
\end{table*}

\begin{table*}[!htbp]
\caption{Comparison of multimodal fusion strategies with \textbf{TiDE} (+ \textbf{LLAMA3}).}
\centering
\begin{adjustbox}{width=1.00\linewidth}
%
\end{adjustbox}
\end{table*}

\begin{table*}[!htbp]
\caption{Comparison of multimodal fusion strategies with \textbf{FiLM} (+ \textbf{LLAMA3}).}
\centering
\begin{adjustbox}{width=1.00\linewidth}
%
\end{adjustbox}
\end{table*}

\begin{table*}[!htbp]
\caption{Comparison of multimodal fusion strategies with \textbf{FEDformer} (+ \textbf{LLAMA3}).}
\centering
\begin{adjustbox}{width=1.00\linewidth}
%
\end{adjustbox}
\end{table*}
\FloatBarrier
\begin{table*}[!htbp]
\caption{Comparison of multimodal fusion strategies with \textbf{Autoformer} (+ \textbf{LLAMA3}).}
\centering
\begin{adjustbox}{width=1.00\linewidth}
%
\end{adjustbox}
\end{table*}

\begin{table*}[!htbp]
\caption{Comparison of multimodal fusion strategies with \textbf{Crossformer} (+ \textbf{LLAMA3}).}
\centering
\begin{adjustbox}{width=1.00\linewidth}
%
\end{adjustbox}
\end{table*}

\begin{table*}[!htbp]
\caption{Comparison of multimodal fusion strategies with \textbf{Reformer} (+ \textbf{LLAMA3}).}
\centering
\begin{adjustbox}{width=1.00\linewidth}
%
\end{adjustbox}
\end{table*}

\begin{table*}[!htbp]
\caption{Comparison of multimodal fusion strategies with \textbf{Informer} (+ \textbf{LLAMA3}).}
\centering
\begin{adjustbox}{width=1.00\linewidth}
%
\end{adjustbox}
\end{table*}
\FloatBarrier
\begin{table*}[!htbp]
\caption{Comparison of multimodal fusion strategies with \textbf{TSMixer} (+ \textbf{LLAMA3}).}
\centering
\begin{adjustbox}{width=1.00\linewidth}
%
\end{adjustbox}
\end{table*}

\begin{table*}[!htbp]
\caption{Comparison of multimodal fusion strategies with \textbf{Transformer} (+ \textbf{LLAMA3}).}
\centering
\begin{adjustbox}{width=1.00\linewidth}
%
\end{adjustbox}
\end{table*}

\begin{table*}[!htbp]
\caption{Comparison of multimodal fusion strategies with \textbf{Nonstationary Transformer} (+ \textbf{Doc2Vec}).}
\centering
\begin{adjustbox}{width=1.00\linewidth}
%
\end{adjustbox}
\end{table*}

\begin{table*}[!htbp]
\caption{Comparison of multimodal fusion strategies with \textbf{iTransformer} (+ \textbf{Doc2Vec}).}
\centering
\begin{adjustbox}{width=1.00\linewidth}
%
\end{adjustbox}
\end{table*}

\begin{table*}[!htbp]
\caption{Comparison of multimodal fusion strategies with \textbf{Koopa} (+ \textbf{Doc2Vec}).}
\centering
\begin{adjustbox}{width=1.00\linewidth}
%
\end{adjustbox}
\end{table*}

\begin{table*}[!htbp]
\caption{Comparison of multimodal fusion strategies with \textbf{DLinear} (+ \textbf{Doc2Vec}).}
\centering
\begin{adjustbox}{width=1.00\linewidth}
%
\end{adjustbox}
\end{table*}
\FloatBarrier
\begin{table*}[!htbp]
\caption{Comparison of multimodal fusion strategies with \textbf{PatchTST} (+ \textbf{Doc2Vec}).}
\centering
\begin{adjustbox}{width=1.00\linewidth}
%
\end{adjustbox}
\end{table*}

\begin{table*}[!htbp]
\caption{Comparison of multimodal fusion strategies with \textbf{FiLM} (+ \textbf{Doc2Vec}).}
\centering
\begin{adjustbox}{width=1.00\linewidth}
%
\end{adjustbox}
\end{table*}

\begin{table*}[!htbp]
\caption{Comparison of multimodal fusion strategies with \textbf{TiDE} (+ \textbf{Doc2Vec}).}
\centering
\begin{adjustbox}{width=1.00\linewidth}
%
\end{adjustbox}
\end{table*}

\begin{table*}[!htbp]
\caption{Comparison of multimodal fusion strategies with \textbf{FEDformer} (+ \textbf{Doc2Vec}).}
\centering
\begin{adjustbox}{width=1.00\linewidth}
%
\end{adjustbox}
\end{table*}
\FloatBarrier
\begin{table*}[!htbp]
\caption{Comparison of multimodal fusion strategies with \textbf{Autoformer} (+ \textbf{Doc2Vec}).}
\centering
\begin{adjustbox}{width=1.00\linewidth}
%
\end{adjustbox}
\end{table*}

\begin{table*}[!htbp]
\caption{Comparison of multimodal fusion strategies with \textbf{Crossformer} (+ \textbf{Doc2Vec}).}
\centering
\begin{adjustbox}{width=1.00\linewidth}
%
\end{adjustbox}
\end{table*}

\begin{table*}[!htbp]
\caption{Comparison of multimodal fusion strategies with \textbf{Reformer} (+ \textbf{Doc2Vec}).}
\centering
\begin{adjustbox}{width=1.00\linewidth}
%
\end{adjustbox}
\end{table*}

\begin{table*}[!htbp]
\caption{Comparison of multimodal fusion strategies with \textbf{Transformer} (+ \textbf{Doc2Vec}).}
\centering
\begin{adjustbox}{width=1.00\linewidth}
%
\end{adjustbox}
\end{table*}
\FloatBarrier
\begin{table*}[!htbp]
\caption{Comparison of multimodal fusion strategies with \textbf{TSMixer} (+ \textbf{Doc2Vec}).}
\centering
\begin{adjustbox}{width=1.00\linewidth}
%
\end{adjustbox}
\end{table*}

\begin{table*}[!htbp]
\caption{Comparison of multimodal fusion strategies with \textbf{Informer} (+ \textbf{Doc2Vec}).}
\centering
\begin{adjustbox}{width=1.00\linewidth}
%
\end{adjustbox}
\end{table*}








\end{document}